\documentclass[conference]{IEEEtran}  % Comment this line out if you need a4paper
\usepackage{times}

\usepackage[numbers]{natbib}
\usepackage{multicol}
\usepackage[bookmarks=true]{hyperref}
\usepackage{graphicx}
\usepackage[utf8]{inputenc} % allow utf-8 input
\usepackage[T1]{fontenc}    % use 8-bit T1 fonts
\usepackage{url}            % simple URL typesetting
\usepackage{booktabs}       % professional-quality tables
\usepackage{amsfonts}       % blackboard math symbols
\usepackage{nicefrac}       % compact symbols for 1/2, etc.
\usepackage{microtype}      % microtypography
\usepackage{xcolor}         % colors
\usepackage{graphicx}
\usepackage{makecell}
\usepackage{multirow}
\usepackage{floatrow}
\usepackage{wrapfig}
\usepackage{soul}
\usepackage{amsmath}
\usepackage{subcaption}
\usepackage{wrapfig,lipsum,booktabs}
\usepackage{xcolor,colortbl}
\usepackage{svg}
\usepackage{bbding}
\usepackage{enumitem}

\usepackage{booktabs} % For improved table aesthetics
\usepackage{adjustbox} % For resizing the table
\usepackage{tabularx} % For dynamic column width

\usepackage{acronym}
\acrodef{vl}[VL]{Vision-Language}
\acrodef{vla}[VLA]{Vision-Language-Action}
\acrodef{tdc}[TDC]{Task-Demonstration-Controller}
\acrodef{llm}[LLM]{Large Language Model}
\acrodef{ik}[IK]{Inverse Kinematics}
\acrodef{rl}[RL]{Reinforcement Learning}
\acrodef{rl}[IL]{Imitation Learning}

\usepackage{mydefinitions}
\usepackage{pifont} % For \ding{55} (X mark)
\usepackage{amssymb} % For \checkmark
\usepackage{tabularx}
\usepackage{xcolor}
\usepackage{amsmath}
\usepackage{amsfonts}
\usepackage{amssymb}
\usepackage{algorithm}
\usepackage{algorithmic}
\usepackage{xspace}
\usepackage{tikz}
\usepackage{pgfplots}
\usepackage{listings}  % pseudo code

\pgfplotsset{compat=1.18}

\IEEEoverridecommandlockouts                              % This command is only needed if 
\title{\huge{\textsc{RoboVerse}}: Towards a Unified Platform, Dataset and Benchmark for Scalable and Generalizable Robot Learning}

\author{
Haoran Geng\textsuperscript{1*},
Feishi Wang\textsuperscript{1,2,3*},
Songlin Wei\textsuperscript{2*},
Yuyang Li\textsuperscript{2,9*},
Bangjun Wang\textsuperscript{3*},
Boshi An\textsuperscript{2*},
\\
Charlie Tianyue Cheng\textsuperscript{1*},
Haozhe Lou\textsuperscript{3},
Peihao Li\textsuperscript{1,4},
Yen-Jen Wang\textsuperscript{1},
Yutong Liang\textsuperscript{2},
Dylan Goetting\textsuperscript{1},
\\
Chaoyi Xu\textsuperscript{2},
Haozhe Chen\textsuperscript{5},
Yuxi Qian\textsuperscript{6},
Yiran Geng\textsuperscript{2},
Jiageng Mao\textsuperscript{3},
Weikang Wan\textsuperscript{2},
Mingtong Zhang\textsuperscript{3},
\\
Jiangran Lyu\textsuperscript{2},
Siheng Zhao\textsuperscript{3},
Jiazhao Zhang\textsuperscript{2},
Jialiang Zhang\textsuperscript{1,2},
Chengyang Zhao\textsuperscript{7},
Haoran Lu\textsuperscript{2},
\\
Yufei Ding\textsuperscript{1,2},
Ran Gong\textsuperscript{8},
Yuran Wang\textsuperscript{2},
Yuxuan Kuang\textsuperscript{2,3},
Ruihai Wu\textsuperscript{2},
Baoxiong Jia\textsuperscript{9},
Carlo Sferrazza\textsuperscript{1},
\\
Hao Dong\textsuperscript{2},
Siyuan Huang\textsuperscript{9$\dagger$},
Yue Wang\textsuperscript{3$\dagger$},
Jitendra Malik\textsuperscript{1$\dagger$},
Pieter Abbeel\textsuperscript{1$\dagger$}
\\
\\
\textsuperscript{1}UC Berkeley~
\textsuperscript{2}PKU~
\textsuperscript{3}USC~
\textsuperscript{4}UMich~
\textsuperscript{5}UIUC~
\textsuperscript{6}Stanford~
\textsuperscript{7}CMU~
\textsuperscript{8}UCLA~
\textsuperscript{9}BIGAI
\\
{* equal contribution \quad $\dagger$ equal advising \quad 
Correspondence to: Haoran Geng <ghr@berkeley.edu>}
}

\begin{document}

\twocolumn[{
\renewcommand\twocolumn[1][]{#1}%
\begin{center}
    \maketitle
    \centering
    \captionsetup{type=figure}
    \includegraphics[width=\linewidth]{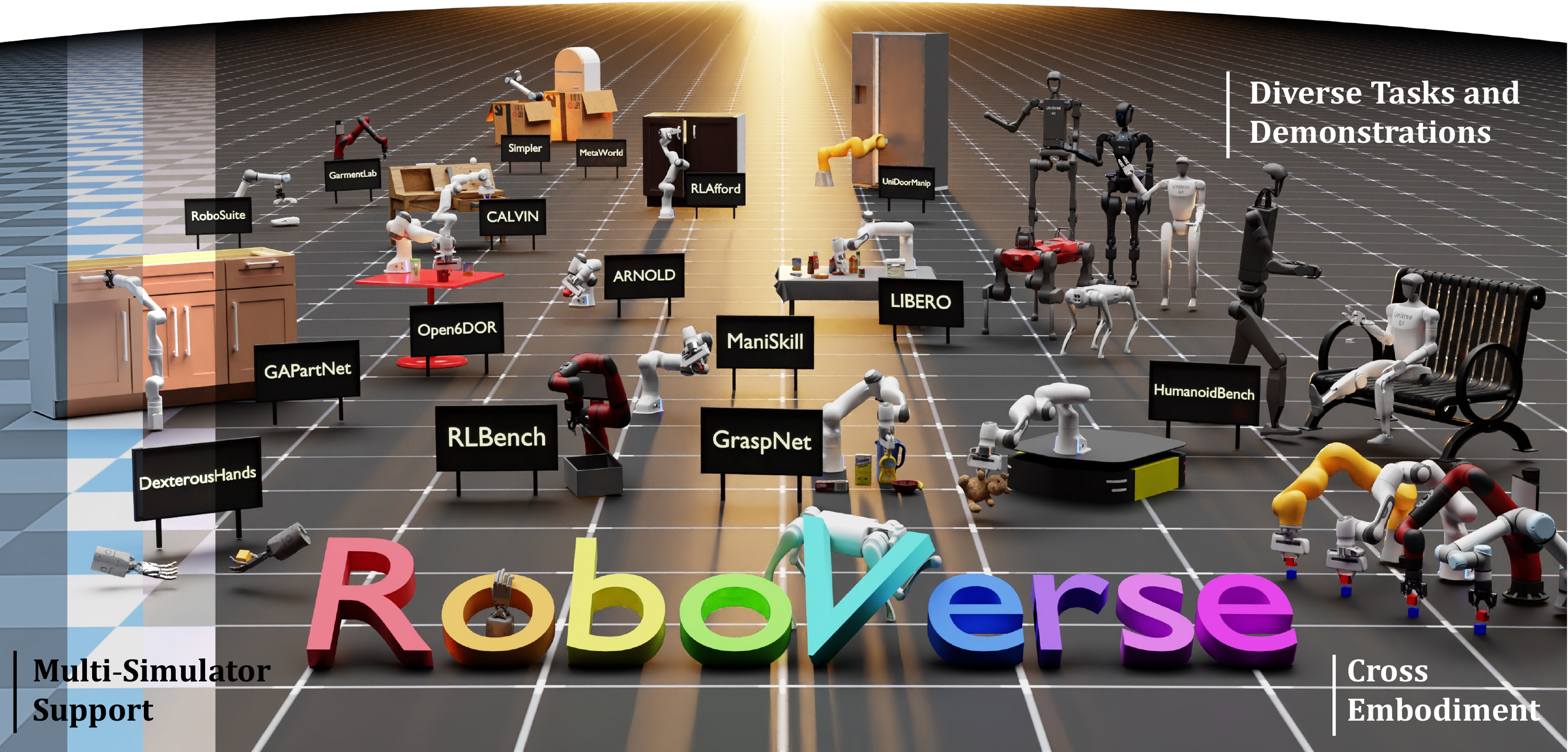}
    \captionof{figure}{
        \textsc{RoboVerse} comprises a scalable simulation platform, a large-scale synthetic dataset, and unified benchmarks. The simulation platform supports seamless integration of new tasks and demonstrations through unified protocols, ensuring flexibility and extensibility. The dataset includes over 1,000 diverse tasks and more than 10 million transitions, constructed through large-scale data migration, cross-embodiment transfer, and robust augmentation and randomization. 
    }
    \label{fig:teaser}
\end{center}
}]

\begin{abstract}
Data scaling and standardized evaluation benchmarks have driven significant advances in natural language processing and computer vision. However, robotics faces unique challenges in scaling data and establishing reliable evaluation protocols. Collecting real-world robotic data is resource-intensive and inefficient, while benchmarking in real-world scenarios remains highly complex. Synthetic data and simulation offer promising alternatives, yet existing efforts often fall short in data quality, diversity, and benchmark standardization.
To address these challenges, we introduce \textsc{RoboVerse}, a comprehensive framework comprising a \textit{simulation platform}, a \textit{synthetic dataset}, and \textit{unified benchmarks}. Our simulation platform supports multiple simulators and robotic embodiments, enabling seamless transitions between different environments. The synthetic dataset, featuring high-fidelity physics and photorealistic rendering, is constructed through multiple approaches including migration from public datasets, policy rollout, and motion planning, \etc enhanced by data augmentation. Additionally, we propose unified benchmarks for imitation learning and reinforcement learning, enabling consistent evaluation across different levels of generalization.
At the core of the \textit{simulation platform} is \textsc{MetaSim}, an infrastructure that abstracts diverse simulation environments into a universal interface. It restructures existing simulation environments into a simulator-agnostic configuration system, as well as an API aligning different simulator functionalities, such as launching simulation environments, loading assets with initial states, stepping the physics engine, \etc. This abstraction ensures interoperability and extensibility. 
Comprehensive experiments demonstrate that \textsc{RoboVerse} enhances the performance of imitation learning, reinforcement learning, and world model learning, improving sim-to-real transfer. These results validate the reliability of our dataset and benchmarks, establishing RoboVerse as a robust solution for advancing simulation-assisted robot learning.
Code and dataset can be found at: 
\href{https://roboverseorg.github.io/}{https://roboverseorg.github.io/}.

\end{abstract}

% \newpage

\section{Introduction}

{Large-scale datasets}, combined with {well-established benchmarks}, have fueled rapid advancements in natural language processing (NLP)~\cite{radford2019language, brown2020language} and computer vision (CV)~\cite{deng2009imagenet,krizhevsky2012imagenet,kirillov2023segment,ravi2024sam2segmentimages,liu2022hoi4d,he2022masked}.
Specifically, large-scale data provides ample training examples that bolster learning, while uniform benchmarks enable standardized evaluation and fair comparison across different methods. However, replicating these successes in robotics remains challenging due to the difficulty of collecting high-quality, diverse data and the lack of widely recognized evaluation protocols.

{Real-world approaches}~\cite{padalkar2023open,khazatsky2024droid} to constructing datasets and benchmarks, though authentically reflecting the complexities of operational environments, face significant practical constraints. First, collecting demonstrations is {time-consuming} and {resource-intensive}, and the resulting data is often hardware-dependent or modality-specific, limiting its adaptability to new scenarios. Additionally, establishing standardized and widely applicable benchmarks is inherently challenging since reproducing identical conditions for fair comparisons is nearly impossible. For instance, object placements can vary across rollouts, ambient lighting fluctuates under natural sunlight, and background environments may change. Consequently, scaling real-world datasets, evaluating policies, and iterating development in real-world scenarios remain cost-prohibitive and difficult to standardize.

{Simulators}, on the other hand, present a promising alternative for large-scale dataset and benchmark construction. By providing efficient computation, synthetic assets, and omniscient information in reproducible settings, they enable cost-effective dataset construction and consistent performance evaluation. Recent works, exemplified by~\cite{zhang2024dexgraspnet20learninggenerative,jiang2024dexmimicgen,chenobject,geng2023sage,rudin2022learningwalkminutesusing,xu2023unidexgrasp,lyu2024scissorbot, kuang2024ramretrievalbasedaffordancetransfer, wei2024droma, zhang2024dexgraspnet20learninggenerative, wan2023unidexgrasp++, li2024ag2manip, li2023manipllm, qi2024shapellm}, have demonstrated the potential of simulation-based methods in various robotic tasks. Despite these advantages, several challenges impede the broader adoption of synthetic datasets and benchmarks. 
First, utilizing simulators often {demands considerable expertise} due to both the complexity of simulator design and the relative immaturity of many platforms, which complicates the data construction process. Second, simulators vary widely in their internal architectures and external interfaces, making it {laborious to transfer data and models or adapt workflows} from one to another. 
Consequently, reusing existing synthetic datasets and benchmarks is difficult, resulting in {a fragmented ecosystem} that further hinders convenient construction and effective use of large-scale data in simulation environments.

To fully harness the potential of simulation in robotics, we introduce \textsc{RoboVerse}, a scalable simulation platform that {unifies existing simulators under a standardized format and a single infrastructure}, a large-scale synthetic dataset, and unified benchmarks. 
To achieve this, we first propose \textsc{MetaSim}, the core infrastructure of the \textsc{RoboVerse}. Through careful design, \textsc{MetaSim} establishes a universal configuration system for agents, objects, sensors, tasks, and physics parameters while exposing a simulator-agnostic interface for simulation setup and control. This architecture enables seamless integration of tasks, assets and robot trajectories from diverse simulation environments with minimal adaptation effort.
\textsc{MetaSim} provides three key capabilities:
(1) \textit{Cross-Simulator Integration}: Enables seamless switching between different simulators, fostering unified benchmarking and facilitating the transfer of environments and demonstrations across platforms.
(2) \textit{Hybrid Simulation}: Combines the strengths of multiple simulators---such as pairing advanced physics engines with superior renderers---to generate scalable and high-quality synthetic data.
(3) \textit{Cross-Embodiment Transfer}: Allows the retargeting of trajectories across various robot arms with parallel grippers, maximizing dataset reuse from heterogeneous sources.

\textsc{MetaSim} enables \textsc{RoboVerse} to systematically enhance the workflow for building and scaling simulation environments and datasets. 
Our method features:
\begin{itemize}
\item \textit{Scalable and Diverse Data Generation}: By aligning multiple benchmarks and task trajectories and leveraging a robust multi-source integration and data filtering pipeline, we generate large-scale, high-quality datasets. Additionally, our data randomization and augmentation pipeline enhances data diversity and volume, further enriching the dataset for comprehensive model training; 

\item \textit{Realistic Simulation and Rendering}: With \textsc{MetaSim}'s hybrid simulation capability, we enable the fusion of advanced physics engines and rendering systems across multiple simulators and renderers. Combined with carefully curated scenes, materials, and lighting assets, \textsc{RoboVerse} enhances realism in physical interactions and sensory observations;

\item \textit{Unified Benchmarking and Evaluation}: We unify widely used benchmarks into a cohesive system, streamlining algorithm development and performance comparison within a structured evaluation framework. Additionally, we introduce a standardized benchmarking protocol to assess varying levels of generalization and sim-to-real transferability.

\item \textit{Highly Extensibility and Scalability:} The aligned APIs and infrastructure streamline development and enable efficient algorithm integration, testing, and deployment across diverse simulation environments. Additionally, we develop real-to-sim frameworks, multiple teleoperation methods, and AI-generative systems for scalable task and data creation.
\end{itemize}

Leveraging these workflows in \textsc{RoboVerse}, we construct the largest and most diverse high-quality synthetic dataset and benchmark to date, all in a unified format. This dataset includes $\sim$500k unique, high-fidelity trajectories covering 276 task categories and $\sim$5.5k assets. Additionally, we generate over 50 million high-quality state transitions to support policy learning. 

Beyond dataset and benchmark construction, we explore the potential of \textsc{RoboVerse} through extensive experiments on imitation learning (\sref{sec:il_exp}), reinforcement learning (\sref{sec:rl_exp}), and world model learning (\sref{sec:world_model_exp}). Our results demonstrate that \textsc{RoboVerse} enables reliable policy learning and evaluation, supports strong sim-to-sim and (\sref{sec:exp_sim2sim2real}) sim-to-real transfer  (\sref{sec:exp_sim2real}) via high-fidelity physics and rendering, and facilitates efficient data expansion through teleoperation  (\sref{sec:dataset_teleop}), trajectory augmentation  (\sref{sec:data_traj_augmentation}), domain randomization (\sref{sec:data_domain_randomization}) and generative models (\sref{sec:data_llm}). These findings highlight the framework's robustness, scalability, and real-world applicability.

\section{Related Work}

\subsection{Robotics Simulators}
Advancements in computer graphics have contributed to the development of high-fidelity simulators, which are widely used in robotics research and development. CoppeliaSim~\cite{coppeliaSim}, Bullet~\cite{coumans2021pybullet}, and MuJoCo~\cite{todorov2012mujoco} provide accurate physics simulations and are extensively utilized in applications such as reinforcement learning and robotic benchmarking~\cite{blum2020rl, yang2021open, panerati2021learning, chiappa2024latent}. More simulators have been developed to fully exploit parallelism for better efficiency. Isaac Gym~\cite{makoviychuk2021isaacgym}, Isaac Sim~\cite{IsaacSim}, SAPIEN~\cite{gu2023maniskill2,tao2024maniskill3}, MuJoCo MJX~\cite{todorov2012mujoco, zakka2025mujocoplayground}, and Genesis~\cite{xian24genesis} utilize GPU power for enhanced performance, enabling large-scale reinforcement learning and efficient data collection, significantly improving training speed and scalability. Some simulators focus on bridging the simulation-reality gap (Sim-to-Real Gap), incorporating technologies including ray-tracing and customized renderers for photo-realistic rendering~\cite{IsaacSim,tao2024maniskill3}.
% introduce highly parallel simulation capabilities, leveraging GPU power for enhanced performance.
% Parallel simulation enables large-scale reinforcement learning and efficient data collection, significantly improving training speed and scalability. 
Furthermore, Isaac Sim~\cite{IsaacSim} and Genesis~\cite{xian24genesis} offer high-fidelity soft-body and liquid simulation, expanding the scope of realistic robotic interactions. \textsc{RoboVerse} proposes a unified platform that supports multiple simulators, facilitating seamless transitions between them and enabling hybrid integration to utilize the strengths of each simulator.

\subsection{Large-Scale Robotics Dataset}

The scarcity of large-scale, high-quality, and diverse datasets in the robotics community has long been recognized. 
Several works have shown the possibility of collecting demonstration data directly on real robots. RoboNet~\cite{dasari2019robonet} is a large-scale manipulation dataset containing roughly 162k trajectories from multiple robot platforms. DROID~\cite{khazatsky2024droid} has collected over 76k contact-rich robotic manipulation demonstrations across 86 tasks. RH20T~\cite{fang2024rh20t} proposed a dataset with over 100k demonstrations and 147 tasks. At the same time, RT-1~\cite{brohan2022rt} set the record further to 130k demonstrations on over 700 tasks. Recently, Open X-Embodiment~\cite{padalkar2023open} has demonstrated a promising approach to unite the community’s efforts, collecting over 1M trajectories on 160,266 tasks with 22 different embodiments.
At this stage, real-world datasets became difficult to scale up due to the proportional effort and cost required to collect more demonstrative trajectories.
% Improving the quality of robotics datasets also demands better sensors and equipment. Thus, the academic race on general robotic datasets has arrived at a pause, while companies and institutions from the industry continue to push the boundary~\cite{contributors2024agibotworldrepo, mitash2023armbench}.

Simulation-based data collection provides a promising solution to the high cost and inefficiencies of real-world datasets. Hussing \etal~\cite{hussing2023robotic} proposed a dataset containing 256M transitions on 256 tasks for offline compositional reinforcement learning. RoboCasa~\cite{nasiriany2024robocasa} introduced a dataset of 100 tasks and over 100k trajectories for generalist robots. DexGraspNet-2.0~\cite{zhang2024dexgraspnet} has collected over 400M demonstrations for dexterous grasping. Despite these efforts, synthetic datasets often exist in disparate simulators, leading to a fragmented ecosystem with limited diversity and quality. Moreover, simulation-based data often fails to capture complex physics and diverse task variations found in the real world~\cite{li24simpler,erez2015simulation}, potentially causing overfitting to specific simulators and hampering generalization to real-world scenarios.

\textsc{RoboVerse} provides a unified solution for large-scale, high-quality, and diverse synthetic data. It enables agents to train on a large set of environments and simulators to reduce overfitting, thereby improving the robustness of the learned policies.

% Companies and institutions from the industry continue to push the boundary. AgiBot-World~\cite{contributors2024agibotworldrepo} released a large-scale high-quality dataset with 

% and may lose utility when robot control systems, morphologies, or sensors are changed or upgraded.
% Besides this, several datasets~\cite{nasiriany2024robocasa, zhang2024dexgraspnet} collected through simulators show some possible ways to handle these problems.\yufei{problems with sim data: too little(demo trajs), fragmented(across different simulators), } However, training with data from a single simulator can lead to overfitting to a specific physics engine or rendering style. To address this, \textsc{RoboVerse} standardizes APIs across multiple simulators, enabling agents to train in diverse environments. This approach helps prevent overfitting to any one simulator, thereby improving the robustness of the learned policies.

\subsection{Benchmarking in Robotics}

Benchmarking remains a critical yet highly challenging problem in the robotics community. Compared to supervised learning tasks, it is relatively difficult to evaluate the performance of a robotics model. Meta-World~\cite{yu2019metaworld} is an early attempt in multi-task benchmarking. This is followed by RLBench~\cite{james2019rlbench}, BEHAVIOR-1K~\cite{li2023behavior}, Habitat~\cite{szot2021habitat}, and ManiSkill~\cite{mu2021maniskill,gu2023maniskill2,tao2024maniskill3,shukla2024maniskillhabbenchmarklowlevelmanipulation}, covering a large variety of robotic tasks. Grutopia~\cite{wang2024grutopia} and InfiniteWorld~\cite{ren2024infiniteworld} make a leap toward general-purpose robot benchmarking.

Despite significant efforts dedicated to these benchmarks, it is not guaranteed that the results are reproducible across different benchmarks. The uncertainty comes from multiple aspects including simulation accuracy, rendering style and asset properties~\cite{li24simpler, erez2015simulation}.
To address these challenges, \textsc{RoboVerse} enables researchers to evaluate their policies across multiple benchmarks and simulators seamlessly, without familiarizing themselves with each one individually.

\section{Infrastructure: \textsc{MetaSim}}

\subsection{\textsc{MetaSim} Overview}

\begin{figure}[tb] \centering
    \includegraphics[width=0.7\linewidth]{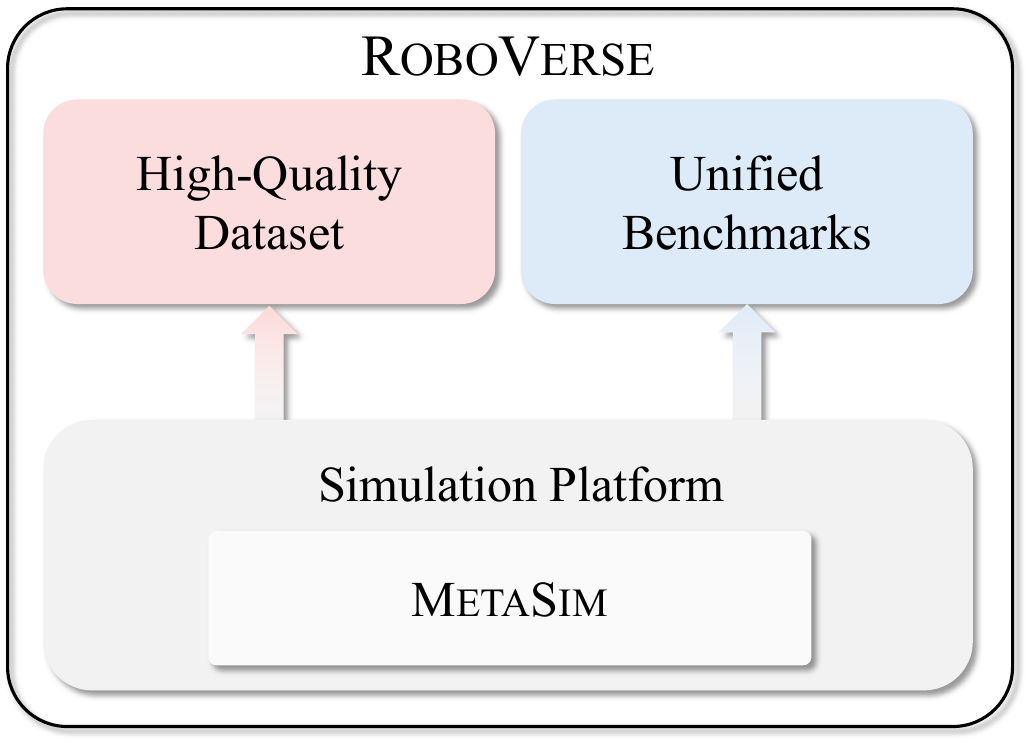}
    \caption{\textsc{RoboVerse} consists of a simulation platform, a large-scale, high-quality dataset, and unified benchmarks. At the core of the simulation platform is \textsc{MetaSim}, the infrastructure of \textsc{RoboVerse}. Powered by \textsc{MetaSim}, the simulation platform facilitates dataset creation and benchmark construction.
    % beyond what existing simulation environments can readily support
    }
    \label{fig:overview}
\end{figure}

\begin{figure*}[htb] \centering
    \includegraphics[width=0.98\linewidth]{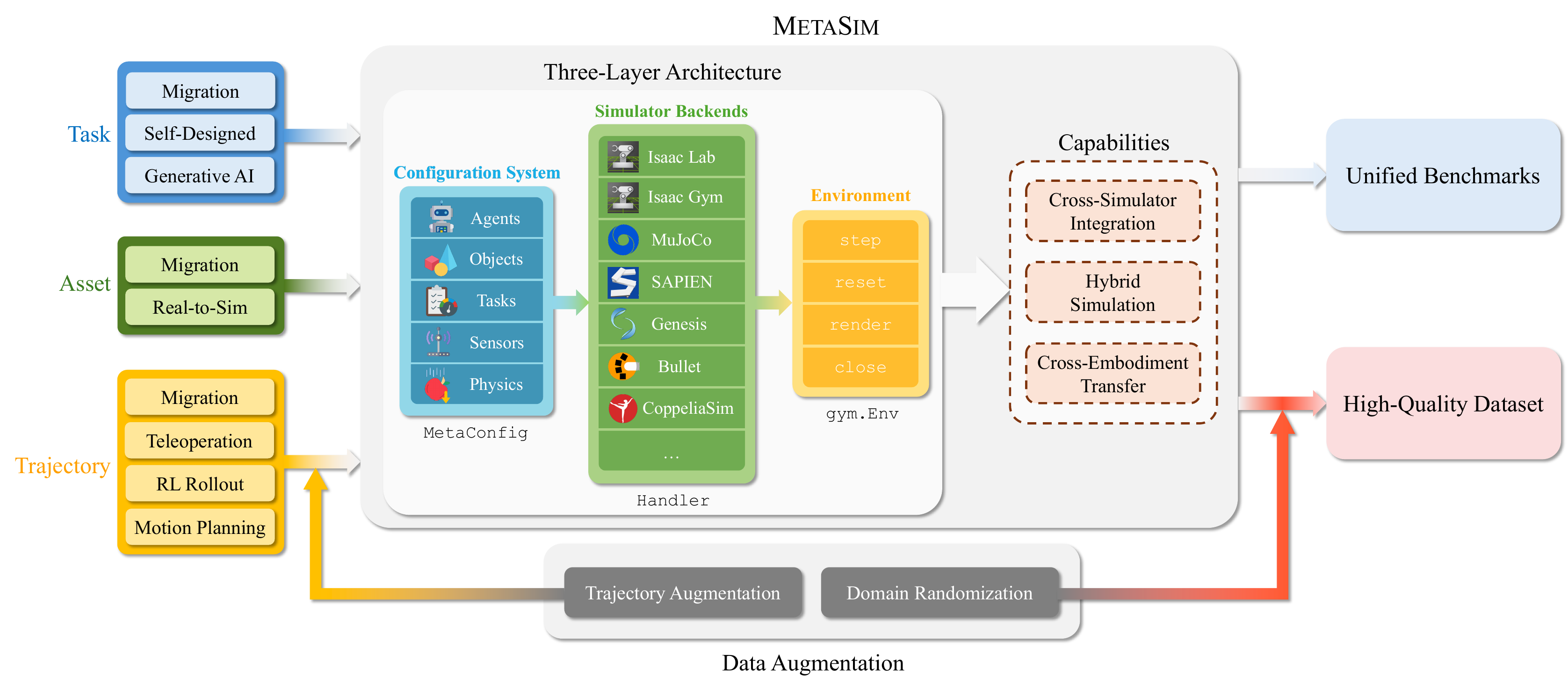}
    \caption{\textsc{MetaSim} provides a universal configuration system, aligned simulator backends, and a Gym~\cite{towers2024gymnasium} environment wrapper. This three-layer architecture abstracts simulation environments into simulator-agnostic specifications and aligns simulator backends, enabling three key capabilities: {cross-simulator integration}, {hybrid simulation} and {cross-embodiment transfer}. Based on \textsc{MetaSim}, we build a pipeline to collect tasks, assets and trajectories from diverse public sources in a unified format, employ data augmentation methods, and ultimately generate a large-scale high-quality dataset along with unified benchmarks. This data pipeline forms the foundation of \textsc{RoboVerse}, facilitating the generation of large-scale datasets and construction of unified benchmarks.
    % \textsc{MetaSim} with three capabilities facilitates the data pipeline that previous simulation environments find hard to support.
    } 
    \label{fig:workflow}
\end{figure*}

% Existing simulation environments provide diverse datasets and benchmarks, but they remain fragmented across different simulators with incompatible formats. 
% This fragmentation is due to their tight coupling with particular simulation environments.
% To build a large-scale high-quality dataset and unified benchmarks, 
We present \textsc{MetaSim}, a high-level interface above specific simulation environment implementations.
It is also the core infrastructure of \textsc{RoboVerse}. 
As illustrated in \fref{fig:overview}, \textsc{MetaSim} empowers the \textsc{RoboVerse} simulation platform, allowing for the generation of a large-scale high-quality dataset, as well as the construction of a unified benchmark.

\subsection{\textsc{MetaSim} Implementation}

As illustrated in \fref{fig:workflow}, \textsc{MetaSim} employs a three-layer architecture including a universal configuration system, a simulator-agnostic interface, and a user-friendly environment wrapper.
The universal configuration system unifies specifications for a simulation scenario and ensures consistent format across simulators.
The simulator-agnostic interface interprets these specifications, translates them into simulator-specific commands, and therefore aligns different simulator backends.
In addition, the environment wrappers encapsulate the simulator-agnostic interface into a standarized learning environment, such as a Gym~\cite{towers2024gymnasium} environment.
We describe each layer with more details in the following sections.
% \begin{itemize}
    % \item A universal configuration system that unifies specifications for a simulation scenario and ensures consistent format across simulators;
    % \item A simulator-agnostic interface that interprets these specifications, translates them into simulator-specific commands, and therefore aligns different simulator backends; 
    % \item User-friendly wrappers that encapsulate the simulator-agnostic interface into a standarized learning environment, \eg, Gym~\cite{towers2024gymnasium} environment. 
% \end{itemize}

\subsubsection{Universal Configuration System}

\begin{figure}[tb] \centering
    \includegraphics[width=0.98\linewidth]{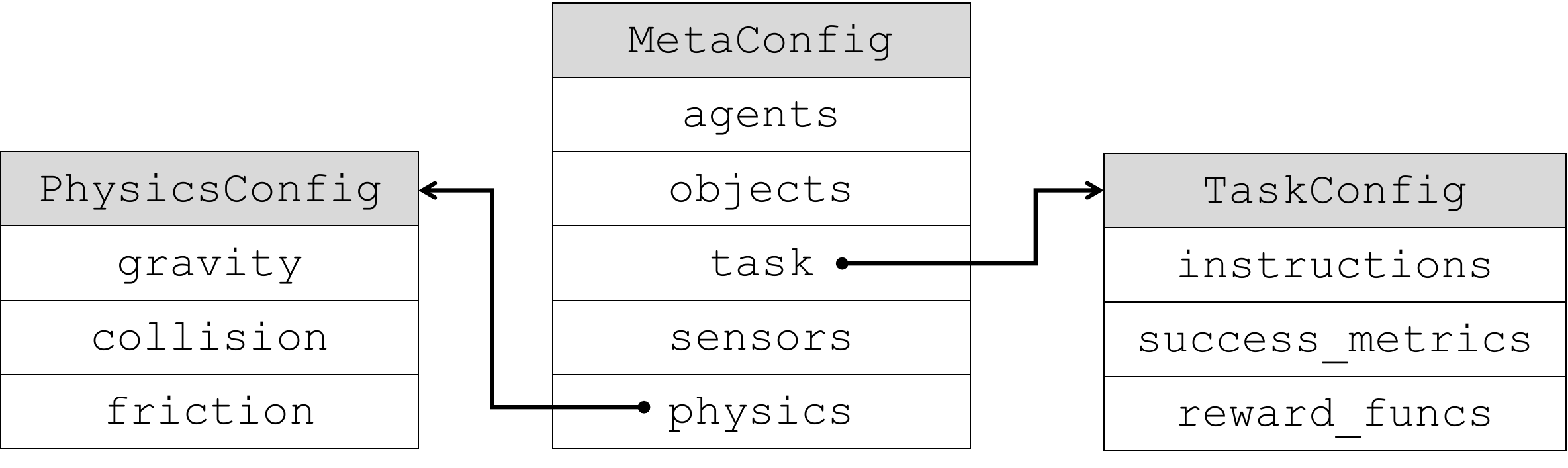}
    \caption{The \texttt{MetaConfig} is a nested dataclass that abstracts the core components in any simulation environment in a simulator-agnostic way.}
    \label{fig:metacfg}
\end{figure}

% Any scenario in a simulation environment has the following crucial components: agents, objects, tasks, sensors, and physics parameters. 
A typical simulation environment comprises agents, objects, tasks, sensors, and physics parameters.
They collectively define who performs the actions (agents), what the environment looks like (objects), what the agents should do (tasks, including instructions, success metrics, and rewards), how the environment is perceived and measured (sensors), and the governing physical laws (physics parameters). 
Ideally, these components should be simulator-agnostic, requiring a unified standard of simulation scenarios. Such a standard would enable researchers to work across different simulators seamlessly and integrate existing efforts from the community through cross-simulation.

Based on such a principle, we design a configuration system, \texttt{MetaConfig}, to abstract simulation scenarios in a simulator-agnostic way. As illustrated in \fref{fig:metacfg}, \texttt{MetaConfig} is a nested class that contains the above-mentioned core components. It can be interpreted by different simulator backends to build the corresponding simulation.
Additionally, \texttt{MetaConfig} supports optional simulator-specific hyperparameters (\eg, solver type), allowing fully leveraging the unique features of different simulators through customization.

\subsubsection{Aligned Simulator Backends}

Different simulators have their own implementations and specializations. However, routine operations -- such as initializing a scene, loading objects, stepping the physics engine, retrieving observations, time management, and determining success states -- tend to follow similar patterns.
To standardize these shared operations, we create a unified interface through a \texttt{Handler} class. 
Each simulator has its own handler instance implementing this interface. 
% The base class, \texttt{BaseHandler}, defines the abstract interface, while specific simulator handlers (\eg, \texttt{SapienHandler}) implement its abstract methods using the native APIs of their respective simulators (\eg, SAPIEN~\cite{xiang2020sapien} APIs).
The handler class implements the common methods including \texttt{launch()}, \texttt{get\_states()}, and \texttt{set\_states()}, \etc, spanning the whole lifecycle of simulating a task.
The usage of the APIs is illustrated in \Cref{code:env}. More information is provided in the supplementary materials.

% Please refer to 
% - https://www.overleaf.com/learn/latex/Code_listing
% - https://tex.stackexchange.com/a/331807
% - https://tex.stackexchange.com/a/279245

\definecolor{main-color}{rgb}{0.6627, 0.7176, 0.7764}
\definecolor{back-color}{rgb}{0.1686, 0.1686, 0.1686}
\definecolor{string-color}{rgb}{0.3333, 0.5254, 0.345}
\definecolor{key-color}{rgb}{0.8, 0.47, 0.196}

\definecolor{codegreen}{rgb}{0,0.6,0}
\definecolor{codegray}{rgb}{0.5,0.5,0.5}
\definecolor{codepurple}{rgb}{0.58,0,0.82}
\definecolor{backcolour}{rgb}{0.95,0.95,0.95}

\lstdefinestyle{mystyle}
{
    language = Python,
    basicstyle = \ttfamily\footnotesize,
    backgroundcolor = {\color{backcolour}},
    stringstyle = {\color{string-color}},
    keywordstyle = {\color{key-color}},
    keywordstyle = [2]{\color{codegreen}},
    otherkeywords = {},
    morekeywords = [2]{handler},
    breakatwhitespace=false,         
    breaklines=true,
    keepspaces=true,
    %% Padding
    % reference
    % - https://tex.stackexchange.com/a/715207
    % - https://tex.stackexchange.com/a/167950
    frame=bt,
    framerule=0pt,
    framextopmargin=3mm, 
    framexbottommargin=3mm, 
    framexleftmargin=5mm, xleftmargin=5mm,
}

\begin{lstfloat}[bt]
\begin{lstlisting}[style=mystyle]
class Env:
    def __init__(self, handler):
        self.handler = handler
        handler.launch()

    def reset(self):
        handler.set_states()
        states = handler.get_states()
        return get_observation(states), \
               handler.get_extra()

    def step(self, action):
        handler.set_states(action=action)
        handler.step()
        states = handler.get_states()
        return get_observation(states), \
               get_reward(states), \
               get_success(states) \
               get_termination(states), \
               get_time_out(states), \
               handler.get_extra()

    def render(self):
        return handler.render()

    def close(self):
        handler.close()
\end{lstlisting}
\caption{Pseudocode for \texttt{gym.Env} implementation. Each method of \texttt{gym.Env} is implemented by calling the corresponding methods of the \texttt{Handler} class.}
\label{code:env}
\end{lstfloat}

\subsubsection{User-Friendly Environment Wrapper}

Gym~\cite{towers2024gymnasium} is a widely adopted paradigm in reinforcement learning and robotics, in which the \texttt{gym.Env} class is fundamental to building learning environments. 
We define a wrapper to easily transform a \texttt{Handler} into an environment equipped with Gym APIs (\texttt{step()}, \texttt{reset()}, \texttt{render()}, and \texttt{close()}). 
As shown in \Cref{code:env}, these methods are implemented by leveraging the underlying \texttt{Handler} methods.

\subsection{\textsc{MetaSim} Capabilities}
\textsc{MetaSim} offers the following three key capabilities.

\subsubsection{Cross-Simulator Integration} 
Seamlessly switching between different simulators, allowing tasks and trajectories from one simulator to be utilized in other simulators. This capability enables efficient task and trajectory integration, unified benchmark construction, and sim-to-sim transfer for reinforcement learning training. 
% With this capability, imagine you can design tasks using MuJoCo~\cite{todorov2012mujoco} on your Macbook, and then seamlessly sync it to the server to render high-quality images using Isaac Sim~\cite{IsaacSim}. 
For example, tasks from Meta-World~\cite{yu2019metaworld} can be used by Isaac Gym~\cite{makoviychuk2021isaacgym} for fast parallel training, after which the generated trajectories can be deployed in Isaac Sim~\cite{IsaacSim} for rendering.

\subsubsection{Hybrid Simulation}
\textsc{MetaSim} supports combining the physics engine of one simulator and the renderer of another simulator at the same time, allowing users to benefit from advantages owned by different simulators. Specifically, using a single command, one could launch a simulator with a powerful renderer (\eg, Isaac Sim~\cite{IsaacSim}) with a simulator that has an accurate physics engine (\eg, MuJoCo~\cite{todorov2012mujoco}) to form an even more powerful simulation, enabling high-quality data generation.

\subsubsection{Cross-Embodiment Transfer}
Reusing the trajectories across different gripper-based robot morphologies by retargeting the end-effector pose, which allows the integration of data collected from diverse robots into a unified format. 
\section{\textsc{RoboVerse} Dataset}

\subsection{Dataset Overview}
On top of \textsc{MetaSim}, we generate large-scale high quality dataset by incorporating
multiple data collection methods. Overall, there are three key data types to collect:
tasks, assets, and robot trajectories. The main source of these data is migration
from existing simulation environments. Beyond migration, we explore various methods
to collect these data, such as using large language models to generate new tasks,
leveraging the real-to-sim toolset~\cite{lou2024robogsphysicsconsistentspatialtemporal}
to reconstruct assets from the real world, using teleoperation to collect new
trajectories, \etc. Additionally, we leverage data augmentation methods for both
trajectories and visual observations. Finally, we report the statistics for current
progress of data migration in \textsc{RoboVerse}.

\subsection{Tasks, Assets and Trajectories Collection: Migration}
Leveraging the \textsc{RoboVerse} format and infrastructure, we seamlessly integrate a wide
range of benchmarks and datasets into our system with a unified format and clean
codebase. We apply the following approaches to collect tasks and demonstrations.

\begin{itemize}
    \item \textbf{Direct Migration from Other Simulation Environments}

        Some benchmarks provide essential components % \yufei{do we need to define these components beforehand? or just say "according to \textsc{MetaConfig}}, enabling direct
        integration into \textsc{RoboVerse}. We define environment
        configurations for task initialization and evaluation, then convert trajectory
        data and asset formats for seamless compatibility. Notably, \textsc{RoboVerse}
        streamlines this migration process by first aligning formats in the
        original simulator and automatically ensuring compatibility across all
        simulators.

    \item \textbf{Motion Planning and RL Rollout} When benchmarks provide only
        partial manipulation data, such as keypoint trajectories or grasping poses,
        we use motion planning to generate complete trajectories. If no explicit
        manipulation data is available but pre-existing policies or reinforcement
        learning frameworks exist, we either utilize these policies or train new
        ones to collect demonstration data through rollouts. To ensure high data
        quality and consistency with our system standards, we carefully adapt
        the success checker and rigorously filter both planned and collected
        trajectories.
\end{itemize}

With the techniques mentioned above, we migrated multiple existing manipulation
datasets into \textsc{RoboVerse}. Currently, we support ManiSkill~\cite{mu2021maniskill,gu2023maniskill2,tao2024maniskill3},
RLBench~\cite{james2019rlbench}, CALVIN~\cite{mees2022calvin}, Meta-World~\cite{yu2019metaworld},
\texttt{robosuite}~\cite{zhu2020robosuite}, MimicGen~\cite{mandlekar2023mimicgen},
% RoboCasa~\cite{nasiriany2024robocasa},
GAPartNet~\cite{geng2023gapartnet}, Open6DOR~\cite{ding2024open6dor}, ARNOLD~\cite{gong2023arnold},
LIBERO~\cite{liu2023libero}, SIMPLER~\cite{li24simpler}, GraspNet~\cite{fang2020graspnet},
GarmentLab~\cite{lu2024garmentlab}, and UniDoorManip~\cite{li2024unidoormanip}.

We also integrated datasets from a wider range of embodiments, including
dexterous hands, quadrupeds, and humanoids, covering tasks such as dexterous manipulation,
locomotion, navigation, and whole-body control. Currently, we have migrated VLN-CE
R2R~\cite{Krantz2020BeyondTN} and RxR~\cite{ku2020room} for navigation, as well as
HumanoidBench~\cite{sferrazza2024humanoidbench} and Humanoid-X~\cite{uh1} for locomotion
and whole-body control.

\textsc{RoboVerse} simplifies and standardizes the migration process, and we will
continue to maintain and expand it.

\subsection{Tasks, Assets and Trajectories Collection: Teleoperation and Generation}
\label{sec:dataset_teleop}
\begin{itemize}[leftmargin=*]
    \item \textbf{Teleoperation System for Trajectory Collection}
        \label{sec:dataset _teleop}. As shown in ~\fref{fig:teleop}, \textsc{RoboVerse}
        integrates teleoperation systems within the \textsc{MetaSim}
        infrastructure, offering a flexible and efficient solution for high-quality
        data collection. It supports various robotic systems, including arms,
        dexterous hands~\cite{pavlakos2024reconstructing}, and bimanual setups,
        enabling seamless teleoperation across different simulators. To mitigate
        the high cost and complexity of professional equipment, we introduce an
        interactive motion control system utilizing accessible devices such as
        keyboards, joysticks, mobile apps (we developed a new app for Android and
        iOS to control robotic arms; see supplementary materials for more
        details.), motion capture (Mocap)~\cite{vlasic2007practical}, and VR
        systems~\cite{cheng2024open,qin2023anyteleop}. These devices' integrated
        sensors capture motion data, allowing natural, gesture-based control
        along with real-time, high-frequency communication for precise, low-cost
        remote operation. Further details are provided in the supplementary materials.

    \item \textbf{AI-Assisted Task Generation}. \label{sec:data_llm}Leveraging
        the generalization capability of large generative models, AI-assisted task
        generation provides a mechanism to diversify task varieties and scenario
        distribution. By learning from example placements, it acquires a sense of
        spatial and semantic constraints~\cite{antero2024harnessing} (\eg by
        demonstrating specific constraints, it can learn to spread out objects to
        avoid potential overlap \etc.). It can arrange
        objects originally from different benchmarks into a physically plausible
        scenes based on \textsc{MetaSim}, as shown in ~\fref{fig:ai-assisted}. Incorporating
        randomization in robot and object selection~\cite{katara2024gen2sim} with
        their initial poses, large generative models can generate various initial states. The system can automatically
        output all the required configuration files in unified format for instant
        visualization and user-friendly editing. After task generation, we will process
        a two-step filtering to avoid errors and hallucinations: (1) \textit{Format
        Validation}: Tasks that fail to meet \textsc{RoboVerse} format standards
        are discarded. (2) \textit{Feasibility Check}: Since trajectory data is
        collected via human teleoperation, tasks deemed unreasonable by the teleoperator
        are removed. By unleashing the extrapolative and few-shot learning abilities of
        large generative models, we integrate assets under a uniform schema automatically,
        driving task generation that spans multiple simulators and
        benchmarks.

    \item \textbf{Real-to-Sim for Asset Construction}. Video-based reconstruction
        proves to be a valuable source for data and asset creation by leveraging
        Real-to-Sim techniques. Our approach integrates multiple reconstruction
        pipelines to extract high-fidelity assets from video data. First, we
        initialize the structure using COLMAP~\cite{schoenberger2016sfm,schoenberger2016mvs}
        and employ Gaussian Splatting~\cite{kerbl3Dgaussians} for high-quality
        rendering. Next, we infer physical properties by feeding both semantic
        and original images into a Vision-Language Model (VLM)~\cite{zhou2022learning}.
        For geometry reconstruction, we estimate surface normals from video~\cite{ye2024stablenormal},
        apply surfel splatting~\cite{Huang2DGS2024}, and utilize TSDF-based
        methods with dynamic filtering to reconstruct detailed meshes~\cite{ye2024gaustudio}.
        By leveraging semantic masks~\cite{ravi2024sam2segmentimages}, we
        selectively extract components from both Gaussian and mesh
        representations. To further enhance realism, we infer and learn object
        kinematics directly from video~\cite{liu2024differentiablerobotrendering},
        ensuring accurate motion representations. Finally, we formulate URDF models
        by refining key attributes such as coordinate frames, orientation, axis alignment,
        scale, relative 6-DoF poses, and PD control parameters~\cite{lou2024robogsphysicsconsistentspatialtemporal}.
        This pipeline effectively bridges the gap between real-world video data and
        simulation-ready assets, enhancing robotic learning and simulation
        fidelity. We also present comparative experiments in the supplementary materials,
        demonstrating that our methods significantly enhance real-world policy
        performance.

        \begin{figure}[bt]
            \centering
            \includegraphics[width=1.0 \linewidth]{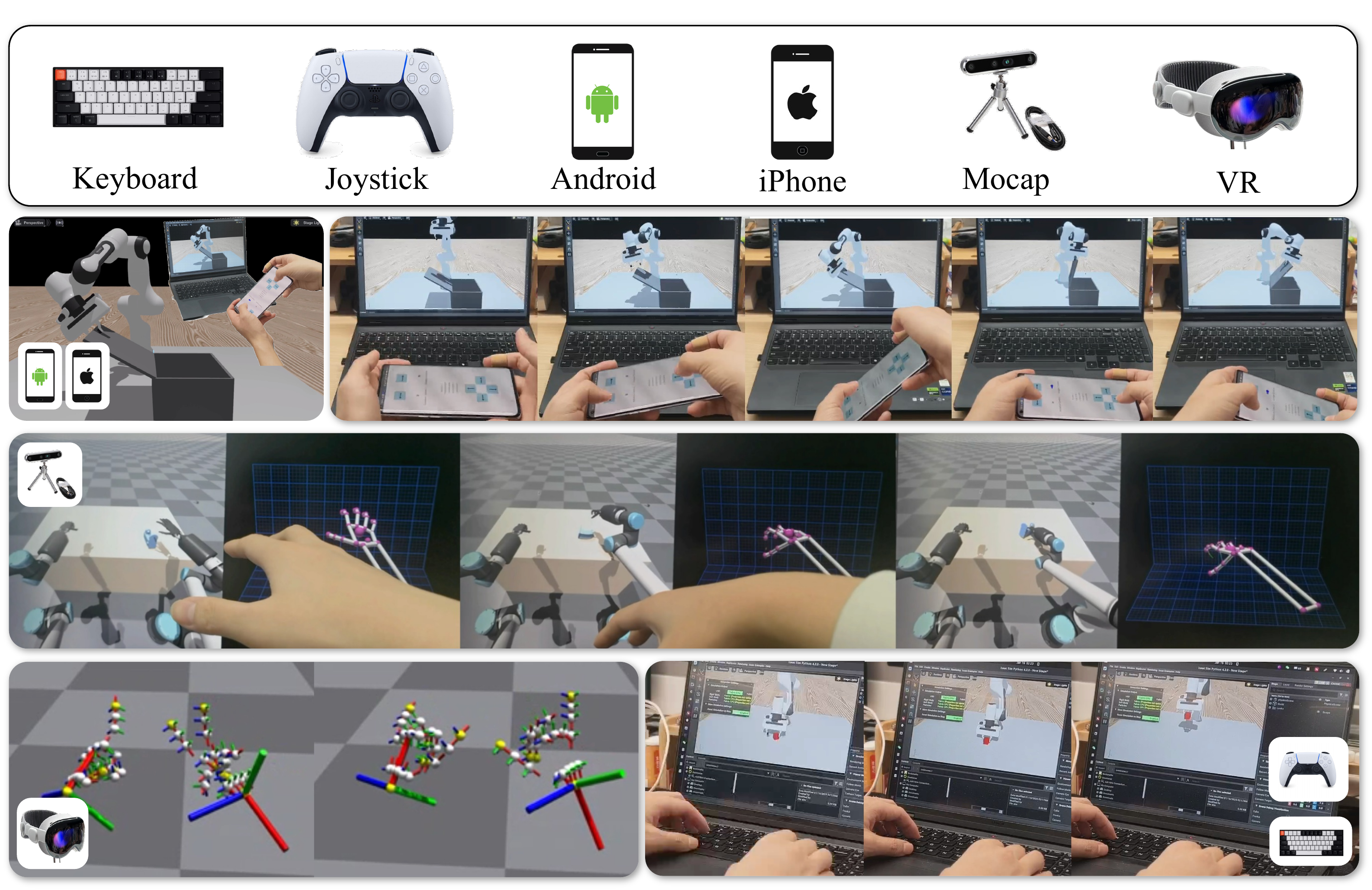}
            \caption{\textbf{Teleoperation System.} \textsc{RoboVerse} supports
            various user-friendly teleoperation approaches. Currently, it
            enables teleoperation via a phone app (second row), motion capture (middle),
            VR devices (bottom left), as well as keyboard and joystick (bottom
            right). These methods allow control of robotic arms, dexterous hands,
            and bimanual systems across different simulators. }
            \label{fig:teleop}
        \end{figure}

        \begin{figure}[bt]
            \centering
            \includegraphics[width=\linewidth]{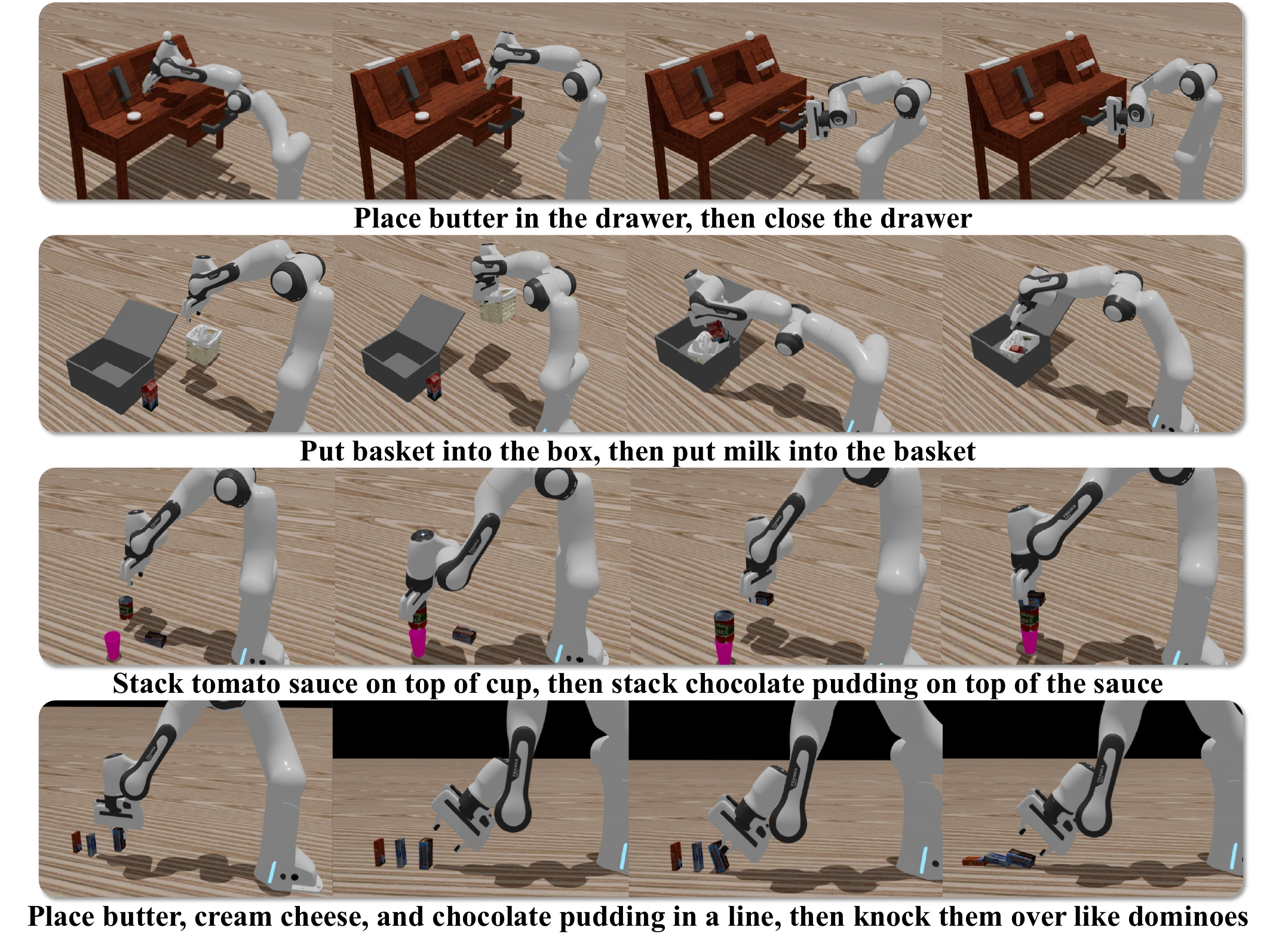}
            \vspace{-6mm}
            \caption{\textbf{AI-Assisted Task Generation.} \textsc{RoboVerse} supports an
            AI-assisted task generation framework that leverages large generative
            models' extrapolation capabilities to generate non-trivial and
            semantically rich tasks. Combined with our teleoperation system, it
            enables the generation of diverse and high-quality data.
            % By utilizing LLMs' extrapolation capabilities, our system not only understands spatial constraints and geometric feasibility but also generalizes across multiple simulators and benchmarks. This approach ensures robust task adaptation, allowing seamless integration of cross-domain tasks while maintaining consistency in object interaction, physics constraints, and manipulation strategies.
            }
            \label{fig:ai-assisted}
        \end{figure}

        \begin{figure}[bt]
            \centering
            \includegraphics[width=\linewidth]{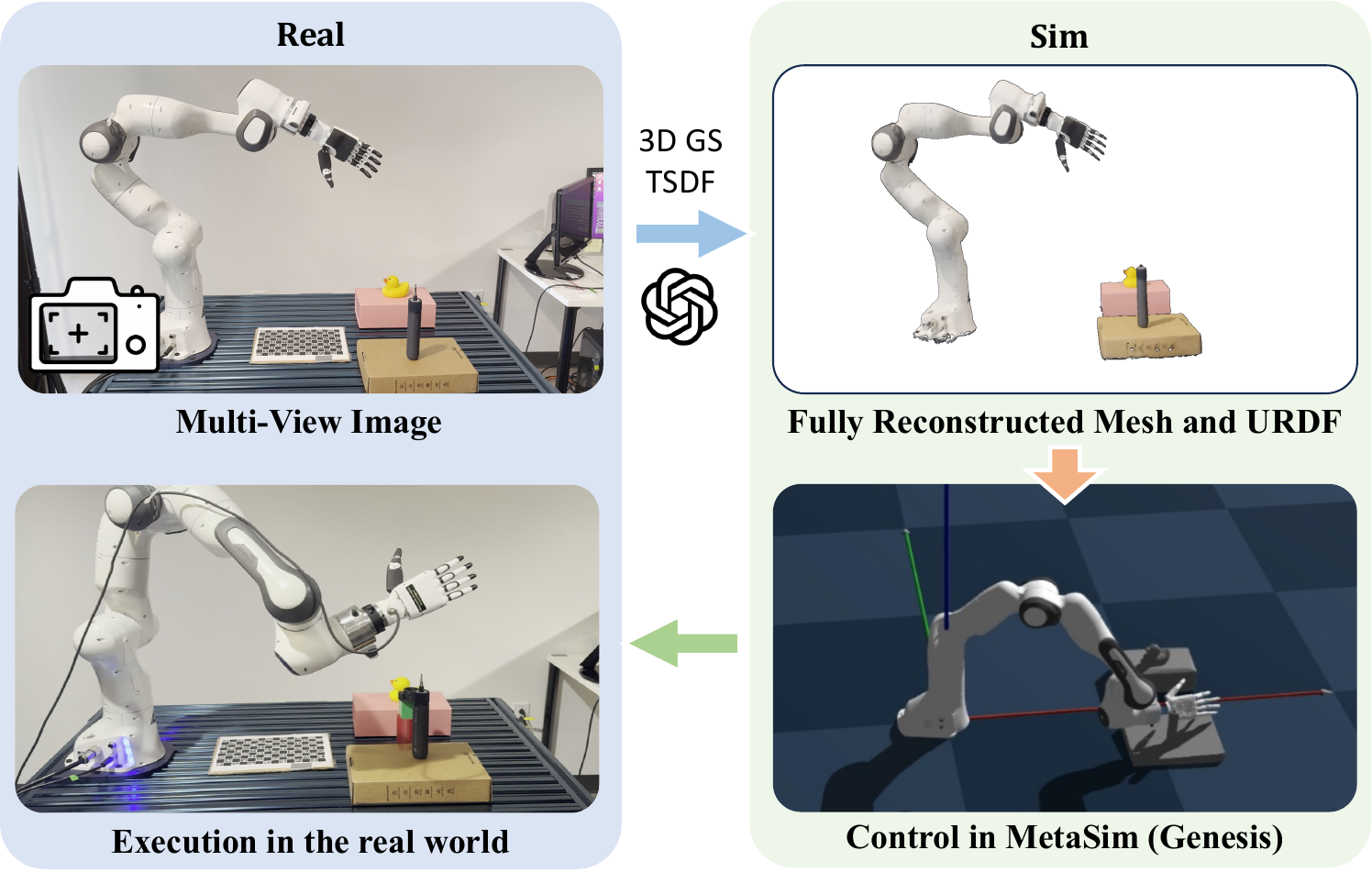}
            \caption{\textbf{Real-to-Sim Tools. }We use a mobile device to
            capture multi-view images, reconstruct a high-quality mesh, build a
            URDF using VLM, and then perform actions in both \textsc{RoboVerse} and the real
            world.}
            \label{fig:real2sim}
        \end{figure}
\end{itemize}

\subsection{Data Augmentation}
\label{sec:data_augmentation}

\subsubsection{Trajectory Augmentation}
\label{sec:data_traj_augmentation}

With the unified simulation interface and data format, \textsc{RoboVerse} enables
significantly more efficient data augmentation and supports advanced augmentation
techniques. Beyond the visual randomization detailed in Benchmark Protocol~\cite{calli2015benchmarking},
we also provide robust trajectory space augmentation.
% This approach has proven to be surprisingly effective in enhancing task generalization, further extending the applicability and robustness of trained models.
We offer an API to generate large-scale robot trajectory datasets from a limited
number of source demonstrations. Following the MimicGen~\cite{mandlekar2023mimicgen}
framework, for most tasks, we can decompose them into a sequence of object-centric
subtasks $(S_{1}(o_{S_1}), S_{2}(o_{S_2}), \dots, S_{M}(o_{S_M}))$, where the robot's
trajectory within each subtask $S_{i}(o_{S_i})$ is relative to a single object’s
coordinate frame ($o_{S_i}\in \mathcal{O}$, $\mathcal{O}$ is the set of objects in
the task $\mathcal{M}$). Additionally, we assume that the sequence of subtasks
in each task is predefined. By leveraging this minimal human annotation regarding
the order of subtasks, we can efficiently divide each source demo into
contiguous object-centric manipulation segments $\{\tau_{i}\}_{i=1}^{M}$ (each of
which corresponds to a subtask $S_{i}(o_{i})$) using a simulator, and then generate
extensive trajectory datasets for various task variants (in our case: variations
in the initial and goal state distributions of objects ($D$) and robots ($R$))
using MimicGen~\cite{mandlekar2023mimicgen}.
% \todo{maybe need some precise and objective word?}
This approach has been shown to significantly benefit generalization in imitation
learning~\cite{mandlekar2023mimicgen,jiang2024dexmimicgen,wang2024cyberdemo,garrett2024skillmimicgen,nasiriany2024robocasa},
particularly in scenarios where the number of source demonstrations is limited. For
further details, please refer to the supplementary materials.

\subsubsection{Domain Randomization}
\label{sec:data_domain_randomization}

We implement domain randomization in the Isaac Sim~\cite{IsaacSim} handler of \textsc{MetaSim}.
This involves four types of randomization:
\begin{itemize}[leftmargin=*]
    \item \textbf{Table, Ground, and Wall.} Walls (and ceilings) can be added
        for tasks that lack a predefined scene. Customizable tables can also be included
        for tasks that are performed on tabletops. The visual materials for
        these elements are randomly selected from a curated subset of ARNOLD~\cite{gong2023arnold}
        and vMaterials~\cite{vMaterials}. The table has ${\sim}$300 material
        options, while the wall and ground each have around ${\sim}$150 material
        options.

    \item \textbf{Lighting Condition.} Two types of lighting scenarios can be
        specified: distant light and cylinder light arrays. For distant light, the
        light's polar angles are randomized. %from $0$ degree to $45$ degree.
        For cylinder light, a random $n\times m$ matrix of cylinder lights with
        random size is added at a fixed height above the agents. In both scenarios,
        the intensity and color temperature of the lights are randomized within
        a reasonable range.

    \item \textbf{Camera Poses.} We carefully select 59 candidate camera poses, with
        the majority positioned to face the robot directly and a smaller subset placed
        at side-facing angles.

    \item \textbf{Reflection Properties.} The roughness, specular, and metallic
        properties of each surface are randomized within reasonable ranges.
\end{itemize}

These randomization options can be freely combined. For example, a scene can include
a customized table, walls with a ceiling, and a set of cylinder lights to simulate
an indoor environment. For details, please refer to the supplementary materials.

\subsection{\textsc{RoboVerse} Dataset}
\begin{table}[ht!]
    \Large
    \centering
    \caption{Migration progress statistics for manipulation tasks in \textsc{RoboVerse} }
    \label{tab:data_statistics}
    \begin{adjustbox}{width=\textwidth} % Ensure the table fits within page margins
        \begin{tabular}{l|c|ccc}
            \toprule
            Source Benchmark & \makecell{Source\\Simulator} & \makecell{\# Task\\Categories} & \# Trajectories & \# Assets \\
            \midrule
            ManiSkill~\cite{mu2021maniskill,gu2023maniskill2,tao2024maniskill3} & SAPIEN & 6 & 19k & 1.7k \\
            RLBench~\cite{james2019rlbench}& CoppeliaSim & 80 & 150k & 100 \\
            CALVIN~\cite{mees2022calvin} & Pybullet & 7 & 20k & 7 \\
            MetaWorld~\cite{yu2019metaworld}& MuJoCo & 5 & 5k & 6 \\
            RoboSuite~\cite{zhu2020robosuite}\&MimicGen~\cite{mandlekar2023mimicgen}& MuJoCo  & 6 & 6k &12\\
            GAPartNet~\cite{geng2023gapartnet} & IsaacGym & 4 & 4k & 151 \\
            Open6DOR~\cite{ding2024open6dor}& IsaacGym & 69 & 10k & 207 \\
            ARNOLD~\cite{gong2023arnold}& IsaacSim & 6 & 3k & 30 \\
            LIBERO~\cite{liu2023libero}& MuJoCo  & 10 & 15k & 15\\
            Simpler~\cite{li24simpler}& SAPIEN  & 6 & 30k & 52 \\
            RLAfford~\cite{geng2023rlafford} & IsaacGym & 4 & 40k & 40\\
            GraspNet~\cite{fang2020graspnet} & - & 58 & 200k & 42 \\
            GarmentLab~\cite{lu2024garmentlab} & IsaacSim & 6 & 6k & 3k \\
            UniDoorManip~\cite{li2024unidoormanip} & IsaacGym & 7 & 1k & 140 \\
            GAPartManip~\cite{cui2024gapartmanip} & IsaacSim & 2 & 1.5k & 42 \\
            \midrule
            % DexterousHands~\cite{chen2022towards} & - &  & & \\
            % \midrule
            % Navigation-R2R~\cite{Krantz2020BeyondTN}\&RxR~\cite{ku2020room} & IsaacSim & 2 & 30k & 90 \\
            % \midrule
            % HumanoidBench~\cite{sferrazza2024humanoidbench} & - &  & & \\
            % Humanoid-X~\cite{uh1} & - &  & & \\
            % \midrule
            Total & - & 276 & 510.5k & 5.5k \\
            \bottomrule
        \end{tabular}
    \end{adjustbox}
\end{table}
\begin{figure*}[hbt]
    \centering
    \includegraphics[width=\linewidth]{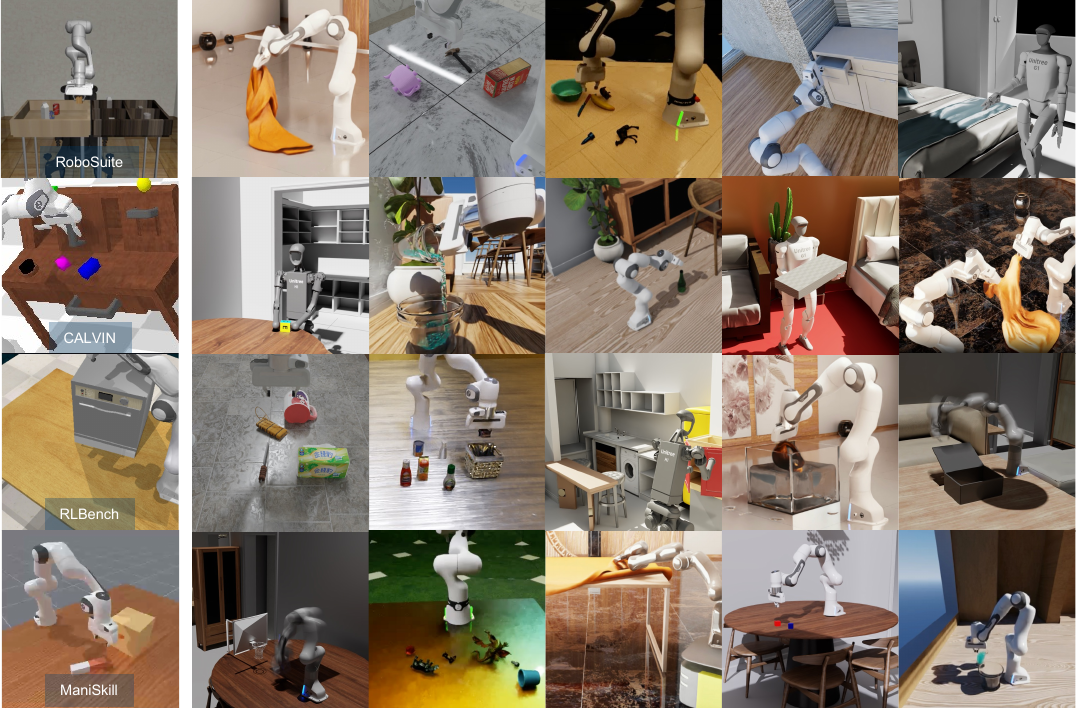}
    \caption{\textbf{Dataset Comparison and Gallery}. Left: other representative
    synthetic robotics datasets. Right: the \textsc{RoboVerse} dataset.}
    \label{fig:gallery}
\end{figure*}
\subsubsection{Dataset Statistics}

\paragraph{Manipulation Dataset}

We migrate diverse manipulation datasets from existing source benchmarks~\cite{mu2021maniskill,gu2023maniskill2,tao2024maniskill3,james2019rlbench,mees2022calvin,yu2019metaworld,zhu2020robosuite,mandlekar2023mimicgen,geng2023gapartnet,ding2024open6dor,gong2023arnold,liu2023libero,li24simpler,geng2023rlafford,fang2020graspnet,lu2024garmentlab,li2024unidoormanip,cui2024gapartmanip}
into \textsc{RoboVerse}. The number of task categories, trajectories and assets contributed
by each source benchmarks is summarized in \tref{tab:data_statistics}. In total,
this migration results in 276 task categories, 510.5k trajectories, and 5.5k
assets. Representitive tasks with rich domain randomization are shown in \fref{fig:gallery}.

\paragraph{Navigation Dataset}
% \haoran{jiaozhao, please add details here}

We migrate vision-and-language navigation (VLN) tasks into \textsc{RoboVerse}.
Note that there exists various VLN tasks with different settings; here, we
particularly focus on VLN in continuous environments (VLN-CE)~\cite{Krantz2020BeyondTN},
as it more closely resembles real-world scenarios~\cite{cheng2024navila,zhang2024uni,zhang2024navid}.
Specifically, we construct our dataset based on \textsc{RoboVerse} by
integrating MatterPort 3D scenes~\cite{chang2017matterport3d} (90 scenes) and
off-the-shelf instructions from R2R~\cite{Krantz2020BeyondTN} (10k episodes) and
RxR~\cite{ku2020room} (20k episodes). We provide two types of mobile embodiments,
including the Unitree Dog (a legged robot) and the JetBot (a wheeled robot), which
support different control policies. A detailed elaboration on the navigation
dataset is provided in the supplementary materials.

\paragraph{Humanoid Dataset}
We migrate HumanoidBench~\cite{sferrazza2024humanoidbench} tasks for reinforcement
learning benchmarks and integrate tasks, policies, and data samples from Humanoid-X~\cite{uh1} and SkillBlender~\cite{kuang2024skillblender}. Additionally, we re-implement the UH-1 inference pipeline within our
framework. The pretrained policy successfully enables humanoid robots to follow
demonstrated poses while maintaining stable locomotion across multiple
simulators based on \textsc{RoboVerse}.

\section{\textsc{RoboVerse} Benchmark}
\begin{figure*}[ht]
    \centering
    \includegraphics[width=\linewidth]{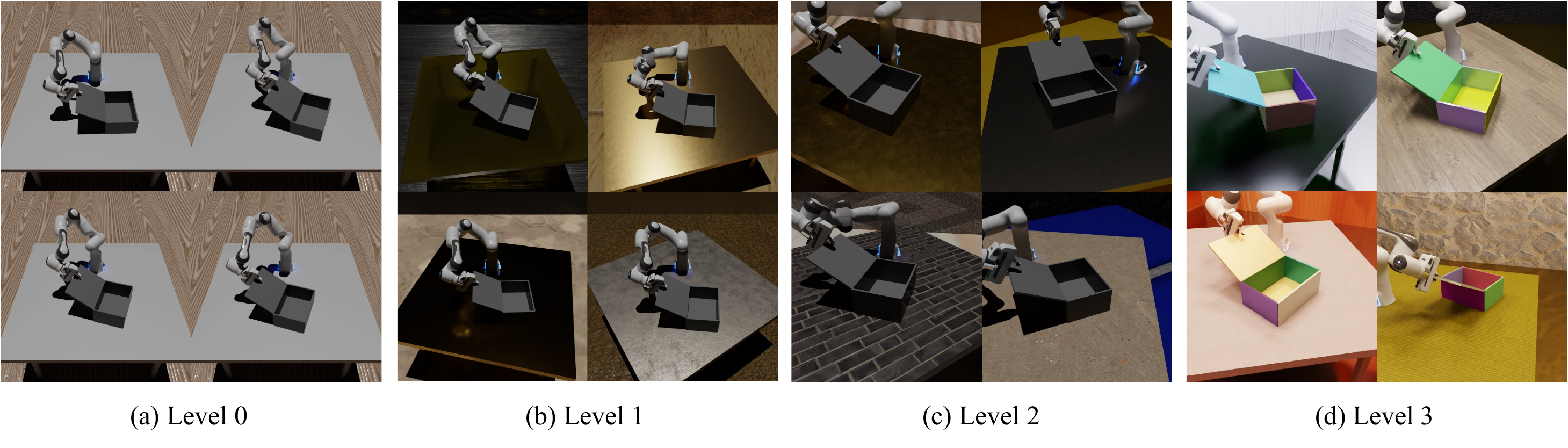}
    \caption{\textbf{Benchmark Protocol:} We define a four-level generalization benchmarking
    protocol, allocating 90\% of the data for training and 10\% for
    generalization evaluation. From left to right, Levels $0$ to $3$ corresponds
    to task space generalization, environment radomization, camera randomization,
    lighting and reflection randomization, respectively. }
    \label{fig:bench_levels}
\end{figure*}

\subsection{Benchmark Overview}
With the collected tasks, assets, and trajectories, \textsc{RoboVerse} establishes standardized
benchmarks for robot learning, including both imitation learning and
reinforcement learning. We define a unified training and evaluation protocol within
the \textsc{RoboVerse} platform and implement standardized baselines and learning frameworks
for benchmarking. Specifically, for imitation learning, we introduce different
levels of generalization benchmarks to assess the generalization capability of
models.

\subsection{Imitation Learning Benchmark}

\label{sec:il_benchmark} For each imitation learning benchmark, we establish a
standardized evaluation framework with a fixed set of demonstrations and a
controlled evaluation environment. Policies must be trained exclusively on the provided
training data and assessed within this environment to ensure fair comparison. To
rigorously test generalization capability, we curate training data from specific
domains and evaluate policies on unseen samples, challenging their adaptability
to novel scenarios. We systematically categorize visual generalization factors into
multiple levels, including task space generalization, environment setup generalization,
camera setting generalization, and lighting and reflection generalization. Each
level introduces controlled variations to assess a policy’s adaptability and
robustness in increasingly diverse and challenging conditions.

\paragraph{Level 0: Task Space Generalization}
We establish a controlled evaluation by standardizing the environment with consistent
camera, materials, lighting, and other parameters. The task space, including
object initialization and instructions, is split into 90\% training and 10\% validation
to assess generalization within a fixed setting, as shown in \fref{fig:bench_levels}
(a).

\paragraph{Level 1: Environment Randomization}
Building on the standardized setup, we introduce scene randomization while keeping
the camera, materials, and lighting fixed~\cite{martinez2020unrealrox}. By
varying house, table, and ground configurations, we create diverse visual inputs
to test robustness against environmental changes~\cite{kadian2020sim2real}. A
fixed set of predefined randomized scenes ensures structured evaluation, as shown
in \fref{fig:bench_levels} (b).

\paragraph{Level 2: Camera Randomization}
To assess generalization across camera variations, we introduce different
viewing heights and angles using carefully annotated, realistic camera poses.
Following the 90/10 training/testing split, we ensure consistent and rigorous evaluation,
as illustrated in \fref{fig:bench_levels} (c).

\paragraph{Level 3: Lighting and Reflection Randomization}
Real-world environments involve diverse materials and lighting conditions~\cite{vaccaro2013robotic}.
To simulate these challenges, we randomize lighting and reflections, curating realistic
object materials and illumination setups~\cite{dai2022domain}. This enhances
robustness testing under varying conditions, as shown in \fref{fig:bench_levels}
(d).

\subsection{Reinforcement Learning Benchmark}

In addition to imitation learning, \textsc{RoboVerse} offers a comprehensive
reinforcement learning (RL) benchmark designed to accommodate a diverse range of
tasks, robot embodiments, and simulation backends. Specifically, we integrate
the PPO~\cite{schulman2017proximal} algorithm from both Stable-Baselines3~\cite{raffin2021stable}
and rsl\_rl~\cite{rudin2022learningwalkminutesusing} into our \textsc{MetaSim}
interface, enabling straightforward task definition, seamless environment switching,
and standardized performance logging.

Building upon this infrastructure, we have successfully ported multiple humanoid
control tasks from the HumanoidBench~\cite{sferrazza2024humanoidbench} benchmark
into \textsc{RoboVerse}. Through our adapted interface for rsl\_rl~\cite{rudin2022learningwalkminutesusing}, we have
efficiently extended framework compatibility to support the TD-MPC2~\cite{hansen2022tdmpc,
hansen2024tdmpc2}
algorithm from the original benchmark while preserving implementation fidelity.

\section{Experimental Results}

% \koushil{Somewhere in this section (or elsewhere), I would have liked to see the following:

% (a) computation performance overhead of using Roboverse instead of an underlying simulator.  For eg.: How much extra time would be spent to simulate MuJoCo through Roboverse vs simulating directly in MuJoCo?

% (b) While using RL, is it possible to use Roboverse to randomly sample a simulator?  This way you do domain randomization by just changing simulators...  If so, do mention this.}

\subsection{Overview}
We conduct extensive experiments to validate the effectiveness and practicality of \textsc{RoboVerse}. First, we evaluate baselines on representative tasks from various benchmark sources to ensure the reliability of the collected datasets and established benchmarks. This includes assessments of both imitation learning baselines \sref{sec:il_exp} and reinforcement learning baselines \sref{sec:rl_exp}.

Then we further demonstrate the strength of the high-quality synthetic dataset. We find that synthetic data could significantly boost world model learning. 

\subsection{Results on the Imitation Learning Benchmark}
\label{sec:il_exp}

\begin{table*}[ht!]
    \centering
    \caption{\textbf{Baseline Results on \textsc{RoboVerse} Imitation Learning Benchmark.} 
    We report baseline results on representative tasks from various benchmark sources to validate the effectiveness and reliability of the \textsc{RoboVerse} benchmark.
    % We report the success rate xxx\todo{stats!}. We \textbf{bold} the best result for models within the same category and \underline{underline} the second. 
}
    \resizebox{\textwidth}{!}{
    \begin{tabular}{c|c|cccccc|c}
    \toprule
    \multicolumn{2}{c|}{Representative Task}
     &PickCube & StackCube & CloseBox & MoveSliderLeft & PickChocolatePudding & NutAssembly &  Average \\
    % \midrule
    \multicolumn{2}{c|}{Benchmark Source}
    & ManiSkill  & ManiSkill & RLBench &  CALVIN & LIBERO & RoboSuite & - \\
    \midrule
    Diffusion Policy~\cite{chi2023diffusionpolicy} & 78M & 52.7 & 53.8 & 51.5 &76.5 & 50.0 & 7.1 & 48.6 \\
    ACT~\cite{zhao2023learning} & 84M & 31.7 & 36.7 & 68.3 &  85.0 & 78.3 & 0.0 & 50.0\\
    % \midrule
    % OpenVLA**~\cite{kim24openvla} & 7B & 40.0 & - & - &  45.0 & - & -  & - \\
    
    \bottomrule
    \end{tabular}
    }
    \label{table:benchmark}
\end{table*}

\begin{table*}[ht!]
    \centering
    \caption{\textbf{Generalization Performance on Imitation Learning Benchmark.} This table presents the experimental results for each generalization level in our benchmark across different tasks and methodologies. The tasks are divided into distinct levels (Level 0, Level 1, Level 2, and Level 3) to evaluate performance under progressively challenging scenarios. }
    \resizebox{\textwidth}{!}{
    \begin{tabular}{c|cccc|cccc|cccc}
    \toprule
    \multirow{2}{*}{Task and Generalization Level} & \multicolumn{4}{c|}{MoveSliderLeft} & \multicolumn{4}{c|}{CloseBox}& \multicolumn{4}{c}{PickCube} \\
    & Level 0 & Level 1 & Level 2 & Level 3 & Level 0 & Level 1 & Level 2 & Level 3& Level 0 & Level 1 & Level 2 & Level 3 \\
    \midrule
    Diffusion Policy~\cite{chi2023diffusionpolicy} & 76.5 & 81.3 & 72.0 & 60.0 & 51.5 & 42.8 & 20.0 & 10.4 &52.7&11.1&0.0&0.0\\
    ACT~\cite{zhao2023learning} & 85.0 & 83.3 & 43.3 & 16.6 & 68.3 & 73.3 & 0.0 & 20.0 & 31.7 & 30.0 & 6.7 & 3.3\\
    \midrule
    % Octo \cite{octo_2023} &  &  &  &  &  &  &  &  \\
    % RDT \cite{liu2024rdt} &  &  &  &  &  &  &  &  \\
    OpenVLA\footnotemark[1]~\cite{kim24openvla} & 45.0 & 40.0 & 35.0 & 30.0 & 0.0 & 0.0 & 0.0 & 0.0 & 40.0 & 15.0 & 0.0 & 0.0 \\
    \bottomrule
    \end{tabular}
    }
    \label{table:generalization_benchmark}
\end{table*}

\footnotetext[1]{Due to resource and time constraints, we uniformly sample 20 testing scenarios for the OpenVLA baseline.}
\begin{table}[bt]
    \huge
    \centering
    \begin{adjustbox}{width=\textwidth}
        \begin{tabular}{c|cc|ccc}
            \toprule
            \multirow{2}{*}{Method} & \multicolumn{2}{c|}{Simple} & \multicolumn{3}{c}{Language-conditioned Grasping} \\ 
           &  PickCube & MoveSliderLeft & Object Set 1 & Object Set 2 & Object Set 3 \\
            \midrule
            OpenVLA~\cite{kim24openvla} & 40.0 & 45.0 & 46.0 & 33.3 & 14.4\\
            Octo~\cite{octo_2023} & 50.0 & 30.0  & 42.0 & 14.4 & 2.2 \\
             \bottomrule
        \end{tabular}
    \end{adjustbox}
    \caption{\textbf{Vision-Language-Action (VLA) Model Results on \textsc{RoboVerse} Imitation Learning Benchmark.} Constrained with time and resources, we report VLA models' results on two simple tasks from \textsc{RoboVerse} and grasping tasks with diverse and challenging language instructions. We split 58 objects in GraspNet into three sets, each containing progressively more challenging objects based on their geometry.}
    \label{table:vla}
\end{table}
\subsubsection{Baseline and Task Selection}
To genuinely reflect the data quality of the \textsc{RoboVerse} dataset and provide a standard benchmark for all kinds of imitation learning policy models, we select both prevailing specialist and generalist models as baselines of our \textsc{RoboVerse} benchmark. Specifically, for specialist models, we integrate ACT~\cite{zhao2023learning} and Diffusion Policy~\cite{chi2023diffusionpolicy}. For generalist models, We benchmark our approach on OpenVLA~\cite{kim24openvla} and Octo~\cite{octo_2023}, both of which we fine-tuned using our synthetic dataset. ACT is one of the most widely used methods in bi-manual manipulation. Diffusion Policy~\cite{chi2023diffusionpolicy} is the first work that applies the conditional denoising diffusion process as a robot visuomotor policy and achieves great generalization capabilities. % OpenVLA is the largest open-source vision-language-action model with 7B parameters.

Leveraging the \textsc{RoboVerse} format and infrastructure design, we are able to evaluate models on different tasks within a unified platform. To fully test policy models' performance under versatile settings, we select one representative task from each of the source benchmarks integrated by the \textsc{RoboVerse} dataset as shown in \tref{table:benchmark}. The experiment subset includes \texttt{PickCube} and \texttt{StackCube} from ManiSkill~\cite{mu2021maniskill}, \texttt{CloseBox} from RLBench~\cite{james2019rlbench}, \texttt{MoveSliderLeft} from CALVIN~\cite{mees2022calvin}, \texttt{PickChocolatePudding} from LIBERO~\cite{liu2023libero}, and \texttt{NutAssembly} from \texttt{robosuite}~\cite{zhu2020robosuite}. These tasks not only demand precise pick-and-place skills but also require contact-rich physical interactions with articulated objects. Through these tasks, the benchmark results can provide a comprehensive reflection of each model's performance under different scenarios.

\subsubsection{Implementation Details}
Due to time and resource constraints, we implement specialist and generalist models using different strategies, and all the results are obtained under the single-task setting. The training and evaluation settings follow the $90/10$ \textsc{RoboVerse} benchmark protocol as specified in \sref{sec:il_benchmark}. During evaluations, we randomly select ten task settings from training sets and another ten from the validation sets. The reported success rates are computed as the averages over three random seeds. 

For each step, the inputs are $256 \times 256 \times 3$ RGB images and a short language description depending on the task settings. 
For specialist models, we train from scratch with action in $9$-dim robot joint state space. 
For generalist models, the action is pre-processed into delta end-effector position space from absolute end-effector position space, and The gripper action is discretized into binary values $\{0, +1\}$. Owing to the lack of time and resources, we are only able to fine-tune the generalist models in the single-task setting. During evaluations, we employ cuRobo~\cite{curobo_report23} as the inverse-kinematics solver to transform the action to robot joint state space. Specific model implementation details and hyperparameters are provided in supplementary materials.

\subsubsection{Experiment Results}
We present the imitation learning benchmark results in \tref{table:benchmark} and the generalization evaluation in \tref{table:generalization_benchmark}. We further fine-tune large vision-language-action models on both simple and complex language-conditioned tasks, as shown in ~\tref{table:vla}.

\subsection{Results on the Reinforcement Learning Benchmark}
\label{sec:rl_exp}
Using Stable-Baselines3~\cite{raffin2021stable} and rsl\_rl~\cite{rudin2022learningwalkminutesusing} implementations of PPO, we train policies on tasks from IsaacLab~\cite{mittal2023orbit} under consistent hyperparameters. 

For additional tasks (humanoid, dexterous hand), the same PPO-based workflow applies. We successfully migrate the HumanoidBench~\cite{sferrazza2024humanoidbench} from MuJoCo to \textsc{RoboVerse}, enabling training across multiple simulators (Isaac Sim and MuJoCo) with consistent interfaces. Experiment results demonstrate stable policy convergence across simulators, achieving comparable performance to native MuJoCo baselines. Leveraging the generalizability of rsl\_rl~\cite{rudin2022learningwalkminutesusing}, we further extend the benchmark to support TD-MPC2~\cite{hansen2022tdmpc, hansen2024tdmpc2} algorithm \koushil{This was already mentioned right at the end of Section III  ??  Why repeat here?}, which exhibits robust training dynamics in all environments. For implementation details, reward curve, and extended experimental results, please refer to the supplementary materials.

\subsection{Augmentation Experiments}
\begin{figure}[hbt] \centering
    \includegraphics[width=1.0\linewidth]{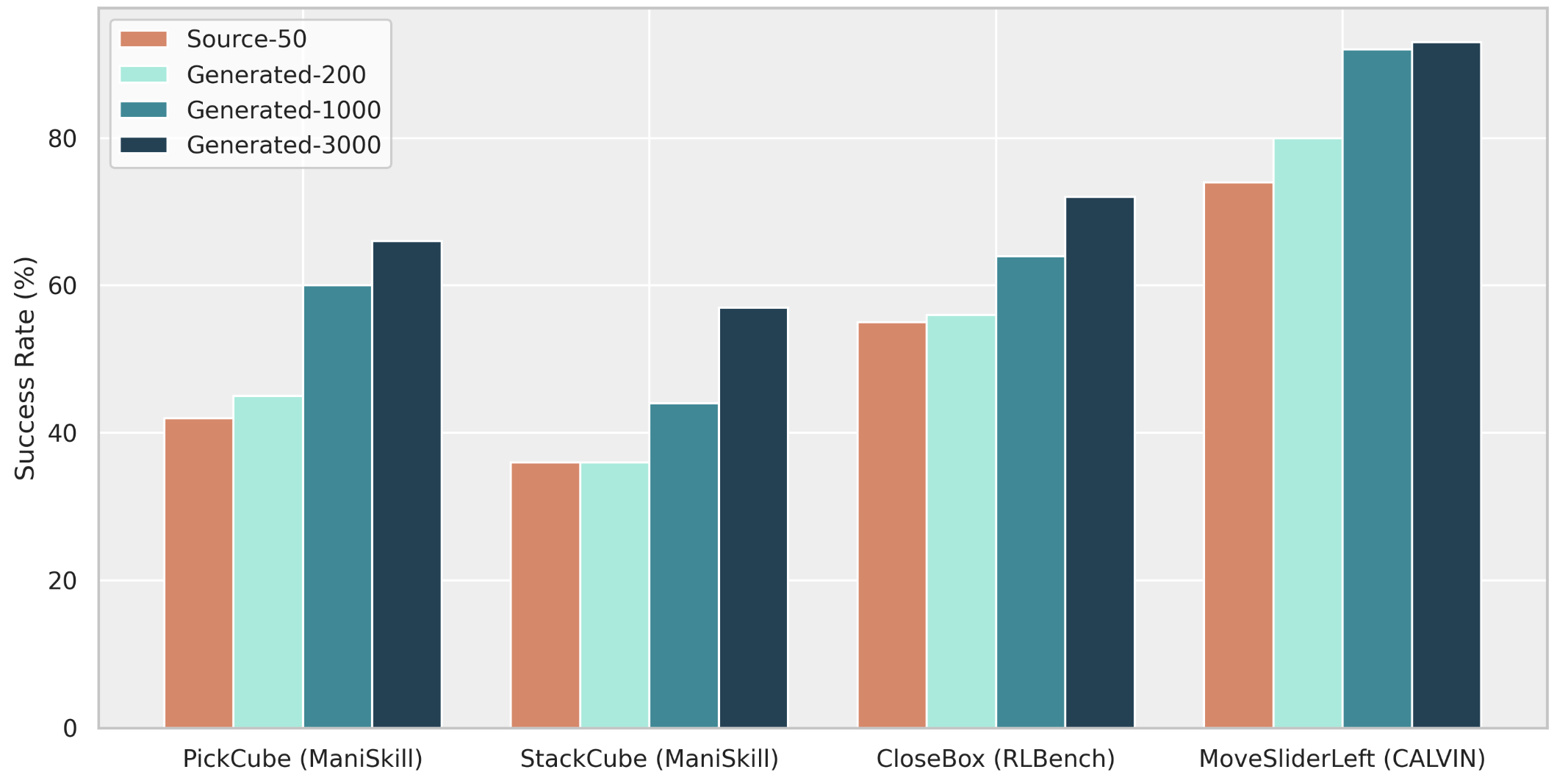}
    \caption{\textbf{Effectiveness of Trajectory Augmentation.} Success rates of policy trained with augmented dataset and source dataset.
    }
    \label{fig:data_aug}
\end{figure}

To verify the effectiveness of our trajectory augmentation API, on four representative tasks, we compare the success rates of trained Diffusion Policy on 50 source demonstrations and 200, 1000, and 3000 generated augmentation demonstrations under the imitation learning setting. The results presented in ~\fref{fig:data_aug} demonstrate a consistent improvement in model performance as the number of generated data increases, highlighting both the effectiveness and scalability of the trajectory augmentation API. 

\subsection{World Model Learning}
\label{sec:world_model_exp}

\begin{figure}[tb]
    \centering
    \includegraphics[width=\linewidth]{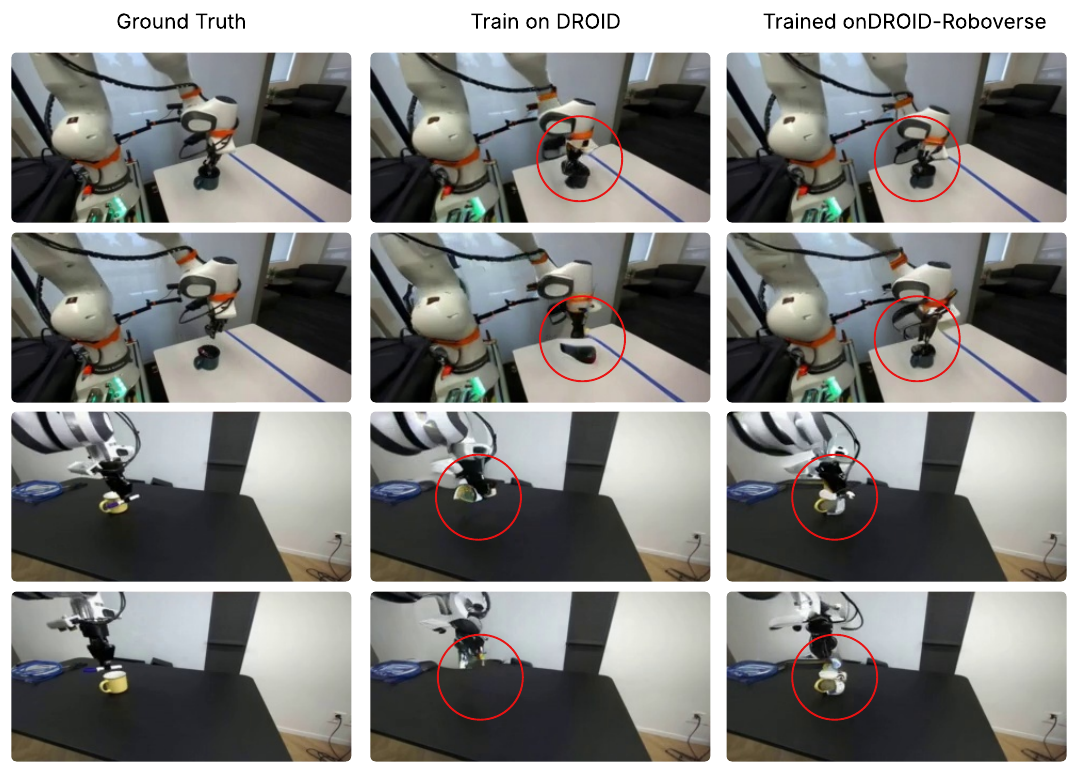}
    \caption{\textbf{Ablation Study of Action-conditioned World Model Learning.} 
    We compare the qualitative results of an action-conditioned world model trained on pure DROID and DROID-RoboVerse datasets, with evaluations sampled from the DROID dataset.}
    \label{fig:world_model_droid_vs_droid-roboverse}
\end{figure}

\vspace{10pt}

Recent advances in general-purpose video generation and interactive world models~\cite{polyak2024moviegencastmedia, bruce2024geniegenerativeinteractiveenvironments} have shown promising progress. Yet, the scarcity of gigantic-scale robotic datasets still impedes the development of robust world models for a wide range of robotic applications. In this session, we demonstrate how synthetic data from the \textsc{RoboVerse} simulation can augment real-world datasets to train more capable robotics world models.

When a model is trained exclusively on 50,000 episodes from the DROID dataset~\cite{khazatsky2024droid}, it generally respects action conditions but struggles to accurately capture physical interactions between the gripper and target objects. Notably, the objects appear “warped” during contact with the gripper, as shown in \fref{fig:world_model_droid_vs_droid-roboverse}. By incorporating an additional 50,000 synthetic episodes from \textsc{RoboVerse} to create a combined dataset of  100,000 episodes, the model predictions improve with regard to preserving object geometry.
% (figure 9). 
However, merely “watching videos” remains insufficient for learning the intricate physical interactions in DROID.

In contrast, training solely on the \textsc{RoboVerse}-50K or on the DROID-RoboVerse-100K dataset and then validating on \textsc{RoboVerse} samples, we observe that the generated frames are physically more realistic in most scenes, with details in the supplementary materials. This improvement can be attributed to the extensive randomization and augmentation available in \textsc{RoboVerse}. Conversely, a model trained solely on DROID data fails to transfer effectively to the \textsc{RoboVerse} scene. We hypothesize that this shortcoming stems from limited samples per scene coverage in DROID and incomplete gripper visibility in the camera view.

\subsection{Imitating the \textsc{RoboVerse} Dataset Enables Direct Sim-to-Real Transfer}
\label{sec:exp_sim2real}

% \begin{figure*}[tb]
%     \centering
%     \includegraphics[width=\linewidth]{fig/humanoid_failure.png}
%     \caption{\textbf{Simulation Comparison and Failures of Direct Sim-to-Real.} Training a locomotion policy directly in current parallel simulation, \eg, Isaac Gym~\cite{makoviychuk2021isaacgym}, and deploying it to the real world often results in significant failures. Similar failures are observed in sim-to-sim transfers. We observe that MuJoCo~\cite{todorov2012mujoco} demonstrates higher fidelity to real-world dynamics in this context.}
%     \label{fig:huamnoid_failure}
% \end{figure*}

\begin{figure*}[tb]
    \centering
    \includegraphics[width=\linewidth]{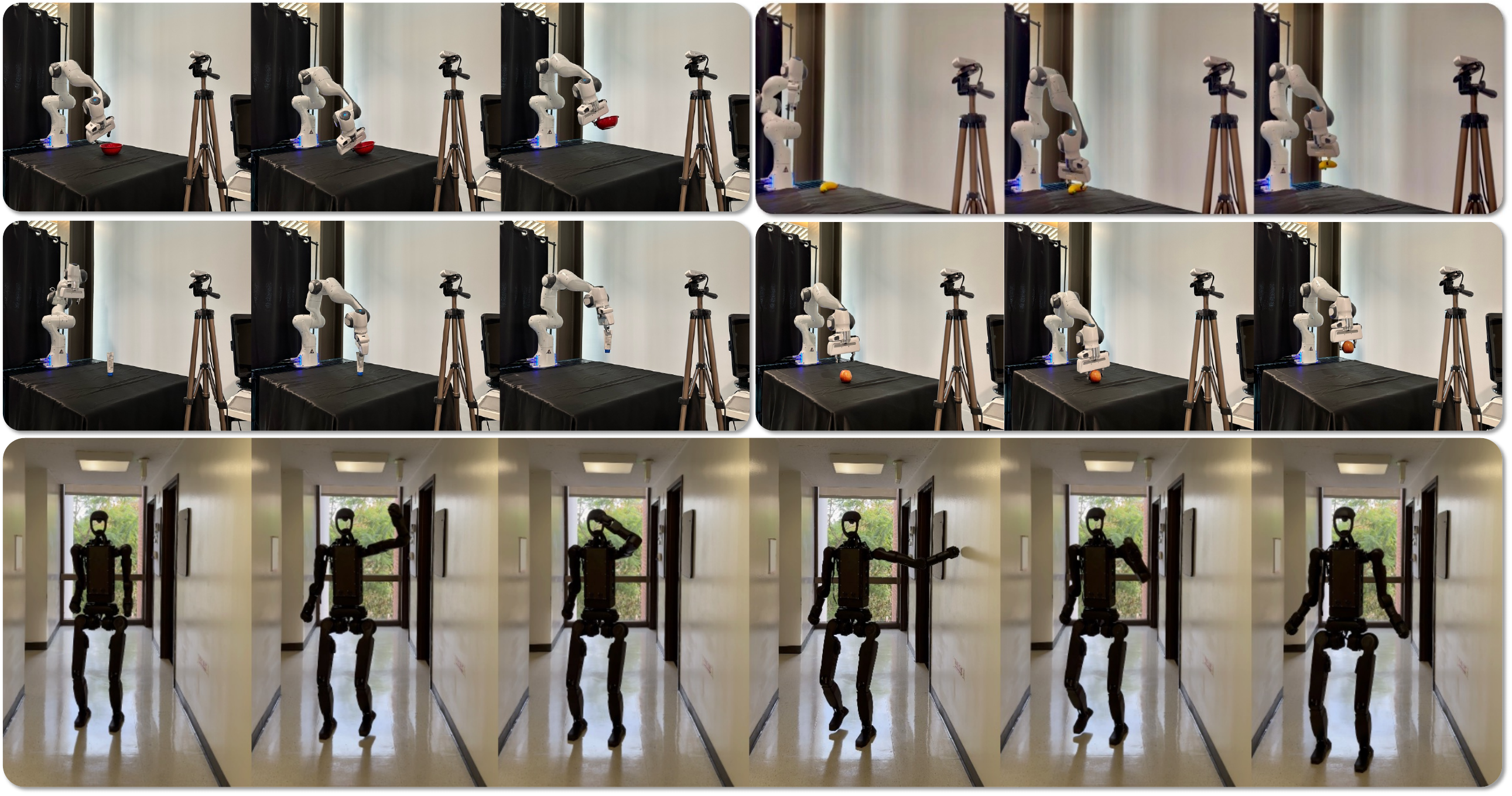}
    \caption{\haoran{add more} \textbf{Sim-to-Real and Sim-to-Sim-to-Real Experiment Results.} We demonstrate that learning within the \textsc{RoboVerse} framework enables seamless direct Sim-to-Real transfer for manipulating unseen objects in new environments (imitation learning) and Sim-to-Sim-to-Real transfer for whole-body humanoid control (reinforcement learning).}
    \label{fig:sim2real}
\end{figure*}

\begin{figure}[tb]
    \centering
    \includegraphics[width=\linewidth]{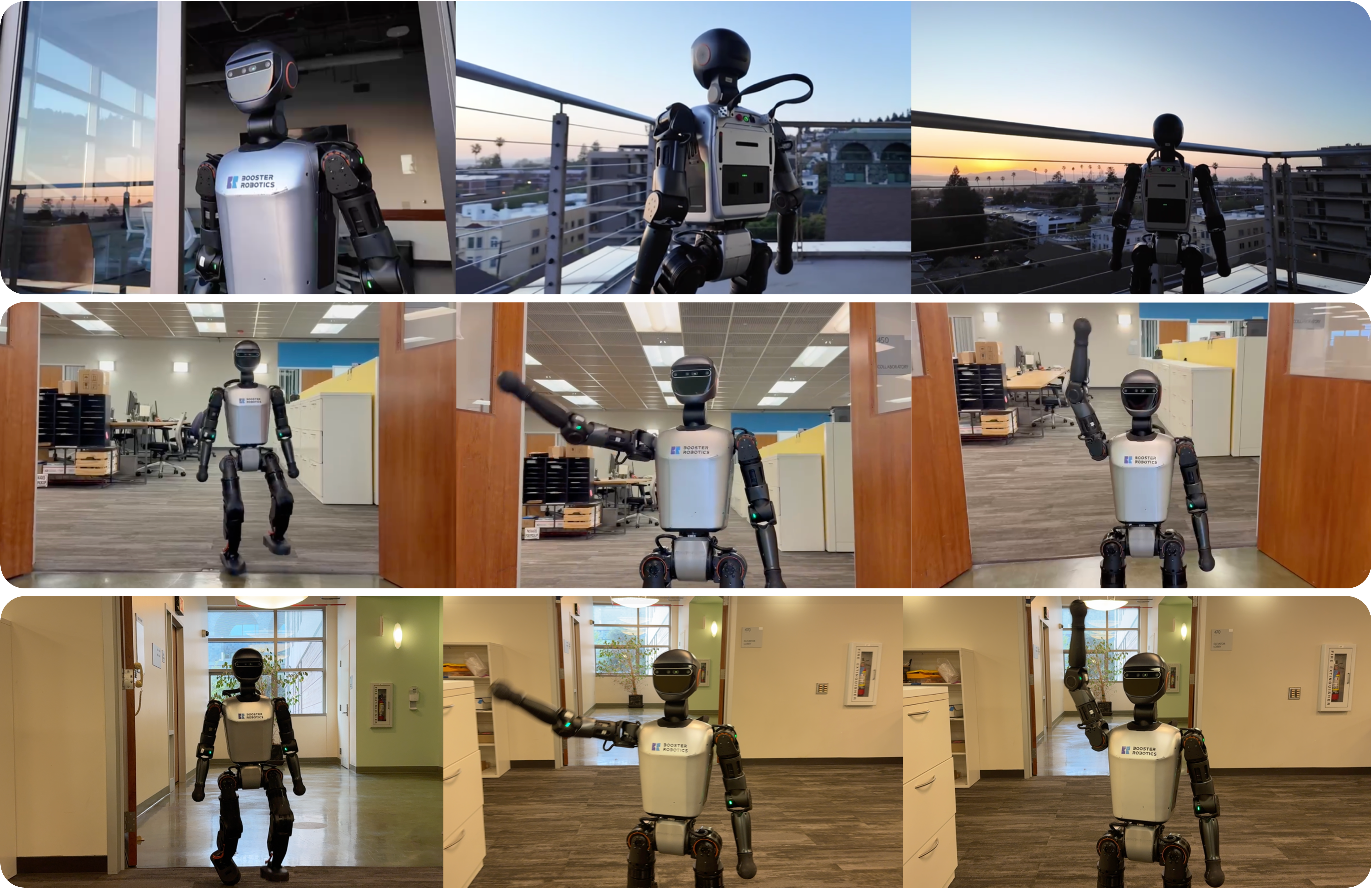}
    \caption{\textbf{Generalization of Sim-to-Sim-to-Real.} This figure shows the in-the-wild generalization ability of our lower-body RL policy with upper-body PD control by the sim-to-sim-to-real approach.}
    \label{fig:sim2sim2real}
\end{figure}

 The \textsc{RoboVerse} system seamlessly integrates a powerful physics engine with a high-quality renderer, ensuring the generation of realistic, high-fidelity data. To demonstrate its potential, we conduct experiments validating its effectiveness in direct sim-to-real transfer. 
 As shown in \fref{fig:sim2real}, we fine-tune OpenVLA~\cite{kim24openvla} on the \textsc{RoboVerse} dataset and transfer the learned policy to real-world scenarios without additional fine-tuning. The model successfully manipulates unseen objects in previously unseen real-world environments, showcasing the robustness and generalization capabilities of our system. The quantitative results on more challenging language-guided tasks, as shown in ~\tref{tab:direct_sim2real}, further demonstrate the high success rate of models trained on the \textsc{RoboVerse} dataset. Additional details are provided in the supplementary materials.
 \begin{table}[tb]
\centering
\caption{\textbf{Direct Sim-to-Real.} We fine-tune two baseline models using demonstrations adapted from GraspNet~\cite{fang2020graspnet} to validate the effectiveness of the RoboVerse dataset. The final performance score for each task is reported, where a baseline receives 1 point for successfully grasping the target. Additionally, we adopt the partial reward scheme from OpenVLA~\cite{kim24openvla}, awarding 0.5 points when the gripper makes contact with the target.}

\resizebox{\textwidth}{!}{
\begin{tabular}{c|ccc}
\toprule
\makecell{GraspNet Objects}  & \makecell{Pick up \\ Wash Soap} & \makecell{Lift \\Mouth Rinse} & \makecell{Grasp \\ Green Dish}\\
\midrule
Octo~\cite{octo_2023} &  5.0/10.0 & 3.0/10.0 & 6.0/10.0 \\
OpenVLA~\cite{kim24openvla} & 7.0/10.0 & 8.0/10.0 & 5.0/10.0  \\
\bottomrule
\end{tabular}}

\label{tab:direct_sim2real}
\end{table}

\subsection{Reinforcement Learning in \textsc{RoboVerse} Enables Sim-to-Sim-to-Real Transfer}
\label{sec:exp_sim2sim2real}
Large-scale parallel environments offer significant potential for large-scale exploration and are highly effective for reinforcement learning (RL) tasks. However, while they provide excellent efficiency, their accuracy may be limited in certain scenarios~\cite{dulacarnold2019challengesrealworldreinforcementlearning}. % We also demonstrated some failure cases directly from sim-to-real transfer (see ~\fref{fig:huamnoid_failure}). 
To address this problem, Sim-to-sim evaluation and fine-tuning present promising solutions~\cite{li24simpler}. As shown in \fref{fig:sim2sim2real}, \textsc{RoboVerse} platform seamlessly supports such functionalities, enabling robust sim-to-sim and sim-to-real transitions. We further demonstrate the effectiveness of sim-to-sim-to-real generalization through comprehensive experiments, highlighting the platform's ability to bridge simulation and real-world performance.

% \section{Conclusion}

\section{Limitations}

\koushil{In addition to this, you should have a section on Conclusion.}

While \textsc{RoboVerse} provides a comprehensive and scalable platform, several limitations remain. First, the integration of a unified format for non-rigid objects is not yet fully supported, which we leave for future work to develop. Additionally, while our large-scale dataset presents significant potential for pretraining a foundation model, this exploration falls beyond the scope of this paper due to resource constraints.
Furthermore, despite our extensive efforts to fully reimplement and optimize all baseline methods within the \textsc{RoboVerse} baselines, some implementations may still be suboptimal. Our primary goal is not to directly compare policy performance but to demonstrate that the system is comprehensive, supports diverse policies, and ensures strong alignment between simulation and real-world performance. While we have made every effort to build a robust platform, it is inevitable that some oversights or errors may remain. We encourage the broader research community to contribute to maintaining and refining the baselines, fostering collaboration to further enhance the platform's capabilities.

%%% Acknowledgement
\section*{Acknowledgement}
We thank Hanyang Zhou and Sicheng He for providing valuable suggestions for setting up robotics hardware. We thank Yufeng Chi and Sophia Shao for providing humanoid robots for testing. We thank Jie Yang and Muzhi Han for valuable discussion. We thank Koushil Sreenath for insightful feedback. We thank Jiawei Yang, Sumeet Batra, and Gaurav Sukhatme for their generous help.
Pieter Abbeel holds concurrent appointments as a professor at UC Berkeley and as an Amazon Scholar. This paper describes work performed at UC Berkeley and is not associated with Amazon.

\koushil{Many of the references cite arXiv versions - if these papers are published, please cite the published versions instead - check this for all references here...}

\newpage

\bibliographystyle{plainnat}
\bibliography{ref}

\begin{thebibliography}{145}
\providecommand{\natexlab}[1]{#1}
\providecommand{\url}[1]{\texttt{#1}}
\expandafter\ifx\csname urlstyle\endcsname\relax
  \providecommand{\doi}[1]{doi: #1}\else
  \providecommand{\doi}{doi: \begingroup \urlstyle{rm}\Url}\fi

\bibitem[Antero et~al.(2024)Antero, Blanco, O{\~n}ativia, Sall{\'e}, and Sierra]{antero2024harnessing}
Unai Antero, Francisco Blanco, Jon O{\~n}ativia, Damien Sall{\'e}, and Basilio Sierra.
\newblock Harnessing the power of large language models for automated code generation and verification.
\newblock \emph{Robotics}, 2024.

\bibitem[Authors(2024)]{xian24genesis}
Genesis Authors.
\newblock Genesis: A universal and generative physics engine for robotics and beyond, 2024.
\newblock URL \url{https://github.com/Genesis-Embodied-AI/Genesis}.

\bibitem[Blum et~al.(2020)Blum, Paillet, Laine, and Yoshida]{blum2020rl}
Tamir Blum, Gabin Paillet, Mickael Laine, and Kazuya Yoshida.
\newblock Rl star platform: Reinforcement learning for simulation based training of robots.
\newblock \emph{arXiv preprint arXiv:2009.09595}, 2020.

\bibitem[Brohan et~al.(2022)Brohan, Brown, Carbajal, Chebotar, Dabis, Finn, Gopalakrishnan, Hausman, Herzog, Hsu, et~al.]{brohan2022rt}
Anthony Brohan, Noah Brown, Justice Carbajal, Yevgen Chebotar, Joseph Dabis, Chelsea Finn, Keerthana Gopalakrishnan, Karol Hausman, Alex Herzog, Jasmine Hsu, et~al.
\newblock Rt-1: Robotics transformer for real-world control at scale.
\newblock \emph{arXiv preprint arXiv:2212.06817}, 2022.

\bibitem[Brown et~al.(2020)Brown, Mann, Ryder, Subbiah, Kaplan, Dhariwal, Neelakantan, Shyam, Sastry, Askell, et~al.]{brown2020language}
Tom Brown, Benjamin Mann, Nick Ryder, Melanie Subbiah, Jared~D Kaplan, Prafulla Dhariwal, Arvind Neelakantan, Pranav Shyam, Girish Sastry, Amanda Askell, et~al.
\newblock Language models are few-shot learners.
\newblock \emph{Advances in Neural Information Processing Systems}, 2020.

\bibitem[Bruce et~al.(2024)Bruce, Dennis, Edwards, Parker-Holder, Shi, Hughes, Lai, Mavalankar, Steigerwald, Apps, Aytar, Bechtle, Behbahani, Chan, Heess, Gonzalez, Osindero, Ozair, Reed, Zhang, Zolna, Clune, de~Freitas, Singh, and Rocktäschel]{bruce2024geniegenerativeinteractiveenvironments}
Jake Bruce, Michael Dennis, Ashley Edwards, Jack Parker-Holder, Yuge Shi, Edward Hughes, Matthew Lai, Aditi Mavalankar, Richie Steigerwald, Chris Apps, Yusuf Aytar, Sarah Bechtle, Feryal Behbahani, Stephanie Chan, Nicolas Heess, Lucy Gonzalez, Simon Osindero, Sherjil Ozair, Scott Reed, Jingwei Zhang, Konrad Zolna, Jeff Clune, Nando de~Freitas, Satinder Singh, and Tim Rocktäschel.
\newblock Genie: Generative interactive environments, 2024.
\newblock URL \url{https://arxiv.org/abs/2402.15391}.

\bibitem[Calli et~al.(2015{\natexlab{a}})Calli, Walsman, Singh, Srinivasa, Abbeel, and Dollar]{7254318}
Berk Calli, Aaron Walsman, Arjun Singh, Siddhartha Srinivasa, Pieter Abbeel, and Aaron~M. Dollar.
\newblock Benchmarking in manipulation research: Using the yale-cmu-berkeley object and model set.
\newblock \emph{IEEE Robotics \& Automation Magazine}, 2015{\natexlab{a}}.

\bibitem[Calli et~al.(2015{\natexlab{b}})Calli, Walsman, Singh, Srinivasa, Abbeel, and Dollar]{calli2015benchmarking}
Berk Calli, Aaron Walsman, Arjun Singh, Siddhartha Srinivasa, Pieter Abbeel, and Aaron~M Dollar.
\newblock Benchmarking in manipulation research: The ycb object and model set and benchmarking protocols.
\newblock \emph{arXiv preprint arXiv:1502.03143}, 2015{\natexlab{b}}.

\bibitem[Chang et~al.(2017)Chang, Dai, Funkhouser, Halber, Niebner, Savva, Song, Zeng, and Zhang]{chang2017matterport3d}
Angel Chang, Angela Dai, Thomas Funkhouser, Maciej Halber, Matthias Niebner, Manolis Savva, Shuran Song, Andy Zeng, and Yinda Zhang.
\newblock Matterport3d: Learning from rgb-d data in indoor environments.
\newblock In \emph{International Conference on 3D Vision}, 2017.

\bibitem[Chen et~al.(2024)Chen, Wang, Yang, and Liu]{chenobject}
Yuanpei Chen, Chen Wang, Yaodong Yang, and Karen Liu.
\newblock Object-centric dexterous manipulation from human motion data.
\newblock In \emph{Conference on Robot Learning}, 2024.

\bibitem[Cheng et~al.(2024{\natexlab{a}})Cheng, Ji, Yang, Zou, Kautz, B{\i}y{\i}k, Yin, Liu, and Wang]{cheng2024navila}
An-Chieh Cheng, Yandong Ji, Zhaojing Yang, Xueyan Zou, Jan Kautz, Erdem B{\i}y{\i}k, Hongxu Yin, Sifei Liu, and Xiaolong Wang.
\newblock Navila: Legged robot vision-language-action model for navigation.
\newblock \emph{arXiv preprint arXiv:2412.04453}, 2024{\natexlab{a}}.

\bibitem[Cheng et~al.(2024{\natexlab{b}})Cheng, Li, Yang, Yang, and Wang]{cheng2024open}
Xuxin Cheng, Jialong Li, Shiqi Yang, Ge~Yang, and Xiaolong Wang.
\newblock Open-television: Teleoperation with immersive active visual feedback.
\newblock \emph{arXiv preprint arXiv:2407.01512}, 2024{\natexlab{b}}.

\bibitem[Chi et~al.(2023)Chi, Feng, Du, Xu, Cousineau, Burchfiel, and Song]{chi2023diffusionpolicy}
Cheng Chi, Siyuan Feng, Yilun Du, Zhenjia Xu, Eric Cousineau, Benjamin Burchfiel, and Shuran Song.
\newblock Diffusion policy: Visuomotor policy learning via action diffusion.
\newblock In \emph{Robotics: Science and Systems}, 2023.

\bibitem[Chiappa et~al.(2024)Chiappa, Marin~Vargas, Huang, and Mathis]{chiappa2024latent}
Alberto~Silvio Chiappa, Alessandro Marin~Vargas, Ann Huang, and Alexander Mathis.
\newblock Latent exploration for reinforcement learning.
\newblock \emph{Advances in Neural Information Processing Systems}, 2024.

\bibitem[Collaboration(2023)]{padalkar2023open}
Open X-Embodiment Collaboration.
\newblock Open {X-E}mbodiment: Robotic learning datasets and {RT-X} models, 2023.

\bibitem[Coumans and Bai(2016--2021)]{coumans2021pybullet}
Erwin Coumans and Yunfei Bai.
\newblock Pybullet, a python module for physics simulation for games, robotics and machine learning, 2016--2021.

\bibitem[Coumans and Bai(2016--2023)]{coumans2020}
Erwin Coumans and Yunfei Bai.
\newblock Pybullet, a python module for physics simulation for games, robotics and machine learning, 2016--2023.

\bibitem[Cui et~al.(2024)Cui, Zhao, Wei, Zhang, Geng, Chen, and Wang]{cui2024gapartmanip}
Wenbo Cui, Chengyang Zhao, Songlin Wei, Jiazhao Zhang, Haoran Geng, Yaran Chen, and He~Wang.
\newblock Gapartmanip: A large-scale part-centric dataset for material-agnostic articulated object manipulation.
\newblock \emph{arXiv preprint arXiv:2411.18276}, 2024.

\bibitem[Dai et~al.(2022)Dai, Zhang, Li, Wu, Dong, Liu, Tan, and Wang]{dai2022domain}
Qiyu Dai, Jiyao Zhang, Qiwei Li, Tianhao Wu, Hao Dong, Ziyuan Liu, Ping Tan, and He~Wang.
\newblock Domain randomization-enhanced depth simulation and restoration for perceiving and grasping specular and transparent objects.
\newblock In \emph{European Conference on Computer Vision}, 2022.

\bibitem[Dasari et~al.(2019)Dasari, Ebert, Tian, Nair, Bucher, Schmeckpeper, Singh, Levine, and Finn]{dasari2019robonet}
Sudeep Dasari, Frederik Ebert, Stephen Tian, Suraj Nair, Bernadette Bucher, Karl Schmeckpeper, Siddharth Singh, Sergey Levine, and Chelsea Finn.
\newblock Robonet: Large-scale multi-robot learning.
\newblock \emph{arXiv preprint arXiv:1910.11215}, 2019.

\bibitem[Deitke et~al.(2022)Deitke, VanderBilt, Herrasti, Weihs, Ehsani, Salvador, Han, Kolve, Kembhavi, and Mottaghi]{deitke2022️proc}
Matt Deitke, Eli VanderBilt, Alvaro Herrasti, Luca Weihs, Kiana Ehsani, Jordi Salvador, Winson Han, Eric Kolve, Aniruddha Kembhavi, and Roozbeh Mottaghi.
\newblock Procthor: Large-scale embodied ai using procedural generation.
\newblock \emph{Advances in Neural Information Processing Systems}, 2022.

\bibitem[Deitke et~al.(2023)Deitke, Liu, Wallingford, Ngo, Michel, Kusupati, Fan, Laforte, Voleti, Gadre, VanderBilt, Kembhavi, Vondrick, Gkioxari, Ehsani, Schmidt, and Farhadi]{Deitke2023ObjaverseXLAU}
Matt Deitke, Ruoshi Liu, Matthew Wallingford, Huong Ngo, Oscar Michel, Aditya Kusupati, Alan Fan, Christian Laforte, Vikram~S. Voleti, Samir~Yitzhak Gadre, Eli VanderBilt, Aniruddha Kembhavi, Carl Vondrick, Georgia Gkioxari, Kiana Ehsani, Ludwig Schmidt, and Ali Farhadi.
\newblock Objaverse-xl: A universe of 10m+ 3d objects.
\newblock \emph{ArXiv}, 2023.

\bibitem[Deng et~al.(2009)Deng, Dong, Socher, Li, Li, and Fei-Fei]{deng2009imagenet}
Jia Deng, Wei Dong, Richard Socher, Li-Jia Li, Kai Li, and Li~Fei-Fei.
\newblock Imagenet: A large-scale hierarchical image database.
\newblock In \emph{Conference on Computer Vision and Pattern Recognition}, 2009.

\bibitem[Ding et~al.(2024)Ding, Geng, Xu, Fang, Zhang, Wei, Dai, Zhang, and Wang]{ding2024open6dor}
Yufei Ding, Haoran Geng, Chaoyi Xu, Xiaomeng Fang, Jiazhao Zhang, Songlin Wei, Qiyu Dai, Zhizheng Zhang, and He~Wang.
\newblock Open6dor: Benchmarking open-instruction 6-dof object rearrangement and a vlm-based approach.
\newblock In \emph{International Conference on Intelligent Robots and Systems}, 2024.

\bibitem[Dulac-Arnold et~al.(2019)Dulac-Arnold, Mankowitz, and Hester]{dulacarnold2019challengesrealworldreinforcementlearning}
Gabriel Dulac-Arnold, Daniel Mankowitz, and Todd Hester.
\newblock Challenges of real-world reinforcement learning, 2019.
\newblock URL \url{https://arxiv.org/abs/1904.12901}.

\bibitem[Erez et~al.(2015)Erez, Tassa, and Todorov]{erez2015simulation}
Tom Erez, Yuval Tassa, and Emanuel Todorov.
\newblock Simulation tools for model-based robotics: Comparison of bullet, havok, mujoco, ode and physx.
\newblock In \emph{International Conference on Robotics and Automation}, 2015.

\bibitem[Fang et~al.(2020)Fang, Wang, Gou, and Lu]{fang2020graspnet}
Hao-Shu Fang, Chenxi Wang, Minghao Gou, and Cewu Lu.
\newblock Graspnet-1billion: A large-scale benchmark for general object grasping.
\newblock In \emph{Proceedings of the IEEE/CVF Conference on Computer Vision and Pattern Recognition(CVPR)}, 2020.

\bibitem[Fang et~al.(2024)Fang, Fang, Tang, Liu, Wang, Wang, Zhu, and Lu]{fang2024rh20t}
Hao-Shu Fang, Hongjie Fang, Zhenyu Tang, Jirong Liu, Chenxi Wang, Junbo Wang, Haoyi Zhu, and Cewu Lu.
\newblock Rh20t: A comprehensive robotic dataset for learning diverse skills in one-shot.
\newblock In \emph{International Conference on Robotics and Automation}, 2024.

\bibitem[Freeman et~al.(2021)Freeman, Frey, Raichuk, Girgin, Mordatch, and Bachem]{brax2021github}
C.~Daniel Freeman, Erik Frey, Anton Raichuk, Sertan Girgin, Igor Mordatch, and Olivier Bachem.
\newblock Brax - a differentiable physics engine for large scale rigid body simulation, 2021.
\newblock URL \url{http://github.com/google/brax}.

\bibitem[Fu et~al.(2021)Fu, Jia, Gao, Gong, Zhao, Maybank, and Tao]{fu20213d}
Huan Fu, Rongfei Jia, Lin Gao, Mingming Gong, Binqiang Zhao, Steve Maybank, and Dacheng Tao.
\newblock 3d-future: 3d furniture shape with texture.
\newblock \emph{International Journal of Computer Vision (IJCV)}, 2021.

\bibitem[Garrett et~al.(2024)Garrett, Mandlekar, Wen, and Fox]{garrett2024skillmimicgen}
Caelan Garrett, Ajay Mandlekar, Bowen Wen, and Dieter Fox.
\newblock Skillmimicgen: Automated demonstration generation for efficient skill learning and deployment.
\newblock \emph{arXiv preprint arXiv:2410.18907}, 2024.

\bibitem[Geng et~al.(2023{\natexlab{a}})Geng, Li, Geng, Chen, Dong, and Wang]{geng2023partmanip}
Haoran Geng, Ziming Li, Yiran Geng, Jiayi Chen, Hao Dong, and He~Wang.
\newblock Partmanip: Learning cross-category generalizable part manipulation policy from point cloud observations.
\newblock In \emph{CVPR}, 2023{\natexlab{a}}.

\bibitem[Geng et~al.(2023{\natexlab{b}})Geng, Wei, Deng, Shen, Wang, and Guibas]{geng2023sage}
Haoran Geng, Songlin Wei, Congyue Deng, Bokui Shen, He~Wang, and Leonidas Guibas.
\newblock Sage: Bridging semantic and actionable parts for generalizable articulated-object manipulation under language instructions, 2023{\natexlab{b}}.

\bibitem[Geng et~al.(2023{\natexlab{c}})Geng, Xu, Zhao, Xu, Yi, Huang, and Wang]{geng2023gapartnet}
Haoran Geng, Helin Xu, Chengyang Zhao, Chao Xu, Li~Yi, Siyuan Huang, and He~Wang.
\newblock Gapartnet: Cross-category domain-generalizable object perception and manipulation via generalizable and actionable parts.
\newblock In \emph{Conference on Computer Vision and Pattern Recognition}, 2023{\natexlab{c}}.

\bibitem[Geng et~al.(2023{\natexlab{d}})Geng, An, Geng, Chen, Yang, and Dong]{geng2023rlafford}
Yiran Geng, Boshi An, Haoran Geng, Yuanpei Chen, Yaodong Yang, and Hao Dong.
\newblock Rlafford: End-to-end affordance learning for robotic manipulation.
\newblock In \emph{International Conference on Robotics and Automation}, 2023{\natexlab{d}}.

\bibitem[Gong et~al.(2023)Gong, Huang, Zhao, Geng, Gao, Wu, Ai, Zhou, Terzopoulos, Zhu, et~al.]{gong2023arnold}
Ran Gong, Jiangyong Huang, Yizhou Zhao, Haoran Geng, Xiaofeng Gao, Qingyang Wu, Wensi Ai, Ziheng Zhou, Demetri Terzopoulos, Song-Chun Zhu, et~al.
\newblock Arnold: A benchmark for language-grounded task learning with continuous states in realistic 3d scenes.
\newblock In \emph{International Conference on Computer Vision}, 2023.

\bibitem[Gu et~al.(2023)Gu, Xiang, Li, Ling, Liu, Mu, Tang, Tao, Wei, Yao, et~al.]{gu2023maniskill2}
Jiayuan Gu, Fanbo Xiang, Xuanlin Li, Zhan Ling, Xiqiang Liu, Tongzhou Mu, Yihe Tang, Stone Tao, Xinyue Wei, Yunchao Yao, et~al.
\newblock Maniskill2: A unified benchmark for generalizable manipulation skills.
\newblock \emph{arXiv preprint arXiv:2302.04659}, 2023.

\bibitem[Gu et~al.(2024)Gu, Wang, and Chen]{gu2024humanoidgymreinforcementlearninghumanoid}
Xinyang Gu, Yen-Jen Wang, and Jianyu Chen.
\newblock Humanoid-gym: Reinforcement learning for humanoid robot with zero-shot sim2real transfer, 2024.
\newblock URL \url{https://arxiv.org/abs/2404.05695}.

\bibitem[Ha et~al.(2023)Ha, Florence, and Song]{pmlr-v229-ha23a}
Huy Ha, Pete Florence, and Shuran Song.
\newblock Scaling up and distilling down: Language-guided robot skill acquisition.
\newblock In Jie Tan, Marc Toussaint, and Kourosh Darvish, editors, \emph{Conference on Robot Learning}, Proceedings of Machine Learning Research, 2023.
\newblock URL \url{https://proceedings.mlr.press/v229/ha23a.html}.

\bibitem[Ha et~al.(2024)Ha, Gao, Fu, Tan, and Song]{ha2024umi}
Huy Ha, Yihuai Gao, Zipeng Fu, Jie Tan, and Shuran Song.
\newblock Umi on legs: Making manipulation policies mobile with manipulation-centric whole-body controllers.
\newblock \emph{arXiv preprint arXiv:2407.10353}, 2024.

\bibitem[Hansen et~al.(2022)Hansen, Wang, and Su]{hansen2022tdmpc}
Nicklas Hansen, Xiaolong Wang, and Hao Su.
\newblock Temporal difference learning for model predictive control.
\newblock In \emph{International Conference on Machine Learning}, 2022.

\bibitem[Hansen et~al.(2024)Hansen, Su, and Wang]{hansen2024tdmpc2}
Nicklas Hansen, Hao Su, and Xiaolong Wang.
\newblock Td-mpc2: Scalable, robust world models for continuous control.
\newblock In \emph{International Conference on Learning Representations}, 2024.

\bibitem[He et~al.(2022)He, Chen, Xie, Li, Doll{\'a}r, and Girshick]{he2022masked}
Kaiming He, Xinlei Chen, Saining Xie, Yanghao Li, Piotr Doll{\'a}r, and Ross Girshick.
\newblock Masked autoencoders are scalable vision learners.
\newblock In \emph{Conference on Computer Vision and Pattern Recognition}, 2022.

\bibitem[Hu et~al.(2021)Hu, Shen, Wallis, Allen-Zhu, Li, Wang, Wang, and Chen]{hu2021loralowrankadaptationlarge}
Edward~J. Hu, Yelong Shen, Phillip Wallis, Zeyuan Allen-Zhu, Yuanzhi Li, Shean Wang, Lu~Wang, and Weizhu Chen.
\newblock Lora: Low-rank adaptation of large language models, 2021.
\newblock URL \url{https://arxiv.org/abs/2106.09685}.

\bibitem[Huang et~al.(2024)Huang, Yu, Chen, Geiger, and Gao]{Huang2DGS2024}
Binbin Huang, Zehao Yu, Anpei Chen, Andreas Geiger, and Shenghua Gao.
\newblock 2d gaussian splatting for geometrically accurate radiance fields.
\newblock In \emph{SIGGRAPH 2024 Conference Papers}, 2024.

\bibitem[Hussing et~al.(2023)Hussing, Mendez, Singrodia, Kent, and Eaton]{hussing2023robotic}
Marcel Hussing, Jorge~A Mendez, Anisha Singrodia, Cassandra Kent, and Eric Eaton.
\newblock Robotic manipulation datasets for offline compositional reinforcement learning.
\newblock \emph{arXiv preprint arXiv:2307.07091}, 2023.

\bibitem[James et~al.(2019)James, Freese, and Davison]{james2019pyrep}
Stephen James, Marc Freese, and Andrew~J. Davison.
\newblock Pyrep: Bringing v-rep to deep robot learning.
\newblock \emph{arXiv preprint arXiv:1906.11176}, 2019.

\bibitem[James et~al.(2020)James, Ma, Rovick~Arrojo, and Davison]{james2019rlbench}
Stephen James, Zicong Ma, David Rovick~Arrojo, and Andrew~J. Davison.
\newblock Rlbench: The robot learning benchmark \& learning environment.
\newblock \emph{Robotics and Automation Letters}, 2020.

\bibitem[Jia et~al.(2024)Jia, Chen, Yu, Wang, Niu, Liu, Li, and Huang]{jia2025sceneverse}
Baoxiong Jia, Yixin Chen, Huangyue Yu, Yan Wang, Xuesong Niu, Tengyu Liu, Qing Li, and Siyuan Huang.
\newblock Sceneverse: Scaling 3d vision-language learning for grounded scene understanding.
\newblock In \emph{European Conference on Computer Vision}, 2024.

\bibitem[Jiang et~al.(2024)Jiang, Xie, Lin, Xu, Wan, Mandlekar, Fan, and Zhu]{jiang2024dexmimicgen}
Zhenyu Jiang, Yuqi Xie, Kevin Lin, Zhenjia Xu, Weikang Wan, Ajay Mandlekar, Linxi Fan, and Yuke Zhu.
\newblock Dexmimicgen: Automated data generation for bimanual dexterous manipulation via imitation learning.
\newblock \emph{arXiv preprint arXiv:2410.24185}, 2024.

\bibitem[Kadian et~al.(2020)Kadian, Truong, Gokaslan, Clegg, Wijmans, Lee, Savva, Chernova, and Batra]{kadian2020sim2real}
Abhishek Kadian, Joanne Truong, Aaron Gokaslan, Alexander Clegg, Erik Wijmans, Stefan Lee, Manolis Savva, Sonia Chernova, and Dhruv Batra.
\newblock Sim2real predictivity: Does evaluation in simulation predict real-world performance?
\newblock \emph{IEEE Robotics and Automation Letters}, 2020.

\bibitem[Katara et~al.(2024)Katara, Xian, and Fragkiadaki]{katara2024gen2sim}
Pushkal Katara, Zhou Xian, and Katerina Fragkiadaki.
\newblock Gen2sim: Scaling up robot learning in simulation with generative models.
\newblock In \emph{2024 IEEE International Conference on Robotics and Automation (ICRA)}, 2024.

\bibitem[Kerbl et~al.(2023)Kerbl, Kopanas, Leimk{\"u}hler, and Drettakis]{kerbl3Dgaussians}
Bernhard Kerbl, Georgios Kopanas, Thomas Leimk{\"u}hler, and George Drettakis.
\newblock 3d gaussian splatting for real-time radiance field rendering.
\newblock \emph{ACM Transactions on Graphics}, 2023.
\newblock URL \url{https://repo-sam.inria.fr/fungraph/3d-gaussian-splatting/}.

\bibitem[Khazatsky et~al.(2024)Khazatsky, Pertsch, Nair, Balakrishna, Dasari, Karamcheti, Nasiriany, Srirama, Chen, Ellis, et~al.]{khazatsky2024droid}
Alexander Khazatsky, Karl Pertsch, Suraj Nair, Ashwin Balakrishna, Sudeep Dasari, Siddharth Karamcheti, Soroush Nasiriany, Mohan~Kumar Srirama, Lawrence~Yunliang Chen, Kirsty Ellis, et~al.
\newblock Droid: A large-scale in-the-wild robot manipulation dataset.
\newblock \emph{CoRR}, 2024.

\bibitem[Kim et~al.(2022)Kim, Danielczuk, Huang, and Goldberg]{kim2022ipcgraspsimreducingsim2realgap}
Chung~Min Kim, Michael Danielczuk, Isabella Huang, and Ken Goldberg.
\newblock Ipc-graspsim: Reducing the sim2real gap for parallel-jaw grasping with the incremental potential contact model, 2022.
\newblock URL \url{https://arxiv.org/abs/2111.01391}.

\bibitem[Kim et~al.(2024)Kim, Pertsch, Karamcheti, Xiao, Balakrishna, Nair, Rafailov, Foster, Lam, Sanketi, Vuong, Kollar, Burchfiel, Tedrake, Sadigh, Levine, Liang, and Finn]{kim24openvla}
{Moo Jin} Kim, Karl Pertsch, Siddharth Karamcheti, Ted Xiao, Ashwin Balakrishna, Suraj Nair, Rafael Rafailov, Ethan Foster, Grace Lam, Pannag Sanketi, Quan Vuong, Thomas Kollar, Benjamin Burchfiel, Russ Tedrake, Dorsa Sadigh, Sergey Levine, Percy Liang, and Chelsea Finn.
\newblock Openvla: An open-source vision-language-action model.
\newblock \emph{arXiv preprint arXiv:2406.09246}, 2024.

\bibitem[Kirillov et~al.(2023)Kirillov, Mintun, Ravi, Mao, Rolland, Gustafson, Xiao, Whitehead, Berg, Lo, et~al.]{kirillov2023segment}
Alexander Kirillov, Eric Mintun, Nikhila Ravi, Hanzi Mao, Chloe Rolland, Laura Gustafson, Tete Xiao, Spencer Whitehead, Alexander~C Berg, Wan-Yen Lo, et~al.
\newblock Segment anything.
\newblock In \emph{International Conference on Computer Vision}, 2023.

\bibitem[Krantz et~al.(2020)Krantz, Wijmans, Majumdar, Batra, and Lee]{Krantz2020BeyondTN}
Jacob Krantz, Erik Wijmans, Arjun Majumdar, Dhruv Batra, and Stefan Lee.
\newblock Beyond the nav-graph: Vision-and-language navigation in continuous environments.
\newblock In \emph{European Conference on Computer Vision}, 2020.

\bibitem[Krizhevsky et~al.(2012)Krizhevsky, Sutskever, and Hinton]{krizhevsky2012imagenet}
Alex Krizhevsky, Ilya Sutskever, and Geoffrey~E Hinton.
\newblock Imagenet classification with deep convolutional neural networks.
\newblock \emph{Advances in Neural Information Processing Systems}, 2012.

\bibitem[Ku et~al.(2020)Ku, Anderson, Patel, Ie, and Baldridge]{ku2020room}
Alexander Ku, Peter Anderson, Roma Patel, Eugene Ie, and Jason Baldridge.
\newblock Room-across-room: Multilingual vision-and-language navigation with dense spatiotemporal grounding.
\newblock In \emph{Annual Conference on Empirical Methods in Natural Language Processing}, 2020.

\bibitem[Kuang et~al.()Kuang, Elhafsi, Geng, Pavone, and Wang]{kuang2024skillblender}
Yuxuan Kuang, Amine Elhafsi, Haoran Geng, Marco Pavone, and Yue Wang.
\newblock Skillblender: Towards versatile humanoid whole-body control via skill blending.
\newblock In \emph{CoRL 2024 Workshop on Whole-body Control and Bimanual Manipulation: Applications in Humanoids and Beyond}.

\bibitem[Kuang et~al.(2024)Kuang, Ye, Geng, Mao, Deng, Guibas, Wang, and Wang]{kuang2024ramretrievalbasedaffordancetransfer}
Yuxuan Kuang, Junjie Ye, Haoran Geng, Jiageng Mao, Congyue Deng, Leonidas Guibas, He~Wang, and Yue Wang.
\newblock Ram: Retrieval-based affordance transfer for generalizable zero-shot robotic manipulation, 2024.
\newblock URL \url{https://arxiv.org/abs/2407.04689}.

\bibitem[Li et~al.(2023{\natexlab{a}})Li, Zhang, Wong, Gokmen, Srivastava, Mart{\'\i}n-Mart{\'\i}n, Wang, Levine, Lingelbach, Sun, et~al.]{li2023behavior}
Chengshu Li, Ruohan Zhang, Josiah Wong, Cem Gokmen, Sanjana Srivastava, Roberto Mart{\'\i}n-Mart{\'\i}n, Chen Wang, Gabrael Levine, Michael Lingelbach, Jiankai Sun, et~al.
\newblock Behavior-1k: A benchmark for embodied ai with 1,000 everyday activities and realistic simulation.
\newblock In \emph{Conference on Robot Learning}, 2023{\natexlab{a}}.

\bibitem[Li et~al.(2024{\natexlab{a}})Li, Liu, Li, Han, Geng, Wang, Zhu, Zhu, and Huang]{li2024ag2manip}
Puhao Li, Tengyu Liu, Yuyang Li, Muzhi Han, Haoran Geng, Shu Wang, Yixin Zhu, Song-Chun Zhu, and Siyuan Huang.
\newblock Ag2manip: Learning novel manipulation skills with agent-agnostic visual and action representations.
\newblock \emph{arXiv preprint arXiv:2404.17521}, 2024{\natexlab{a}}.

\bibitem[Li et~al.(2023{\natexlab{b}})Li, Zhang, Geng, Geng, Long, Shen, Zhang, Liu, and Dong]{li2023manipllm}
Xiaoqi Li, Mingxu Zhang, Yiran Geng, Haoran Geng, Yuxing Long, Yan Shen, Renrui Zhang, Jiaming Liu, and Hao Dong.
\newblock Manipllm: Embodied multimodal large language model for object-centric robotic manipulation, 2023{\natexlab{b}}.

\bibitem[Li et~al.(2024{\natexlab{b}})Li, Hsu, Gu, Pertsch, Mees, Walke, Fu, Lunawat, Sieh, Kirmani, Levine, Wu, Finn, Su, Vuong, and Xiao]{li24simpler}
Xuanlin Li, Kyle Hsu, Jiayuan Gu, Karl Pertsch, Oier Mees, Homer~Rich Walke, Chuyuan Fu, Ishikaa Lunawat, Isabel Sieh, Sean Kirmani, Sergey Levine, Jiajun Wu, Chelsea Finn, Hao Su, Quan Vuong, and Ted Xiao.
\newblock Evaluating real-world robot manipulation policies in simulation.
\newblock \emph{arXiv preprint arXiv:2405.05941}, 2024{\natexlab{b}}.

\bibitem[Li et~al.(2024{\natexlab{c}})Li, Zhang, Wu, Zhang, Geng, Dong, and He]{li2024unidoormanip}
Yu~Li, Xiaojie Zhang, Ruihai Wu, Zilong Zhang, Yiran Geng, Hao Dong, and Zhaofeng He.
\newblock Unidoormanip: Learning universal door manipulation policy over large-scale and diverse door manipulation environments.
\newblock \emph{arXiv preprint arXiv:2403.02604}, 2024{\natexlab{c}}.

\bibitem[Liu et~al.(2023)Liu, Zhu, Gao, Feng, Liu, Zhu, and Stone]{liu2023libero}
Bo~Liu, Yifeng Zhu, Chongkai Gao, Yihao Feng, Qiang Liu, Yuke Zhu, and Peter Stone.
\newblock Libero: Benchmarking knowledge transfer for lifelong robot learning.
\newblock \emph{arXiv preprint arXiv:2306.03310}, 2023.

\bibitem[Liu et~al.(2024)Liu, Canberk, Song, and Vondrick]{liu2024differentiablerobotrendering}
Ruoshi Liu, Alper Canberk, Shuran Song, and Carl Vondrick.
\newblock Differentiable robot rendering, 2024.
\newblock URL \url{https://arxiv.org/abs/2410.13851}.

\bibitem[Liu et~al.(2022)Liu, Liu, Jiang, Lyu, Wan, Shen, Liang, Fu, Wang, and Yi]{liu2022hoi4d}
Yunze Liu, Yun Liu, Che Jiang, Kangbo Lyu, Weikang Wan, Hao Shen, Boqiang Liang, Zhoujie Fu, He~Wang, and Li~Yi.
\newblock Hoi4d: A 4d egocentric dataset for category-level human-object interaction.
\newblock In \emph{Proceedings of the IEEE/CVF Conference on Computer Vision and Pattern Recognition}, 2022.

\bibitem[Lou et~al.(2024)Lou, Liu, Pan, Geng, Chen, Ma, Li, Wang, Feng, Shi, Luo, and Shi]{lou2024robogsphysicsconsistentspatialtemporal}
Haozhe Lou, Yurong Liu, Yike Pan, Yiran Geng, Jianteng Chen, Wenlong Ma, Chenglong Li, Lin Wang, Hengzhen Feng, Lu~Shi, Liyi Luo, and Yongliang Shi.
\newblock Robo-gs: A physics consistent spatial-temporal model for robotic arm with hybrid representation, 2024.
\newblock URL \url{https://arxiv.org/abs/2408.14873}.

\bibitem[Lu et~al.(2024)Lu, Wu, Li, Li, Zhu, Ning, Shen, Luo, Chen, and Dong]{lu2024garmentlab}
Haoran Lu, Ruihai Wu, Yitong Li, Sijie Li, Ziyu Zhu, Chuanruo Ning, Yan Shen, Longzan Luo, Yuanpei Chen, and Hao Dong.
\newblock Garmentlab: A unified simulation and benchmark for garment manipulation.
\newblock In \emph{Advances in Neural Information Processing Systems}, 2024.

\bibitem[Lyu et~al.(2024)Lyu, Chen, Du, Zhu, Liu, Wang, and Wang]{lyu2024scissorbot}
Jiangran Lyu, Yuxing Chen, Tao Du, Feng Zhu, Huiquan Liu, Yizhou Wang, and He~Wang.
\newblock Scissorbot: Learning generalizable scissor skill for paper cutting via simulation, imitation, and sim2real.
\newblock \emph{arXiv preprint arXiv:2409.13966}, 2024.

\bibitem[Ma et~al.(2024)Ma, Wang, Jia, Chen, Liu, Li, Chen, and Qiao]{ma2024lattelatentdiffusiontransformer}
Xin Ma, Yaohui Wang, Gengyun Jia, Xinyuan Chen, Ziwei Liu, Yuan-Fang Li, Cunjian Chen, and Yu~Qiao.
\newblock Latte: Latent diffusion transformer for video generation, 2024.
\newblock URL \url{https://arxiv.org/abs/2401.03048}.

\bibitem[Makoviychuk et~al.(2021)Makoviychuk, Wawrzyniak, Guo, Lu, Storey, Macklin, Hoeller, Rudin, Allshire, Handa, and State]{makoviychuk2021isaacgym}
Viktor Makoviychuk, Lukasz Wawrzyniak, Yunrong Guo, Michelle Lu, Kier Storey, Miles Macklin, David Hoeller, Nikita Rudin, Arthur Allshire, Ankur Handa, and Gavriel State.
\newblock Isaac gym: High performance gpu-based physics simulation for robot learning, 2021.

\bibitem[Mandlekar et~al.(2018)Mandlekar, Zhu, Garg, Booher, Spero, Tung, Gao, Emmons, Gupta, Orbay, et~al.]{mandlekar2018roboturk}
Ajay Mandlekar, Yuke Zhu, Animesh Garg, Jonathan Booher, Max Spero, Albert Tung, Julian Gao, John Emmons, Anchit Gupta, Emre Orbay, et~al.
\newblock Roboturk: A crowdsourcing platform for robotic skill learning through imitation.
\newblock In \emph{Conference on Robot Learning}, 2018.

\bibitem[Mandlekar et~al.(2020)Mandlekar, Xu, Mart{\'\i}n-Mart{\'\i}n, Zhu, Fei-Fei, and Savarese]{mandlekar2020human}
Ajay Mandlekar, Danfei Xu, Roberto Mart{\'\i}n-Mart{\'\i}n, Yuke Zhu, Li~Fei-Fei, and Silvio Savarese.
\newblock Human-in-the-loop imitation learning using remote teleoperation.
\newblock \emph{arXiv preprint arXiv:2012.06733}, 2020.

\bibitem[Mandlekar et~al.(2021)Mandlekar, Xu, Wong, Nasiriany, Wang, Kulkarni, Fei-Fei, Savarese, Zhu, and Mart{\'\i}n-Mart{\'\i}n]{mandlekar2021matters}
Ajay Mandlekar, Danfei Xu, Josiah Wong, Soroush Nasiriany, Chen Wang, Rohun Kulkarni, Li~Fei-Fei, Silvio Savarese, Yuke Zhu, and Roberto Mart{\'\i}n-Mart{\'\i}n.
\newblock What matters in learning from offline human demonstrations for robot manipulation.
\newblock \emph{arXiv preprint arXiv:2108.03298}, 2021.

\bibitem[Mandlekar et~al.(2023)Mandlekar, Nasiriany, Wen, Akinola, Narang, Fan, Zhu, and Fox]{mandlekar2023mimicgen}
Ajay Mandlekar, Soroush Nasiriany, Bowen Wen, Iretiayo Akinola, Yashraj Narang, Linxi Fan, Yuke Zhu, and Dieter Fox.
\newblock Mimicgen: A data generation system for scalable robot learning using human demonstrations.
\newblock In \emph{Conference on Robot Learning}, 2023.

\bibitem[Mao et~al.(2024)Mao, Zhao, Song, Shi, Ye, Zhang, Geng, Malik, Guizilini, and Wang]{uh1}
Jiageng Mao, Siheng Zhao, Siqi Song, Tianheng Shi, Junjie Ye, Mingtong Zhang, Haoran Geng, Jitendra Malik, Vitor Guizilini, and Yue Wang.
\newblock Learning from massive human videos for universal humanoid pose control.
\newblock \emph{arXiv preprint arXiv:2412.14172}, 2024.

\bibitem[Martinez-Gonzalez et~al.(2020)Martinez-Gonzalez, Oprea, Garcia-Garcia, Jover-Alvarez, Orts-Escolano, and Garcia-Rodriguez]{martinez2020unrealrox}
Pablo Martinez-Gonzalez, Sergiu Oprea, Alberto Garcia-Garcia, Alvaro Jover-Alvarez, Sergio Orts-Escolano, and Jose Garcia-Rodriguez.
\newblock Unrealrox: an extremely photorealistic virtual reality environment for robotics simulations and synthetic data generation.
\newblock \emph{Virtual Reality}, 2020.

\bibitem[Mees et~al.(2022)Mees, Hermann, Rosete-Beas, and Burgard]{mees2022calvin}
Oier Mees, Lukas Hermann, Erick Rosete-Beas, and Wolfram Burgard.
\newblock Calvin: A benchmark for language-conditioned policy learning for long-horizon robot manipulation tasks.
\newblock \emph{Robotics and Automation Letters}, 2022.

\bibitem[Mittal et~al.(2023)Mittal, Yu, Yu, Liu, Rudin, Hoeller, Yuan, Singh, Guo, Mazhar, Mandlekar, Babich, State, Hutter, and Garg]{mittal2023orbit}
Mayank Mittal, Calvin Yu, Qinxi Yu, Jingzhou Liu, Nikita Rudin, David Hoeller, Jia~Lin Yuan, Ritvik Singh, Yunrong Guo, Hammad Mazhar, Ajay Mandlekar, Buck Babich, Gavriel State, Marco Hutter, and Animesh Garg.
\newblock Orbit: A unified simulation framework for interactive robot learning environments.
\newblock \emph{Robotics and Automation Letters}, 2023.

\bibitem[Mu et~al.(2021)Mu, Ling, Xiang, Yang, Li, Tao, Huang, Jia, and Su]{mu2021maniskill}
Tongzhou Mu, Zhan Ling, Fanbo Xiang, Derek Yang, Xuanlin Li, Stone Tao, Zhiao Huang, Zhiwei Jia, and Hao Su.
\newblock Maniskill: Generalizable manipulation skill benchmark with large-scale demonstrations.
\newblock \emph{arXiv preprint arXiv:2107.14483}, 2021.

\bibitem[Nasiriany et~al.(2024)Nasiriany, Maddukuri, Zhang, Parikh, Lo, Joshi, Mandlekar, and Zhu]{nasiriany2024robocasa}
Soroush Nasiriany, Abhiram Maddukuri, Lance Zhang, Adeet Parikh, Aaron Lo, Abhishek Joshi, Ajay Mandlekar, and Yuke Zhu.
\newblock Robocasa: Large-scale simulation of everyday tasks for generalist robots.
\newblock In \emph{Robotics: Science and Systems}, 2024.

\bibitem[NVidia(2024{\natexlab{a}})]{physx}
NVidia.
\newblock Physx, 2024{\natexlab{a}}.
\newblock URL \url{https://nvidia-omniverse.github.io/PhysX/physx/5.5.0/}.

\bibitem[NVidia(2024{\natexlab{b}})]{vMaterials}
NVidia.
\newblock vmaterials, 2024{\natexlab{b}}.
\newblock URL \url{https://developer.nvidia.com/vmaterials}.

\bibitem[NVIDIA(2025)]{IsaacSim}
NVIDIA.
\newblock Isaacsim simulator, 2025.
\newblock URL \url{https://developer.nvidia.com/isaac/sim}.

\bibitem[{Octo Model Team} et~al.(2024){Octo Model Team}, Ghosh, Walke, Pertsch, Black, Mees, Dasari, Hejna, Xu, Luo, Kreiman, Tan, Sanketi, Vuong, Xiao, Sadigh, Finn, and Levine]{octo_2023}
{Octo Model Team}, Dibya Ghosh, Homer Walke, Karl Pertsch, Kevin Black, Oier Mees, Sudeep Dasari, Joey Hejna, Charles Xu, Jianlan Luo, Tobias Kreiman, {You Liang} Tan, Pannag Sanketi, Quan Vuong, Ted Xiao, Dorsa Sadigh, Chelsea Finn, and Sergey Levine.
\newblock Octo: An open-source generalist robot policy.
\newblock In \emph{Robotics: Science and Systems}, 2024.

\bibitem[Panerati et~al.(2021)Panerati, Zheng, Zhou, Xu, Prorok, and Schoellig]{panerati2021learning}
Jacopo Panerati, Hehui Zheng, SiQi Zhou, James Xu, Amanda Prorok, and Angela~P Schoellig.
\newblock Learning to fly—a gym environment with pybullet physics for reinforcement learning of multi-agent quadcopter control.
\newblock In \emph{International Conference on Intelligent Robots and Systems}, 2021.

\bibitem[Pavlakos et~al.(2024)Pavlakos, Shan, Radosavovic, Kanazawa, Fouhey, and Malik]{pavlakos2024reconstructing}
Georgios Pavlakos, Dandan Shan, Ilija Radosavovic, Angjoo Kanazawa, David Fouhey, and Jitendra Malik.
\newblock Reconstructing hands in 3{D} with transformers.
\newblock In \emph{CVPR}, 2024.

\bibitem[Perez et~al.(2017)Perez, Strub, de~Vries, Dumoulin, and Courville]{perez2017filmvisualreasoninggeneral}
Ethan Perez, Florian Strub, Harm de~Vries, Vincent Dumoulin, and Aaron Courville.
\newblock Film: Visual reasoning with a general conditioning layer, 2017.
\newblock URL \url{https://arxiv.org/abs/1709.07871}.

\bibitem[Podell et~al.(2023)Podell, English, Lacey, Blattmann, Dockhorn, Müller, Penna, and Rombach]{podell2023sdxlimprovinglatentdiffusion}
Dustin Podell, Zion English, Kyle Lacey, Andreas Blattmann, Tim Dockhorn, Jonas Müller, Joe Penna, and Robin Rombach.
\newblock Sdxl: Improving latent diffusion models for high-resolution image synthesis, 2023.
\newblock URL \url{https://arxiv.org/abs/2307.01952}.

\bibitem[Qi et~al.(2023)Qi, Kumar, Calandra, Ma, and Malik]{qi2023hand}
Haozhi Qi, Ashish Kumar, Roberto Calandra, Yi~Ma, and Jitendra Malik.
\newblock In-hand object rotation via rapid motor adaptation.
\newblock In \emph{Conference on Robot Learning}, 2023.

\bibitem[Qi et~al.(2024)Qi, Dong, Zhang, Geng, Han, Ge, Yi, and Ma]{qi2024shapellm}
Zekun Qi, Runpei Dong, Shaochen Zhang, Haoran Geng, Chunrui Han, Zheng Ge, Li~Yi, and Kaisheng Ma.
\newblock Shapellm: Universal 3d object understanding for embodied interaction.
\newblock \emph{arXiv preprint arXiv:2402.17766}, 2024.

\bibitem[Qin et~al.(2023)Qin, Yang, Huang, Van~Wyk, Su, Wang, Chao, and Fox]{qin2023anyteleop}
Yuzhe Qin, Wei Yang, Binghao Huang, Karl Van~Wyk, Hao Su, Xiaolong Wang, Yu-Wei Chao, and Dietor Fox.
\newblock Anyteleop: A general vision-based dexterous robot arm-hand teleoperation system.
\newblock \emph{arXiv preprint arXiv:2307.04577}, 2023.

\bibitem[Radford et~al.(2019)Radford, Wu, Child, Luan, Amodei, Sutskever, et~al.]{radford2019language}
Alec Radford, Jeffrey Wu, Rewon Child, David Luan, Dario Amodei, Ilya Sutskever, et~al.
\newblock Language models are unsupervised multitask learners.
\newblock \emph{OpenAI blog}, 2019.

\bibitem[Raffin et~al.(2021)Raffin, Hill, Gleave, Kanervisto, Ernestus, and Dormann]{raffin2021stable}
Antonin Raffin, Ashley Hill, Adam Gleave, Anssi Kanervisto, Maximilian Ernestus, and Noah Dormann.
\newblock Stable-baselines3: Reliable reinforcement learning implementations.
\newblock \emph{Journal of Machine Learning Research}, 2021.

\bibitem[Ravi et~al.(2024)Ravi, Gabeur, Hu, Hu, Ryali, Ma, Khedr, Rädle, Rolland, Gustafson, Mintun, Pan, Alwala, Carion, Wu, Girshick, Dollár, and Feichtenhofer]{ravi2024sam2segmentimages}
Nikhila Ravi, Valentin Gabeur, Yuan-Ting Hu, Ronghang Hu, Chaitanya Ryali, Tengyu Ma, Haitham Khedr, Roman Rädle, Chloe Rolland, Laura Gustafson, Eric Mintun, Junting Pan, Kalyan~Vasudev Alwala, Nicolas Carion, Chao-Yuan Wu, Ross Girshick, Piotr Dollár, and Christoph Feichtenhofer.
\newblock Sam 2: Segment anything in images and videos, 2024.
\newblock URL \url{https://arxiv.org/abs/2408.00714}.

\bibitem[Ren et~al.(2024)Ren, Li, Luo, Song, Chen, Liufu, Yang, Zheng, Xu, Huang, et~al.]{ren2024infiniteworld}
Pengzhen Ren, Min Li, Zhen Luo, Xinshuai Song, Ziwei Chen, Weijia Liufu, Yixuan Yang, Hao Zheng, Rongtao Xu, Zitong Huang, et~al.
\newblock Infiniteworld: A unified scalable simulation framework for general visual-language robot interaction.
\newblock \emph{arXiv preprint arXiv:2412.05789}, 2024.

\bibitem[Rohmer et~al.(2013)Rohmer, Singh, and Freese]{coppeliaSim}
E.~Rohmer, S.~P.~N. Singh, and M.~Freese.
\newblock Coppeliasim (formerly v-rep): a versatile and scalable robot simulation framework.
\newblock In \emph{Proc. of The International Conference on Intelligent Robots and Systems (IROS)}, 2013.
\newblock URL \url{www.coppeliarobotics.com}.

\bibitem[Rudin et~al.(2022)Rudin, Hoeller, Reist, and Hutter]{rudin2022learningwalkminutesusing}
Nikita Rudin, David Hoeller, Philipp Reist, and Marco Hutter.
\newblock Learning to walk in minutes using massively parallel deep reinforcement learning, 2022.
\newblock URL \url{https://arxiv.org/abs/2109.11978}.

\bibitem[Sch\"{o}nberger and Frahm(2016)]{schoenberger2016sfm}
Johannes~Lutz Sch\"{o}nberger and Jan-Michael Frahm.
\newblock Structure-from-motion revisited.
\newblock In \emph{Conference on Computer Vision and Pattern Recognition (CVPR)}, 2016.

\bibitem[Sch\"{o}nberger et~al.(2016)Sch\"{o}nberger, Zheng, Pollefeys, and Frahm]{schoenberger2016mvs}
Johannes~Lutz Sch\"{o}nberger, Enliang Zheng, Marc Pollefeys, and Jan-Michael Frahm.
\newblock Pixelwise view selection for unstructured multi-view stereo.
\newblock In \emph{European Conference on Computer Vision}, 2016.

\bibitem[Schulman et~al.(2017)Schulman, Wolski, Dhariwal, Radford, and Klimov]{schulman2017proximal}
John Schulman, Filip Wolski, Prafulla Dhariwal, Alec Radford, and Oleg Klimov.
\newblock Proximal policy optimization algorithms.
\newblock \emph{arXiv preprint arXiv:1707.06347}, 2017.

\bibitem[Sferrazza et~al.(2024)Sferrazza, Huang, Lin, Lee, and Abbeel]{sferrazza2024humanoidbench}
Carmelo Sferrazza, Dun-Ming Huang, Xingyu Lin, Youngwoon Lee, and Pieter Abbeel.
\newblock Humanoidbench: Simulated humanoid benchmark for whole-body locomotion and manipulation, 2024.

\bibitem[Shukla et~al.(2024)Shukla, Tao, and Su]{shukla2024maniskillhabbenchmarklowlevelmanipulation}
Arth Shukla, Stone Tao, and Hao Su.
\newblock Maniskill-hab: A benchmark for low-level manipulation in home rearrangement tasks, 2024.
\newblock URL \url{https://arxiv.org/abs/2412.13211}.

\bibitem[{Simulately Wiki}(2025)]{simulatelywiki}
{Simulately Wiki}.
\newblock Simulately wiki, 2025.
\newblock Accessed: 31 Jan 2025.

\bibitem[Song et~al.(2022)Song, Meng, and Ermon]{song2022denoisingdiffusionimplicitmodels}
Jiaming Song, Chenlin Meng, and Stefano Ermon.
\newblock Denoising diffusion implicit models, 2022.
\newblock URL \url{https://arxiv.org/abs/2010.02502}.

\bibitem[Sundaralingam et~al.(2023)Sundaralingam, Hari, Fishman, Garrett, Wyk, Blukis, Millane, Oleynikova, Handa, Ramos, Ratliff, and Fox]{curobo_report23}
Balakumar Sundaralingam, Siva Kumar~Sastry Hari, Adam Fishman, Caelan Garrett, Karl~Van Wyk, Valts Blukis, Alexander Millane, Helen Oleynikova, Ankur Handa, Fabio Ramos, Nathan Ratliff, and Dieter Fox.
\newblock curobo: Parallelized collision-free minimum-jerk robot motion generation, 2023.

\bibitem[Szot et~al.(2021)Szot, Clegg, Undersander, Wijmans, Zhao, Turner, Maestre, Mukadam, Chaplot, Maksymets, et~al.]{szot2021habitat}
Andrew Szot, Alexander Clegg, Eric Undersander, Erik Wijmans, Yili Zhao, John Turner, Noah Maestre, Mustafa Mukadam, Devendra~Singh Chaplot, Oleksandr Maksymets, et~al.
\newblock Habitat 2.0: Training home assistants to rearrange their habitat.
\newblock \emph{Advances in Neural Information Processing Systems}, 2021.

\bibitem[Tao et~al.(2024)Tao, Xiang, Shukla, Qin, Hinrichsen, Yuan, Bao, Lin, Liu, kai Chan, Gao, Li, Mu, Xiao, Gurha, Huang, Calandra, Chen, Luo, and Su]{tao2024maniskill3}
Stone Tao, Fanbo Xiang, Arth Shukla, Yuzhe Qin, Xander Hinrichsen, Xiaodi Yuan, Chen Bao, Xinsong Lin, Yulin Liu, Tse kai Chan, Yuan Gao, Xuanlin Li, Tongzhou Mu, Nan Xiao, Arnav Gurha, Zhiao Huang, Roberto Calandra, Rui Chen, Shan Luo, and Hao Su.
\newblock Maniskill3: Gpu parallelized robotics simulation and rendering for generalizable embodied ai.
\newblock \emph{arXiv preprint arXiv:2410.00425}, 2024.

\bibitem[team(2024)]{polyak2024moviegencastmedia}
Movie~Gen team.
\newblock Movie gen: A cast of media foundation models, 2024.
\newblock URL \url{https://arxiv.org/abs/2410.13720}.

\bibitem[Todorov et~al.(2012)Todorov, Erez, and Tassa]{todorov2012mujoco}
Emanuel Todorov, Tom Erez, and Yuval Tassa.
\newblock Mujoco: A physics engine for model-based control.
\newblock In \emph{International Conference on Intelligent Robots and Systems}, 2012.

\bibitem[Towers et~al.(2024)Towers, Kwiatkowski, Terry, Balis, De~Cola, Deleu, Goul{\~a}o, Kallinteris, Krimmel, KG, et~al.]{towers2024gymnasium}
Mark Towers, Ariel Kwiatkowski, Jordan Terry, John~U Balis, Gianluca De~Cola, Tristan Deleu, Manuel Goul{\~a}o, Andreas Kallinteris, Markus Krimmel, Arjun KG, et~al.
\newblock Gymnasium: A standard interface for reinforcement learning environments.
\newblock \emph{arXiv preprint arXiv:2407.17032}, 2024.

\bibitem[Vaccaro et~al.(2013)Vaccaro, Crisp, Fellner, Jackson, Kleeman, and Pavelka]{vaccaro2013robotic}
Christine~M Vaccaro, Catrina~C Crisp, Angela~N Fellner, Christopher Jackson, Steven~D Kleeman, and James Pavelka.
\newblock Robotic virtual reality simulation plus standard robotic orientation versus standard robotic orientation alone: a randomized controlled trial.
\newblock \emph{Urogynecology}, 2013.

\bibitem[Vlasic et~al.(2007)Vlasic, Adelsberger, Vannucci, Barnwell, Gross, Matusik, and Popovi{\'c}]{vlasic2007practical}
Daniel Vlasic, Rolf Adelsberger, Giovanni Vannucci, John Barnwell, Markus Gross, Wojciech Matusik, and Jovan Popovi{\'c}.
\newblock Practical motion capture in everyday surroundings.
\newblock \emph{ACM transactions on graphics (TOG)}, 2007.

\bibitem[Wan et~al.(2020)Wan, Wang, Liu, Yang, and Song]{wan2020deepclaw}
Fang Wan, Haokun Wang, Xiaobo Liu, Linhan Yang, and Chaoyang Song.
\newblock Deepclaw: A robotic hardware benchmarking platform for learning object manipulation.
\newblock In \emph{2020 IEEE/ASME International Conference on Advanced Intelligent Mechatronics (AIM)}, 2020.

\bibitem[Wan et~al.(2023)Wan, Geng, Liu, Shan, Yang, Yi, and Wang]{wan2023unidexgrasp++}
Weikang Wan, Haoran Geng, Yun Liu, Zikang Shan, Yaodong Yang, Li~Yi, and He~Wang.
\newblock Unidexgrasp++: Improving dexterous grasping policy learning via geometry-aware curriculum and iterative generalist-specialist learning.
\newblock \emph{arXiv preprint arXiv:2304.00464}, 2023.

\bibitem[Wang et~al.(2024{\natexlab{a}})Wang, Chen, Huang, Ben, Wang, Mi, Huang, Zhao, Chen, Yang, et~al.]{wang2024grutopia}
Hanqing Wang, Jiahe Chen, Wensi Huang, Qingwei Ben, Tai Wang, Boyu Mi, Tao Huang, Siheng Zhao, Yilun Chen, Sizhe Yang, et~al.
\newblock Grutopia: Dream general robots in a city at scale.
\newblock \emph{arXiv preprint arXiv:2407.10943}, 2024{\natexlab{a}}.

\bibitem[Wang et~al.(2024{\natexlab{b}})Wang, Qin, Kuang, Korkmaz, Gurumoorthy, Su, and Wang]{wang2024cyberdemo}
Jun Wang, Yuzhe Qin, Kaiming Kuang, Yigit Korkmaz, Akhilan Gurumoorthy, Hao Su, and Xiaolong Wang.
\newblock Cyberdemo: Augmenting simulated human demonstration for real-world dexterous manipulation.
\newblock In \emph{Proceedings of the IEEE/CVF Conference on Computer Vision and Pattern Recognition}, 2024{\natexlab{b}}.

\bibitem[Wang et~al.(2023{\natexlab{a}})Wang, Ling, Yuan, Shridhar, Bao, Qin, Wang, Xu, and Wang]{wang2023gensim}
Lirui Wang, Yiyang Ling, Zhecheng Yuan, Mohit Shridhar, Chen Bao, Yuzhe Qin, Bailin Wang, Huazhe Xu, and Xiaolong Wang.
\newblock Gensim: Generating robotic simulation tasks via large language models.
\newblock \emph{arXiv preprint arXiv:2310.01361}, 2023{\natexlab{a}}.

\bibitem[Wang et~al.(2023{\natexlab{b}})Wang, Xian, Chen, Wang, Wang, Fragkiadaki, Erickson, Held, and Gan]{wang2023robogen}
Yufei Wang, Zhou Xian, Feng Chen, Tsun-Hsuan Wang, Yian Wang, Katerina Fragkiadaki, Zackory Erickson, David Held, and Chuang Gan.
\newblock Robogen: Towards unleashing infinite data for automated robot learning via generative simulation.
\newblock \emph{arXiv preprint arXiv:2311.01455}, 2023{\natexlab{b}}.

\bibitem[Wei et~al.(2024)Wei, Geng, Chen, Deng, Wenbo, Zhao, Fang, Guibas, and Wang]{wei2024droma}
Songlin Wei, Haoran Geng, Jiayi Chen, Congyue Deng, Cui Wenbo, Chengyang Zhao, Xiaomeng Fang, Leonidas Guibas, and He~Wang.
\newblock D3roma: Disparity diffusion-based depth sensing for material-agnostic robotic manipulation.
\newblock In \emph{Conference on Robot Learning}, 2024.
\newblock URL \url{https://openreview.net/forum?id=7E3JAys1xO}.

\bibitem[Xiang et~al.(2020)Xiang, Qin, Mo, Xia, Zhu, Liu, Liu, Jiang, Yuan, Wang, Yi, Chang, Guibas, and Su]{xiang2020sapien}
Fanbo Xiang, Yuzhe Qin, Kaichun Mo, Yikuan Xia, Hao Zhu, Fangchen Liu, Minghua Liu, Hanxiao Jiang, Yifu Yuan, He~Wang, Li~Yi, Angel~X. Chang, Leonidas~J. Guibas, and Hao Su.
\newblock {SAPIEN}: A simulated part-based interactive environment.
\newblock In \emph{Conference on Computer Vision and Pattern Recognition}, 2020.

\bibitem[Xie et~al.(2024)Xie, Zhao, Wu, Liu, Luo, Zhong, Yang, and Yu]{xie2024textreward}
Tianbao Xie, Siheng Zhao, Chen~Henry Wu, Yitao Liu, Qian Luo, Victor Zhong, Yanchao Yang, and Tao Yu.
\newblock Text2reward: Reward shaping with language models for reinforcement learning.
\newblock In \emph{International Conference on Learning Representations}, 2024.
\newblock URL \url{https://openreview.net/forum?id=tUM39YTRxH}.

\bibitem[Xu et~al.(2023)Xu, Wan, Zhang, Liu, Shan, Shen, Wang, Geng, Weng, Chen, et~al.]{xu2023unidexgrasp}
Yinzhen Xu, Weikang Wan, Jialiang Zhang, Haoran Liu, Zikang Shan, Hao Shen, Ruicheng Wang, Haoran Geng, Yijia Weng, Jiayi Chen, et~al.
\newblock Unidexgrasp: Universal robotic dexterous grasping via learning diverse proposal generation and goal-conditioned policy.
\newblock In \emph{Proceedings of the IEEE/CVF Conference on Computer Vision and Pattern Recognition}, 2023.

\bibitem[Yadan(2019)]{Yadan2019Hydra}
Omry Yadan.
\newblock Hydra - a framework for elegantly configuring complex applications, 2019.
\newblock URL \url{https://github.com/facebookresearch/hydra}.

\bibitem[Yang et~al.(2021)Yang, Ji, Wu, and Lai]{yang2021open}
Xintong Yang, Ze~Ji, Jing Wu, and Yu-Kun Lai.
\newblock An open-source multi-goal reinforcement learning environment for robotic manipulation with pybullet.
\newblock In \emph{Annual Conference Towards Autonomous Robotic Systems}, 2021.

\bibitem[Yang et~al.(2024)Yang, Jia, Zhi, and Huang]{yang2024physcene}
Yandan Yang, Baoxiong Jia, Peiyuan Zhi, and Siyuan Huang.
\newblock Physcene: Physically interactable 3d scene synthesis for embodied ai.
\newblock In \emph{Conference on Computer Vision and Pattern Recognition}, 2024.

\bibitem[Ye et~al.(2024{\natexlab{a}})Ye, Nie, Chang, Chen, Zhi, and Han]{ye2024gaustudio}
Chongjie Ye, Yinyu Nie, Jiahao Chang, Yuantao Chen, Yihao Zhi, and Xiaoguang Han.
\newblock Gaustudio: A modular framework for 3d gaussian splatting and beyond.
\newblock \emph{arXiv preprint arXiv:2403.19632}, 2024{\natexlab{a}}.

\bibitem[Ye et~al.(2024{\natexlab{b}})Ye, Qiu, Gu, Zuo, Wu, Dong, Bo, Xiu, and Han]{ye2024stablenormal}
Chongjie Ye, Lingteng Qiu, Xiaodong Gu, Qi~Zuo, Yushuang Wu, Zilong Dong, Liefeng Bo, Yuliang Xiu, and Xiaoguang Han.
\newblock Stablenormal: Reducing diffusion variance for stable and sharp normal.
\newblock \emph{ACM Transactions on Graphics (TOG)}, 2024{\natexlab{b}}.

\bibitem[Ye et~al.(2024{\natexlab{c}})Ye, Li, Kerr, Turkulainen, Yi, Pan, Seiskari, Ye, Hu, Tancik, and Kanazawa]{ye2024gsplatopensourcelibrarygaussian}
Vickie Ye, Ruilong Li, Justin Kerr, Matias Turkulainen, Brent Yi, Zhuoyang Pan, Otto Seiskari, Jianbo Ye, Jeffrey Hu, Matthew Tancik, and Angjoo Kanazawa.
\newblock gsplat: An open-source library for {Gaussian} splatting.
\newblock \emph{arXiv preprint arXiv:2409.06765}, 2024{\natexlab{c}}.
\newblock URL \url{https://arxiv.org/abs/2409.06765}.

\bibitem[Yu et~al.(2019)Yu, Quillen, He, Julian, Hausman, Finn, and Levine]{yu2019metaworld}
Tianhe Yu, Deirdre Quillen, Zhanpeng He, Ryan Julian, Karol Hausman, Chelsea Finn, and Sergey Levine.
\newblock Meta-world: A benchmark and evaluation for multi-task and meta reinforcement learning.
\newblock In \emph{Conference on Robot Learning}, 2019.
\newblock URL \url{https://arxiv.org/abs/1910.10897}.

\bibitem[Zakka et~al.(2025)Zakka, Tabanpour, Liao, Haiderbhai, Holt, Luo, Allshire, Frey, Sreenath, Kahrs, Sferrazza, Tassa, and Abbeel]{zakka2025mujocoplayground}
Kevin Zakka, Baruch Tabanpour, Qiayuan Liao, Mustafa Haiderbhai, Samuel Holt, Jing~Yuan Luo, Arthur Allshire, Erik Frey, Koushil Sreenath, Lueder~A. Kahrs, Carlo Sferrazza, Yuval Tassa, and Pieter Abbeel.
\newblock Mujoco playground: An open-source framework for gpu-accelerated robot learning and sim-to-real transfer., 2025.
\newblock URL \url{https://github.com/google-deepmind/mujoco_playground}.

\bibitem[Zeng et~al.(2017)Zeng, Song, Nie{\ss}ner, Fisher, Xiao, and Funkhouser]{zeng20163dmatch}
Andy Zeng, Shuran Song, Matthias Nie{\ss}ner, Matthew Fisher, Jianxiong Xiao, and Thomas Funkhouser.
\newblock 3dmatch: Learning local geometric descriptors from rgb-d reconstructions.
\newblock In \emph{CVPR}, 2017.

\bibitem[Zhang et~al.(2024{\natexlab{a}})Zhang, Liu, Li, Yu, Geng, Ding, Chen, and Wang]{zhang2024dexgraspnet}
Jialiang Zhang, Haoran Liu, Danshi Li, XinQiang Yu, Haoran Geng, Yufei Ding, Jiayi Chen, and He~Wang.
\newblock Dexgraspnet 2.0: Learning generative dexterous grasping in large-scale synthetic cluttered scenes.
\newblock In \emph{Conference on Robot Learning}, 2024{\natexlab{a}}.

\bibitem[Zhang et~al.(2024{\natexlab{b}})Zhang, Liu, Li, Yu, Geng, Ding, Chen, and Wang]{zhang2024dexgraspnet20learninggenerative}
Jialiang Zhang, Haoran Liu, Danshi Li, Xinqiang Yu, Haoran Geng, Yufei Ding, Jiayi Chen, and He~Wang.
\newblock Dexgraspnet 2.0: Learning generative dexterous grasping in large-scale synthetic cluttered scenes, 2024{\natexlab{b}}.

\bibitem[Zhang et~al.(2024{\natexlab{c}})Zhang, Wang, Wang, Li, Liu, Wei, Wang, Zhang, and Wang]{zhang2024uni}
Jiazhao Zhang, Kunyu Wang, Shaoan Wang, Minghan Li, Haoran Liu, Songlin Wei, Zhongyuan Wang, Zhizheng Zhang, and He~Wang.
\newblock Uni-navid: A video-based vision-language-action model for unifying embodied navigation tasks.
\newblock \emph{arXiv preprint arXiv:2412.06224}, 2024{\natexlab{c}}.

\bibitem[Zhang et~al.(2024{\natexlab{d}})Zhang, Wang, Xu, Zhou, Hong, Fang, Wu, Zhang, and Wang]{zhang2024navid}
Jiazhao Zhang, Kunyu Wang, Rongtao Xu, Gengze Zhou, Yicong Hong, Xiaomeng Fang, Qi~Wu, Zhizheng Zhang, and He~Wang.
\newblock Navid: Video-based vlm plans the next step for vision-and-language navigation.
\newblock \emph{Robotics: Science and Systems}, 2024{\natexlab{d}}.

\bibitem[Zhao et~al.(2023)Zhao, Kumar, Levine, and Finn]{zhao2023learning}
Tony~Z Zhao, Vikash Kumar, Sergey Levine, and Chelsea Finn.
\newblock Learning fine-grained bimanual manipulation with low-cost hardware.
\newblock \emph{arXiv preprint arXiv:2304.13705}, 2023.

\bibitem[Zheng et~al.(2025)Zheng, Xue, Zarate, and Song]{zheng2025gstargaussiansurfacetracking}
Chengwei Zheng, Lixin Xue, Juan Zarate, and Jie Song.
\newblock Gstar: Gaussian surface tracking and reconstruction, 2025.
\newblock URL \url{https://arxiv.org/abs/2501.10283}.

\bibitem[Zhou et~al.(2022)Zhou, Yang, Loy, and Liu]{zhou2022learning}
Kaiyang Zhou, Jingkang Yang, Chen~Change Loy, and Ziwei Liu.
\newblock Learning to prompt for vision-language models.
\newblock \emph{International Journal of Computer Vision}, 2022.

\bibitem[Zhu et~al.(2024)Zhu, Wu, Guo, Liu, Cheang, and Kong]{zhu2024irasimlearninginteractiverealrobot}
Fangqi Zhu, Hongtao Wu, Song Guo, Yuxiao Liu, Chilam Cheang, and Tao Kong.
\newblock Irasim: Learning interactive real-robot action simulators, 2024.
\newblock URL \url{https://arxiv.org/abs/2406.14540}.

\bibitem[Zhu et~al.(2020)Zhu, Wong, Mandlekar, Mart\'{i}n-Mart\'{i}n, Joshi, Nasiriany, and Zhu]{zhu2020robosuite}
Yuke Zhu, Josiah Wong, Ajay Mandlekar, Roberto Mart\'{i}n-Mart\'{i}n, Abhishek Joshi, Soroush Nasiriany, and Yifeng Zhu.
\newblock robosuite: A modular simulation framework and benchmark for robot learning.
\newblock In \emph{arXiv preprint arXiv:2009.12293}, 2020.

\end{thebibliography}

%%% Supp
\newpage
\hypersetup{pdfborder={0 0 0}}

\tableofcontents
\newpage

\hypersetup{pdfborder={0 0 1}, linkbordercolor={1 0 0}}
\section{Simulators Overview}

In the field of robotics, simulators play an important role. It is the womb of a robot, taking responsibility for training and testing a robot's behaviors before it was "born" into the real world. Therefore, the functionalities are crucial for a successful robotic application. Users require different functions of simulators according to their specific scenarios: whether it is a photorealistic task which requires accurate rendering of a close-to-real virtual world, or a massive parallel scene that is designed for efficient reinforcement learning. All the requirements may influence the choice of the simulator. In order to reduce the pain users need to endure in getting them familiarized with each new simulator, we incorporated multiple simulators into the \textsc{RoboVerse} platform and listed specifications of the simulators currently supported by \textsc{RoboVerse} in \fref{tab:simulators}.

\begin{table*}[ht]
    \centering
    \renewcommand{\arraystretch}{1.4}
    \begin{tabular}{l|c|c|c|c|c|c}
        \toprule
        \textbf{Simulator} & \textbf{Physics Engine} & \textbf{Rendering} & \textbf{Sensor Support} & \textbf{Dynamics} & \textbf{GPU} & \textbf{Open} \\
        \midrule
        SAPIEN~\cite{xiang2020sapien} & PhysX-5, Warp & \makecell{Rasterization\\RayTracing} & RGBD; Force; Contact & Rigid; Soft; Fluid & \checkmark & \checkmark \\
        \hline
        PyBullet~\cite{coumans2021pybullet} & Bullet & Rasterization & \makecell{RGBD; Force\\IMU; Tactile} & Rigid; Soft; Cloth &  & \checkmark \\
        \hline
        MuJoCo~\cite{todorov2012mujoco} & MuJoCo & Rasterization & \makecell{RGBD; Force\\IMU; Tactile} & Rigid;Soft;Cloth & \checkmark & \checkmark \\
        \hline
        CoppeliaSim~\cite{coppeliaSim} & \makecell{MuJoCo; Bullet\\ODE; Newton; Vortex} & Rasterization & RGBD; Force; Contact & Rigid;Soft;Cloth &  & \checkmark \\
        \hline
        Isaac Sim~\cite{IsaacSim} & PhysX-5 & RayTracing & \makecell{RGBD; Lidar; Force\\Effort; IMU; Contact\\Proximity} & \makecell{Rigid; Soft\\Cloth; Fluid} & \checkmark &  \\
        \hline
        Isaac Gym~\cite{makoviychuk2021isaacgym} & PhysX-5, Flex & Rasterization & RGBD; Force; Contact & Rigid; Soft; Cloth & \checkmark &  \\
        \hline
        Genesis~\cite{xian24genesis} & Genesis & \makecell{Rasterization\\RayTracing} & RGBD; Force; Tactile & Rigid; Soft & \checkmark & \checkmark \\
        \bottomrule
        % \textcolor{blue}{Genesis} & PhysX-5, Flex & Rasterization & RGBD; Force; Contact & Rigid; Soft; Cloth & \checkmark &  \\
        % \hline
    \end{tabular}
    \caption{Comparison of Physics Simulators~\cite{simulatelywiki}. The column \textbf{GPU} denotes whether the simulator can use GPU-accelerated computation. The column \textbf{Open} denotes whether the simulator is open-source.}
    \label{tab:simulators}
\end{table*}

Due to the complexity of physics simulation and rendering, current simulators cannot depict the real world well enough. Our experiments revealed some common issues of nowadays simulators in the basic physics laws. The experimental results on fundamental conservation laws may be a pessimistic sign on our hope of direct sim-to-real transfer of more complicated robotic behaviors.

We conducted experiments on three basic conservation laws of physics in three simulators.

In the experiments for Conservation of Momentum, two rigid bodies are placed in a gravity-free environment, their initial states are set to have an elastic collision.

In the experiments for Conservation of Angular Momentum, one or two rigid bodies are placed in the gravity-free environment, and their initial states are set to rotate. We calculate and record the overall angular momentum as the system evolves.

In the experiments for Conservation of Kinetic Energy, two rigid bodies are placed in the gravity-free environment, and their initial states are set to have a rotation-free elastic collision. This setup allows us to directly observe the conservation of kinetic energy regardless of the results of experiments on angular momentum.

\begin{figure}[ht]
    \centering
    \begin{subfigure}[b]{0.32\textwidth}  % Adjust width (0.3 for 3 images)
        \includegraphics[trim={12 0 10 30}, clip, width=\textwidth]{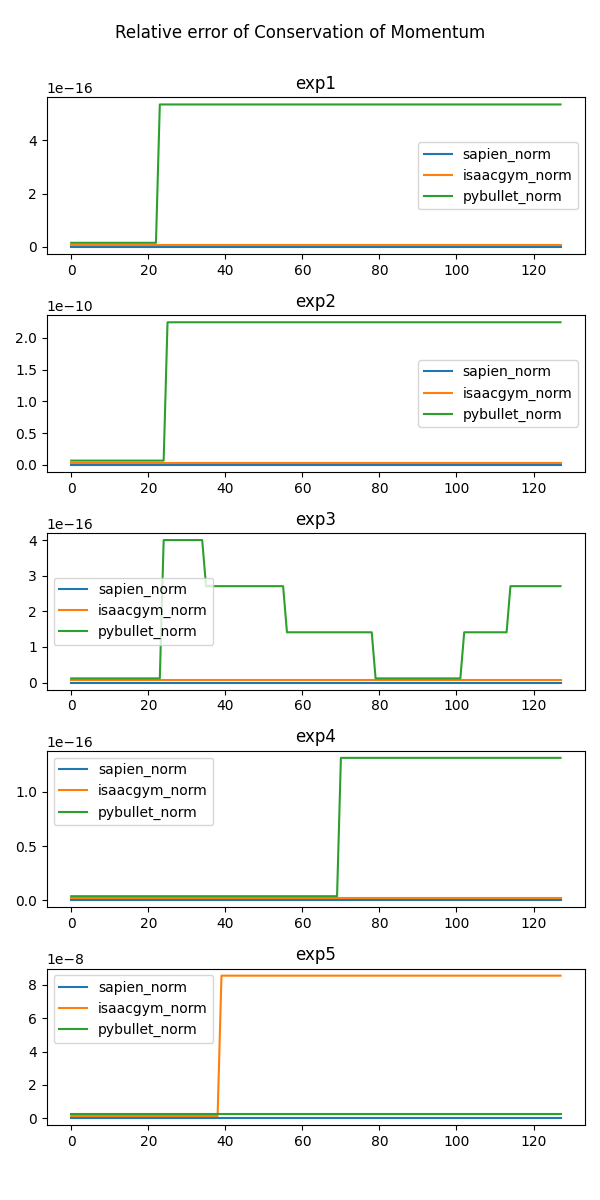}
        \caption{Momentum}
        \label{fig:momentum}
    \end{subfigure}
    \hfill
    \begin{subfigure}[b]{0.32\textwidth}
        \includegraphics[trim={12 0 10 30}, clip, width=\textwidth]{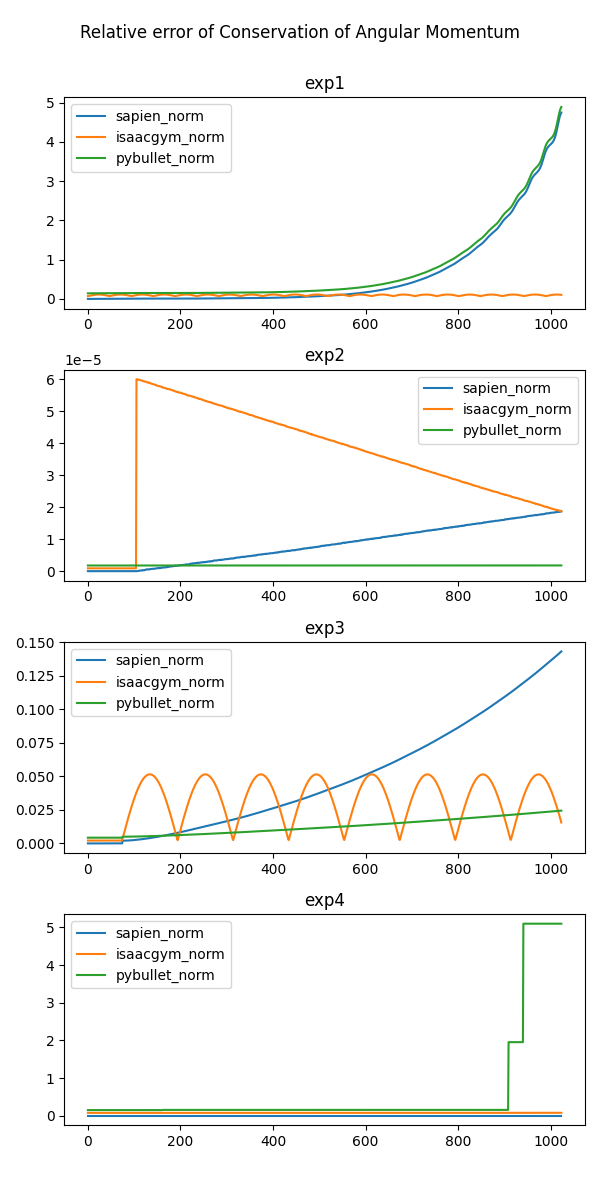}
        \caption{Angular}
        \label{fig:angular}
    \end{subfigure}
    \hfill
    \begin{subfigure}[b]{0.32\textwidth}
        \includegraphics[trim={12 0 10 30}, clip, width=\textwidth]{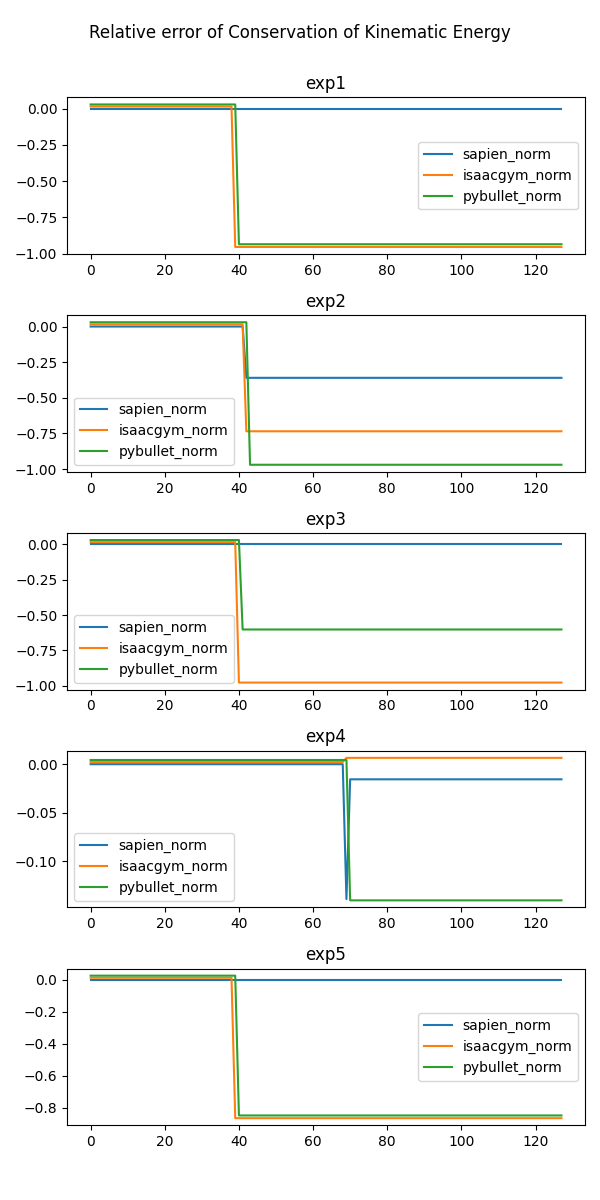}
        \caption{Kinetic Energy}
        \label{fig:energy}
    \end{subfigure}
    \caption{Three series of experiments on conservation laws in simulators. Blue, orange and green lines are data collected from SAPIEN, Isaac Gym and PyBullet respectively.}
    \label{fig:conservation}
\end{figure}

From the results listed in~\fref{fig:conservation}, we can easily notice that basic conservation laws are not kept in the three simulators. However, different simulators behave differently in different experimental setups, which indicates that depending on the needs of different tasks, we may need to choose different simulators for more accurate results. This highlights the necessity of a tool that helps users to easily transfer tasks among simulators.

\section{The \textsc{MetaSim} Framework}

\subsection{Architecture Overview}

\begin{figure*}[hbt] \centering
    \includegraphics[width=0.44\linewidth]{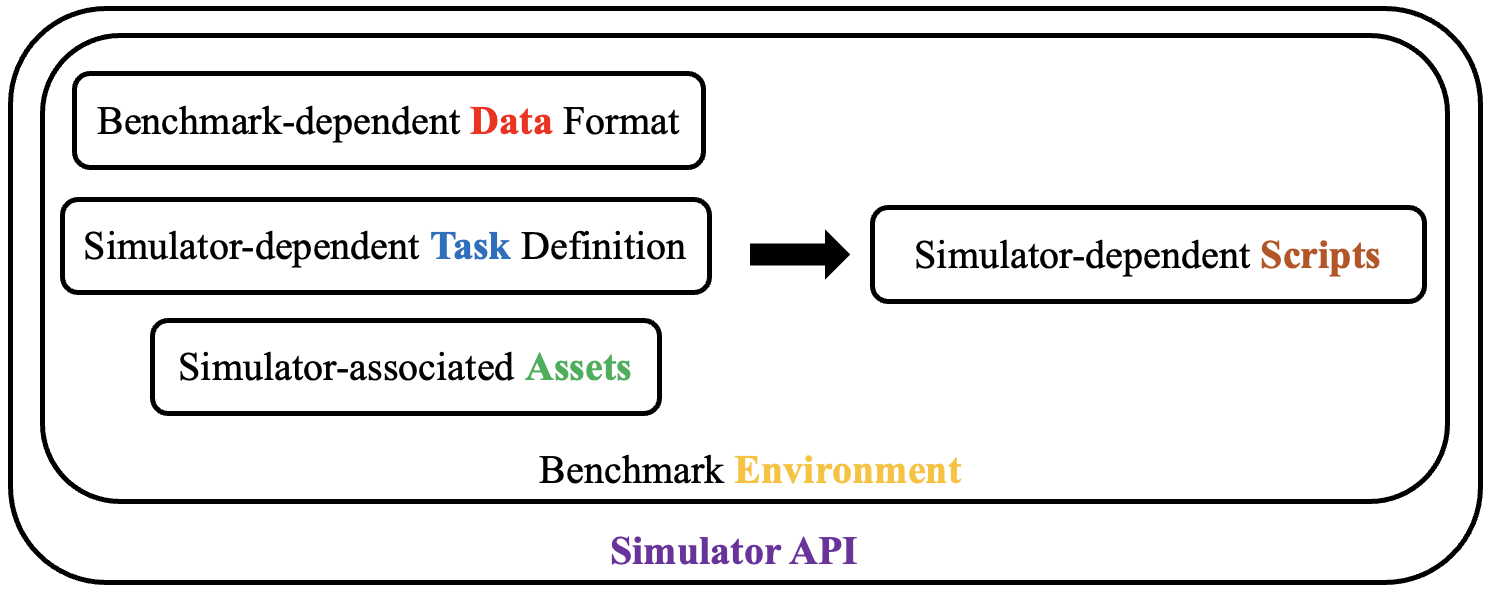}
    \hspace{0.0\linewidth}
    \includegraphics[width=0.54\linewidth]{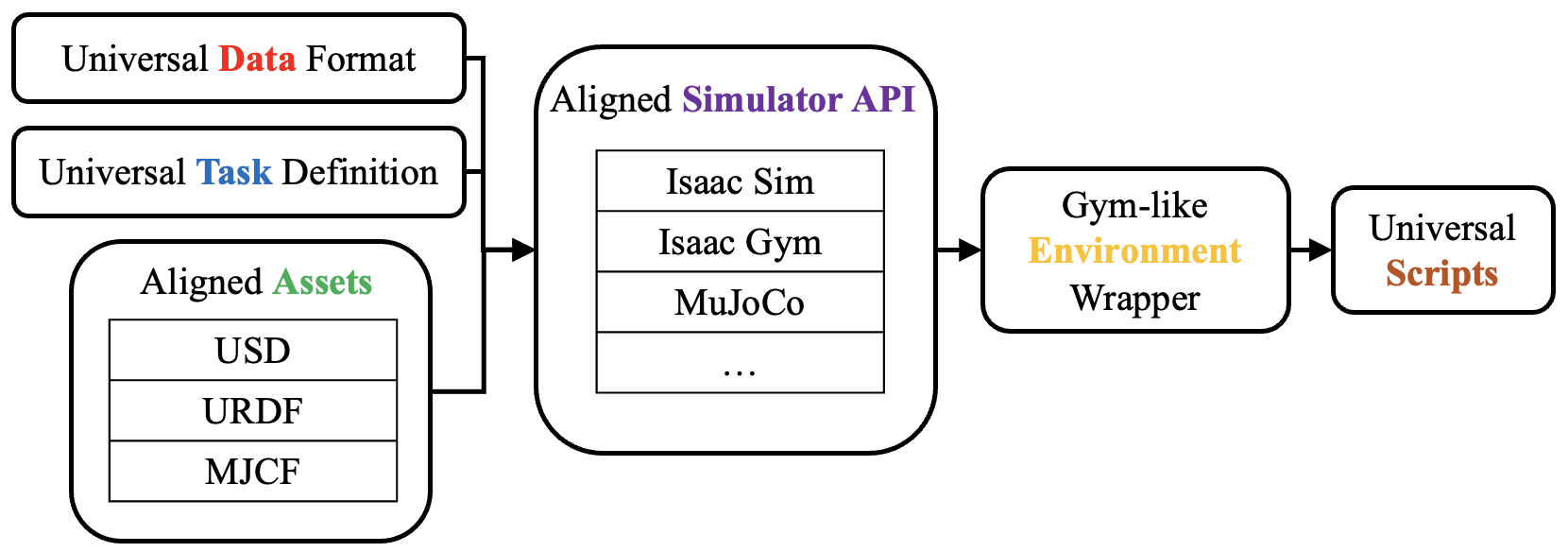}
    \caption{Comparison between the \textsc{MetaSim} and the other simulation environments. Left: Other simulator and benchmark, using self-defined data format, simulator-associated assets, simulator-dependent task definition, and scripts. Right: The \textsc{MetaSim}, decoupling all components to be agnostic to specific simulators or benchmark environments.}
    \label{fig:infra_implement}
\end{figure*}

The \textsc{MetaSim} framework is a unified simulation framework as shown in~\fref{fig:infra_implement}. On the front-end side, it provides user-friendly Gym APIs as well as easy-to-use parallel environment support. On the back-end side, it supports multiple simulators to allow seamless transfer of tasks across simulators. Users only need to master simple skills on writing a simulator-agnostic \texttt{MetaConfig} configuration class, the environment will then be automatically instantiated with the designated back-end simulator.

\subsection{\texttt{MetaConfig} Configuration System}

The \textsc{MetaSim} framework uses \texttt{MetaConfig}, a unified configuration class to describe a scenario in simulation environments.

We designed a configuration system that set up the simulator, define the tasks, set up the domain randomization. In order to run the same setting of environments across different simulators, the configuration system is defined to be simulator-agnostic as much as possible. For simulator-specific settings (\eg \xspace rendering mode, physics engine solver type, \etc), there is a seperate simulator-specific part which defines those things. 

To make changing the settings and debug more easily, we design the configuration system in a Hydra~\cite{Yadan2019Hydra}-like way, making each item in the configuration system can be modified from commandline just like Hydra~\cite{Yadan2019Hydra}. The configuration system is implemented based on Python dataclass, and could therefore use Python type annotation to help user use them.

In order to run the tasks seamlessly across all simulators, it is necessary to define them in a simulator-agnostic way. We configure the task and define its objects list, robot in use, success checker and the reward. The success checker is used to determine when the task is successfully execucated, and is the most difficult part in task definition. To standardize, we offer some structured success checker templates which cover the most cases, and leave option for users to define a callback function for flexibility to implement those stuctured success checker could not cover.

\subsection{Aligned Simulation APIs}

\textsc{MetaSim} support different simulator backends, including Isaac Sim~\cite{IsaacSim}, Isaac Gym~\cite{makoviychuk2021isaacgym}, MuJoCo~\cite{todorov2012mujoco}, PyBullet~\cite{coumans2021pybullet}, SAPIEN~\cite{xiang2020sapien}, CoppeliaSim~\cite{coppeliaSim,james2019pyrep}. The framework is implemented in Python, as these simulators either natively support Python or provide Python APIs.

Common simulator operations are unified in a \texttt{Handler} class.
Each handler supports only tree basic APIs: \texttt{get\_state()}, \texttt{set\_state()} and \texttt{step()}. The \texttt{get\_state()} method takes a descriptive Python dict (\eg, \texttt{\{object\_name: \{'pos': ..., 'rot': ..., '...': ...\}\}}) as input, and returns current simulation states according to the dict in another Python dict structured in the same manner. The \texttt{set\_state()} method also takes a descriptive Python dict as input, and modifies current simulation states to the ones included in the dict. The \texttt{step()} method will prompt the simulation to proceed one timestep.
% which defines two core methods: \texttt{get\_states} and \texttt{set\_states}. Here, \texttt{states} are represented as \texttt{\{object\_name: \{'pos': ..., 'rot': ..., '...': ...\}\}}, where \texttt{object\_name} is the unique identifier of an object, \texttt{pos} denotes its 3D position, and \texttt{rot} represents its orientation as a quaternion. The format is extendable, supporting optional fields like \texttt{dof\_pos} for joint positions in articulated objects and \texttt{dof\_pos\_target} for joint position targets when applying actions. The optional fields can also be introduced to ensure compatibility with soft-body objects.

\subsection{Gym API Wrappers}

To support building learning environments, we define an \texttt{Env} class built on top of \texttt{Handler}. It offers Gymnasium-like APIs (\texttt{step}, \texttt{reset}, \texttt{render}, and \texttt{close}), implementing these methods by leveraging the underlying \texttt{Handler} methods.

It is worth noting that most simulation environments provide the  underlying APIs (corresponding to our \texttt{Handler}) and upper-level environments (corresponding to our \texttt{Env}) seperately, such as SAPIEN~\cite{xiang2020sapien} with ManiSkill~\cite{tao2024maniskill3}, Isaac Sim~\cite{IsaacSim} with IsaacLab~\cite{mittal2023orbit}, CoppeliaSim~\cite{coppeliaSim}/PyRep~\cite{james2019pyrep} with RLBench~\cite{james2019rlbench}, and MuJoCo~\cite{todorov2012mujoco} with MuJoCo Playground~\cite{zakka2025mujocoplayground}. This fact proves our \texttt{Handler} and \texttt{Env} two-level abstraction reasonable.

% \subsection{Frontend APIs}

% The \textsc{MetaSim} framework provides unified APIs for different simulator back-ends.

% Each simulator handler supports only tree basic APIs: \texttt{get_state()}, \texttt{set_state()} and \texttt{step()}. The \texttt{get_state()} method takes a descriptive Python dict as input, and returns current simulation states according to the dict. The \texttt{set_state()} method takes a descriptive Python dict as input, and modifies current simulation states to the ones included in the dict. The \texttt{step()} method will prompt the simulation to proceed one timestep.

% The 

\subsection{Backend Support}

\subsubsection{IsaacSim}

Isaac Sim~\cite{IsaacSim} is an advanced robotics simulation platform developed by NVIDIA. By leveraging high-fidelity physics, GPU acceleration, and photorealistic rendering, it enables rapid prototyping, testing, and deployment of AI-driven robotics solutions in virtual environments. Through seamless integration with NVIDIA's Omniverse framework, Isaac Sim~\cite{IsaacSim} offers robust features such as domain randomization, sensor simulation, and support for large-scale reinforcement learning, making it a powerful tool for both research and industrial applications.

% A key advantage of Isaac Lab is its compatibility with the Isaac ROS infrastructure, which includes valuable models such as foundationpose~\cite{foundationposewen2024,bundlesdfwen2023} and cuRobo~\cite{curobo_report23}, among others.

\subsubsection{IsaacGym}
Isaac Gym~\cite{makoviychuk2021isaacgym} is a physics simulation environment designed for reinforcement learning research. Although it remains available for download, official support has ended. Nevertheless, multiple works published before 2024—such as hora~\cite{qi2023hand}, humanoid-gym~\cite{gu2024humanoidgymreinforcementlearninghumanoid}, and IPC-graspsim~\cite{kim2022ipcgraspsimreducingsim2realgap}—were developed using Isaac Gym.

Key features of Isaac Gym include support for importing URDF and MJCF files with automatic convex decomposition, a GPU-accelerated tensor API for managing environment states and actions, and a range of sensors (\eg, position, velocity, force, torque). Additional capabilities include runtime domain randomization of physics parameters, Jacobian and inverse kinematics support, and customizable friction settings.

\subsubsection{MuJoCo}

MuJoCo~\cite{todorov2012mujoco} is a physics engine and simulation framework designed to accurately model the dynamics and control of complex robotic systems in real-time. Its name, MuJoCo, stands for Multi-Joint dynamics with Contact, highlighting its primary emphasis on efficient computation of contact forces and multi-joint dynamics. The engine supports advanced features such as frictional contact models, user-defined actuators, and customizable sensor modalities, allowing researchers and developers to prototype, test, and refine control algorithms across a wide range of robot morphologies and tasks. 

A key strength of MuJoCo is its computational precision, which enables high simulation throughput and real-time interactive control. It supports rigid-body dynamics, articulated mechanisms, and a variety of constraints, making it suitable for tasks involving locomotion, manipulation, and reinforcement learning. Furthermore, MuJoCo’s flexible XML-based model description streamlines creating and modifying simulated environments, providing a straightforward way to experiment with novel designs. The compatibility between MuJoCo and Brax offers a high-speed, differentiable pipeline crucial for reinforcement learning. This powerful blend of accuracy, speed, and flexibility has solidified MuJoCo’s status as a leading choice in robotics research and machine learning, particularly for advanced control, motion planning, and reinforcement learning applications~\cite{brax2021github}.

\subsubsection{Genesis}
Genesis~\cite{xian24genesis} is a comprehensive physics platform developed for robotics and physics simulation research, unifying multiple core capabilities in a single environment. At its foundation is a universal physics engine, rebuilt from the ground up to simulate diverse materials and physical phenomena while seamlessly integrating various solvers. Alongside this engine, Genesis provides a swift, Python-friendly robotics simulation toolkit, an efficient photo-realistic rendering system, and a data-generation module that converts natural language prompts into multi-modal datasets. We leverage the Genesis backend to support loading, simulation, and rendering in \textsc{RoboVerse} workflow.

\subsubsection{SAPIEN}

SAPIEN~\cite{xiang2020sapien} is a robot simulation framework that allows highly efficient simulation and rendering of robotic tasks. It uses PhysX~\cite{physx} as the underlying physics engine. We supported the released version Sapien 2.2 for the \textsc{MetaSim} framework.

% SAPIEN 2.2 only natively supports single-environment simulation. Therefore,
We use the multipocessing library to support parallel environments in the \texttt{Handler} class for Sapien. When instantiating the environment from configurations, a desired number of processes are forked to run the simulation of different environments. To support the \texttt{get\_states} and \texttt{set\_states} API, data for different environments are distributed to different processes, and the return values are then gathered.

\subsubsection{PyBullet}

PyBullet~\cite{coumans2020} is a fast and easy-to-use robotics simulator. It uses its own physics solvers for accurate and efficient simulations. We supported the released version PyBullet 3.2 for the \textsc{MetaSim} framework.

% PyBullet 3.2 only natively supports single-environment simulation.
We use the same techniques as for Sapien to achieve parallel-environment simulation.

% \subsubsection{Pyrep} 

% \todo{feishi}

\subsection{Hybrid Simulation Implementation}
\textsc{MetaSim} allows launching two simulators in one single process with one command. Taking our demo collection command as example: {\scriptsize \texttt{python collect\_demo.py ---sim=mujoco ---renderer=isaaclab ---task=\$task}}. The implementation is illustrated in \Cref{code:hybrid_sim}.

\begin{lstfloat}
\begin{lstlisting}[style=mystyle]
class HybridEnv:
    def __init__(self, env_physic: Env, env_render: Env):
        ...
    def step(self, action):
        env_physic.handler.set_states(action=action)
        phys_states = env_physic.handler.get_states()
        env_render.handler.set_states(states=phys_states)
        env_render.handler.refresh_render()
        states = env_render.handler.get_states()
        return ...
\end{lstlisting}
\vspace{-0.5cm}
\caption{Pseudocode for implementing hybrid simulation using two different simulator environments simultaneously. The core of this implementation is using \texttt{states} as a unified representation across both simulation environments.}
\label{code:hybrid_sim}
\end{lstfloat}
\section{Asset Conversion}

\subsection{Asset types}
The diverse landscape of robotic assets, stemming from prior research
initiatives~\cite{zhu2020robosuite, james2019rlbench, mu2021maniskill} and a multitude of software platforms~\cite{todorov2012mujoco, makoviychuk2021isaacgym, xiang2020sapien}, necessitates a robust strategy
for managing a wide array of file formats. To facilitate dependable cross-simulator
training and uphold data integrity throughout the development lifecycle, the establishment
of an efficient and reliable asset conversion pipeline is of paramount importance~\cite{erez2015simulation}.
Such a pipeline is crucial for ensuring seamless interoperability, minimizing potential
data loss or inaccuracies, and promoting the uniform application of metadata and
configurations across disparate simulation environments. A selection of frequently
encountered asset formats includes, but is not limited to, MuJoCo XML control files~\cite{todorov2012mujoco},
URDF files~\cite{calli2015benchmarking}, and USD files~\cite{IsaacSim}.

The three predominant file formats in robotics simulation: MJCF, URDF, and USD. Each of them serves distinct purposes and offers unique capabilities. MJCF (MuJoCo Configuration Format) stands out for its exceptional expressiveness in physics simulation, featuring sophisticated capabilities to model complex dynamical systems including tendons, actuators, and advanced joint configurations, along with an integrated compiler for handling complex compile-time computations~\cite{todorov2012mujoco}. URDF (Unified Robot Description Format), while more constrained in its feature set, has emerged as the de facto standard in robotics due to its remarkable cross-platform compatibility and universal adaptability across various simulation environments including Isaac Sim~\cite{IsaacSim}, Isaac Gym~\cite{makoviychuk2021isaacgym}, MuJoCo~\cite{todorov2012mujoco}, Gazebo, and PyBullet~\cite{coumans2021pybullet}, making it ideal for robot model exchange despite its limitations in representing parallel mechanisms or complex sensor configurations~\cite{calli2015benchmarking}. USD (Universal Scene Description), originally developed by Pixar Animation Studios, excels in high-fidelity rendering and scene composition through its sophisticated layering system and variant sets~\cite{Deitke2023ObjaverseXLAU}, making it particularly valuable for applications requiring advanced visual properties and collaborative workflows~\cite{vMaterials}, although its physics simulation capabilities are more limited compared to dedicated robotics formats like MJCF~\cite{erez2015simulation}.

\begin{table}[htbp]
    \centering
    \caption{Comparison of Robot Description Formats}
    \begin{tabularx}
        {\columnwidth}{|X|c|c|c|} \hline \textbf{Features} & \textbf{MJCF} &
        \textbf{URDF} & \textbf{USD} \\ \hline Basic Geometries &
        {\color{green}\ding{51}} & {\color{green}\ding{51}} &
        {\color{green}\ding{51}} \\ \hline Mesh Support &
        {\color{green}\ding{51}} & {\color{green}\ding{51}} &
        {\color{green}\ding{51}} \\ \hline Texture Support &
        {\color{green}\ding{51}} & Limited & {\color{green}\ding{51}} \\ \hline
        Material Properties & {\color{green}\ding{51}} & Basic &
        {\color{green}\ding{51}} \\ \hline Physics Properties &
        {\color{green}\ding{51}} & {\color{green}\ding{51}} & Limited \\ \hline
        Joint Types & Many & Basic & Basic \\ \hline Collision Properties &
        Advanced & Basic & Advanced \\ \hline Deformable Objects &
        {\color{green}\ding{51}} & {\color{red}\ding{55}} &
        {\color{green}\ding{51}} \\ \hline Animation Support & Limited &
        {\color{red}\ding{55}} & {\color{green}\ding{51}} \\ \hline Scene
        Composition & Basic & {\color{red}\ding{55}} & Advanced \\ \hline File
        Format & XML & XML & ASCII/Binary \\ \hline
    \end{tabularx}
    \label{tab:format_comparison}
\end{table}

\subsection{Conversion Pipeline}
Given that our simulation pipeline primarily utilizes Isaac Sim for rendering while many of our assets are originally stored in MJCF format, a two-stage conversion pipeline (MJCF → URDF → USD) becomes necessary and advantageous. This approach leverages URDF as an intermediate format for several reasons. First, while direct conversion from MJCF to USD is theoretically possible, such conversion would be complex and error-prone due to MJCF's rich feature set for physics properties (like tendons and actuators) that lack direct equivalents in USD~\cite{wan2020deepclaw}. Instead, converting to URDF first allows us to standardize the robot's basic kinematic and dynamic properties in a format that has well-established conversion tools and widespread support. The subsequent URDF to USD conversion benefits from Isaac Sim's robust URDF importing capabilities, which have been extensively tested and optimized for robotics applications. This two-stage pipeline thus ensures more reliable asset conversion while maintaining essential physical properties and compatibility across different simulation environments.

\subsubsection{MJCF to URDF conversion}
We implemented our own MJCF to URDF converter by first parsing everything with MuJoCo's
MJCF importer, then exporting all texture, collision mesh and joint information to
the correct URDF format. The inspiration is taken from Genesis~\cite{xian24genesis}, which they built their own class for each asset
object that encode all joint, texture and mesh information. We then recursively generate
the body information to URDF and align everything with texture.

\paragraph{Parsing Link, Joint, and Body Information from the MJCF file}
To parse link, joint, and body information from the MJCF file, we leverage MuJoCo's
parsing capabilities to load the MJCF XML into a MuJoCo model structure. From
this parsed model, we employ a recursive approach, starting from the root body
and descending into each child body to systematically process the hierarchical
structure. For each body, we extract detailed link properties such as name, position,
orientation, inertial characteristics, and associated geometry. Simultaneously,
we parse joint information connected to each body, including joint type, limits,
and axis of motion. All of this extracted link and joint data is systematically
organized and stored in dictionary structures. These dictionaries serve as intermediate
representations, holding all the necessary information from the MJCF model in a
structured format that is readily accessible for subsequent stages of the URDF
conversion process.

\paragraph{Aligning Meshes and textures}
The management of collision meshes across existing asset libraries presents a notable
challenge, as these assets are typically stored in various formats including .msh,
.obj, and .stl files. While URDF natively supports .obj and .stl formats, the conversion
of .msh files into URDF-compatible formats requires careful consideration. Although
MuJoCo's repository provides a conversion utility for transforming .msh files to
.obj format—accomplished by parsing the .msh files through the MuJoCo interface
and subsequently exporting vertex and face information—this approach introduces
potential complications with texture mapping alignment.

The complexity arises from the specific requirements of texture files, which are
predominantly stored as albedo PNG files. These textures depend on precise UV
mapping coordinates within the .obj file to ensure proper alignment. The current
.msh to .obj conversion utility provided in the MuJoCo repository does not
adequately address texture support, leading to potential misalignment issues in the
converted models. This limitation is particularly evident in comprehensive
robotics frameworks such as LIBERO~\cite{liu2023libero} , where both static
and articulated objects frequently exhibit texture alignment discrepancies following
the .msh to .obj conversion process.

Fortunately, we discovered that many asset collections maintain redundant mesh
representations, often including a properly UV-mapped .obj file alongside the .msh
file, typically sharing the same filename or designated as "textured.obj". Leveraging
this observation, we implemented a robust mesh alignment pipeline that follows a
hierarchical decision process:

\begin{itemize}
    \item First, the system searches for an existing .obj file within the same directory
        as the .msh file

    \item If found, this pre-existing .obj file is utilized, ensuring proper
        texture alignment

    \item In the absence of a pre-existing .obj file, the system proceeds with the
        .msh to .obj conversion

    \item In the latter case, users receive a warning notification regarding
        potential texture misalignment issues
\end{itemize}

Following the mesh format resolution, the pipeline systematically maps these
processed mesh files back to their corresponding links within the URDF structure,
maintaining the integrity of the robot's geometric representation while preserving
texture information where possible.

\paragraph{Building URDF}
The assembling procedure after all the conversions become very aparent: we first
processes robot links and joints, incorporating their properties and
relationships into the URDF format. This automated approach ensures a robust and
flexible method for generating URDF files, accommodating a wide range of robot
configurations and properties derived from the preceding conversion steps.

Even though this pipeline roughly works for most of the MJCF, for some specific MJCF
files in some specific folder, we have to modify our conversion approach on a
case by case basis. Below is a table for some special treament we employed to specific
packages, and its conversion success rate:

Despite the general efficacy of the described pipeline across a broad spectrum
of MJCF assets, it is important to acknowledge that certain MJCF files,
particularly those within specific packages or directories, necessitate bespoke conversion
strategies. These exceptions arise due to the inherent complexity and
variability in MJCF file structures across different projects and asset
libraries. To address these unique cases, we have adopted a tailored approach, implementing
case-specific modifications to our conversion pipeline as required. The
subsequent table details instances where such specialized treatment has been
applied, along with the corresponding conversion success rates achieved for each
package.

% \begin{table}[h]
%     \centering
%     \caption{Package-specific conversion details and success rates}
%     \begin{tabularx}
%         {\linewidth}{|X|c|X|} \hline \textbf{Package Name} & \textbf{Success
%         Rate} & \textbf{Special Treatments} \\ \hline RLAfford & XX\% &
%         {\color{red}TODO for Charlie} \\ \hline LIBERO & XX\% & $\bullet$ Custom
%         texture path remapping
%         \newline
%         $\bullet$ Deletion of unnecessary geometries used for collision optimizations
%         \newline
%         $\bullet$ Special orientation fix for meshes\\ \hline ManiSkill2 & XX\% &
%         {\color{red}TODO for Charlie} \\ \hline SAPIEN Assets & XX\% &
%         {\color{red}TODO for Charlie} \\ \hline
%     \end{tabularx}
%     \label{tab:conversion_rates}
% \end{table}

% \paragraph{URDF to MJCF conversion}

% {\color{red}TODO for Charlie}

\subsubsection{URDF to USD conversion}
Isaac Sim has implemented a robust solution for converting URDF files to USD
format. The conversion process comprehensively preserves the robot's structural and
kinematic information, including joint hierarchies, geometric properties, and
physical attributes. The implementation demonstrates exceptional fidelity in translating
complex robotic descriptions, ensuring that all essential components—such as
joint configurations, collision geometries, and visual representations—are
accurately encoded in the resulting USD files.

Given the proprietary nature of Isaac Sim's conversion implementation, we utilize
their framework as an external tool in our pipeline. This approach leverages the
proven reliability and performance of Isaac Sim's converter while maintaining
compatibility with our broader system architecture. The conversion process serves
as a critical bridge between standard robotics formats and the high-performance USD
representation required for our simulation environment.

% \paragraph{USD to URDF conversion}
% Similar to the USD to URDF converter, Isaac Sim provides built-in functionality for
% exporting USD files back to URDF format. While we acknowledge the potential
% utility of bidirectional conversion, we did not conduct extensive testing of the
% USD-to-URDF export capability. This decision was primarily motivated by our
% workflow architecture, where all original robot assets are maintained in either MJCF
% or URDF formats, following established practices in the robotics community. Since
% these formats serve as our primary source representations, and USD is used
% exclusively for simulation purposes, maintaining backward conversion fidelity was
% not critical for our pipeline's functionality.

% % 

\section{Task and Data Migration}

\subsection{ManiSkill}

ManiSkill~\cite{mu2021maniskill,gu2023maniskill2,tao2024maniskill3} provides a series of robotic manipulation tasks under single-arm or dual-arm settings.
\paragraph*{Tasks and assets} We migrate basic single-arm tasks and demonstrations to \textsc{RoboVerse}, including the pick-and-place tasks like \texttt{PickCube} and \texttt{PickSingleYCB}, as well as the insertion tasks like \texttt{PegInsertionSide} and \texttt{PlugCharger}. The corresponding assets are manually crafted with primitives or process from the mesh files, with proper physics API set up.
\paragraph*{Demonstrations} For each task, a great number of demonstration trajectories are available in the released data. Noteworthy, the data does not come with the initial scene states, which are obtained by replaying the demonstrations within the SAPIEN simulator. With the specified seed set, the states are recovered by the random samplers.The success checkers are implemented according to the task designs.

\subsection{RLBench}
RLBench~\cite{james2019rlbench} is a large-scale benchmark and learning environment for robotic manipulation, featuring $100$ diverse, hand-designed tasks ranging in complexity, from simple actions like reaching to multi-stage tasks like opening an oven and placing a tray inside. Each task includes an infinite supply of demonstrations generated via waypoint-based motion planning.
\paragraph*{Tasks and assets} We roll out ${\sim}2K$ trajectories in RLBench~\cite{james2019rlbench} for each task, and migrate them to \textsc{RoboVerse}.

\subsection{CALVIN}

CALVIN~\cite{mees2022calvin} provides 6-hour teleopreation trajectories on 4 environments, each involve an articulated table with three blocks in blue, pink, or red.

\paragraph*{Tasks and assets} We migrate the demonstrations in all 4 environments and transform the original assets (URDF for the table, and primitives for the cubes) into USD files with proper physics APIs.

\paragraph*{Demonstrations} We segment the trajectories according to the text annotations, which specified the task category (\textit{\eg}, \texttt{PlaceInSlider}), the text annotation (\textit{\eg}, \textit{place the red block in the slider}), and the timestamps of the demonstration segment. The states of the first frame is adopted as the scene initial states.

\paragraph*{Success checkers} We carefully implement the success checkers according to the original implementation to make sure the failed executions can be filtered out. This is because the coarsely annotated timestamps in the dataset, which may cause the failed execution in part of the demonstrations.

\subsection{Meta-World}
Meta-World~\cite{yu2019metaworld} is a widely used benchmark for multi-task and meta-reinforcement learning, comprising 50 distinct tabletop robotic manipulation tasks involving a Sawyer robot.

\paragraph*{Tasks and Assets}
We integrate five representative tasks into \textsc{RoboVerse}: \textit{DrawerOpen}, \textit{DrawerClose}, \textit{DoorClose}, \textit{WindowOpen}, and \textit{WindowClose}. The corresponding assets are manually converted from MJCF to USD files with appropriate physics APIs.

\textit{Demonstrations}: As the benchmark does not provide demonstrations, we generate trajectories for each task by rolling out reinforcement learning policies from~\cite{xie2024textreward}.

\subsection{Open6DOR}
Open6DOR is a benchmark for open-instruction 6-DoF object rearrangement tasks, which requires embodied agents to move the target objects according to open instructions that specify its 6-DoF pose. 
\paragraph*{Tasks and Assets}
The synthetic object dataset comprises 200+ items spanning 70+ distinct categories. Originally derived from YCB~\cite{7254318} and Objaverse-XL~\cite{Deitke2023ObjaverseXLAU}, the objects are carefully filtered and scaled using a standardized format of mesh representation. 
Overall, the Open6DOR Benchmark consists of
5k+ tasks, divided into the position-track, rotation-track, and 6-DoF-track, each providing manually configured tasks along with
comprehensive and quantitative 3D annotations. 
\paragraph*{Success checkers}
We determine success by comparing the target object's final pose with the annotated ground-truth pose range.

\subsection{ARNOLD}
ARNOLD~\cite{gong2023arnold} is a benchmark for language-conditioned manipulation. The benchmark uses motion planning and keypoints for robot manipulation tasks, focusing on fine-grained language understanding.

\paragraph*{Tasks and Assets}: We integrate six out of eight tasks from ARNOLD~\cite{gong2023arnold} into \textsc{RoboVerse}: picking up objects, reorienting objects, opening/closing drawers, and opening/closing cabinets.

\textit{Demonstrations}: As the benchmark does not use trajectory-level demonstrations, we use motion planning for trajectory generation to interpolate between keypoints

\subsection{\texttt{robosuite} \& MimicGen}
\texttt{robosuite}~\cite{zhu2020robosuite} provides a suite of task environments for robotic manipulation, built on the MuJoCo physics engine. Each task is implemented as a separate class, with most configuration details embedded in the source code. Based on these environments, MimicGen~\cite{mandlekar2023mimicgen} offers thousands of demonstrations, serving as a widely used benchmark for imitation learning.

\paragraph*{Tasks and Assets}
For tasks with separate object description files (MJCF), we directly migrate the corresponding assets through our Asset Conversion pipeline. However, some tasks contain hard-coded assets within the source code, such as a hammer composed of multiple cubes, cylinders and other primitives with carefully designed relative poses. To integrate these tasks, we will manually reconstruct the assets within our framework. We also argue that hard-coded asset and task definitions, as opposed to modular task descriptions, are not scalable for future robotic task benchmarking.

\paragraph*{Demonstrations} 
We convert MimicGen demonstrations into our format. Specifically, we transform the robot actions from 6-DoF Cartesian space representations to joint space. Additionally, the state of the first frame is adopted as the initial scene state.

\paragraph*{Success Checkers} 
We meticulously implement success checkers based on the original definitions to ensure failed executions are effectively filtered out.

% \subsection{RoboCasa}
% \todo{Jiangran}

\subsection{SimplerEnv}
SimplerEnv is a set of tasks and methods designed to do trustworthy benchmarking in simulation for manipulation policies that can reflect the real-world success rate.

There are in total $25$ different tasks in SimplerEnv. We ignore all tasks that are just a subset of another task and migrated in total $6$ tasks and $52$ object assets to \textsc{RoboVerse}. The tasks all use Google Robot.

SimplerEnv provided some controller models trained with RT-1~\cite{brohan2022rt} and RT-X~\cite{padalkar2023open} dataset. We did not use the trajectories from the dataset directly because some environmental settings are different from the environments from SimplerEnv. We used the trained model to collect trajectories. Hooks are inserted into the original SimplerEnv codebase to extract and maintain the recordings at different stages of simulation. We then rollout the model trained with RT-1 dataset on each task to collect the trajectories.  

\subsection{GAPartNet}
For tasks in GAPartNet~\cite{geng2023gapartnet}, we generate both motion planning~\cite{geng2023gapartnet} and reinforcement learning~\cite{geng2023partmanip} trajectories. GAPartNet is implemented in Isaac Gym~\cite{makoviychuk2021isaacgym} with various articulated objects. To integrate it into \textsc{RoboVerse}, we first align all articulated object initial states to the MetaSim format and convert the asset format to USD for compatibility across different simulators.

For trajectory generation:

(1) \textbf{Motion Planning}: GAPartNet~\cite{geng2023gapartnet} introduces a part-centric manipulation approach. We roll out heuristics to generate manipulation trajectories, providing three demonstrations per part with different object and part initial states.
(2) \textbf{Reinforcement Learning Rollout}: The follow-up work, PartManip~\cite{geng2023partmanip}, proposes several reinforcement learning methods. We re-train all policies based on our robot setup and roll out trajectories for dataset collection.
With aligned task configurations, trajectories, and assets, we successfully adapt GAPartNet into \textsc{RoboVerse}.

\subsection{GAPartManip}
Instead of providing direct demonstrations, GAPartManip~\cite{cui2024gapartmanip} offers a large-scale, part-oriented, scene-level dataset with annotations for actionable interaction poses. We utilize the mesh-level grasping pose annotations in this dataset to generate diverse demonstrations for articulated object manipulation.

\paragraph*{Tasks and Assets}

We currently implement two tasks: \texttt{OpenBox} and \texttt{OpenToilet}. For the \texttt{OpenBox} task, we collect 12 object assets from the \textit{Box} category in the original dataset. For the \texttt{OpenToilet} task, we gather 30 objects from the \textit{Toilet} category. We convert these assets into USD files with appropriate physics APIs to ensure compatibility with our simulation environment.

\paragraph*{Demonstrations}

We generate demonstrations for our tasks in simulation using motion planning with cuRobo~\cite{curobo_report23}. First, we filter potential grasping poses for the target object link by assessing their feasibility through motion planning. Specifically, we discard poses that the end-effector cannot reach or that would cause a collision between the robot and the object.
Next, we generate an end-effector pose trajectory to complete the task using heuristics. Based on the object's kinematic tree, we could define an ideal trajectory. We then apply motion planning to perform inverse kinematics, computing the corresponding joint poses of the robot along this trajectory. Finally, we execute the planned trajectory in simulation to verify task completion, saving successful trajectories as demonstrations. The entire demonstration generation process is conducted in Isaac Sim~\cite{IsaacSim}.

\paragraph*{Success Checkers}

To determine task success, we require the manipulated object to be opened by at least 60 degrees for all tasks.

\subsection{GraspNet-1B}
GraspNet-1B~\cite{fang2020graspnet} is a general object grasping dataset for predicting 6 DoF grasping pose given partial pointcloud input. 
It contains 256 realworld tabletop scenes consists of total 88 different objects.
We carefully filter out 58 objects as our target grasping objects based on the availability of purchasing real items because we need to evaluate our policies to grasp them in the real world experiments.
To generate grasping demonstrations, we use cuRobo~\cite{curobo_report23} as motion planner to generate robot end effector trajectories starting from a fixed initial pose and ending to an target object grasping pose.
The grasping pose is obtained from the grasping annotations used to train GraspNet~\cite{fang2020graspnet}.
We also randomized the object positions to generate more diverse layouts.
Finally, we validate the trajectories in our framework and filter out invalid ones by controlling robots to follow the generated grasping trajectories.
In the end, we successfully generated about 100k valid grasping trajectories.

\subsection{GarmentLab}
GarmentLab~\cite{lu2024garmentlab} is the first robotic manipulation benchmark for deformable object and garment manipulation. It integrates 10 categories of versatile garment assets and the total number of USD assets reaches 6k. To generate manipulation demonstrations, we directly roll out the trajectories provided by the official codebase in Isaac Sim and collect the corresponding state information in a parallel process. Although the trajectory provided by the official codebase is limited and hard-coded, we further extend the number of demonstrations by applying different garments and textures, and all the demonstrations are validated by the original success checker. Finally, we have successfully collected 6k trajectories.

\subsection{UniDoorManip}
UniDoorManip~\cite{li2024unidoormanip} provides an articulated manipulation environment reflecting different realistic door manipulation mechanisms, and a large-scale door dataset containing 6 door categories with hundreds of door bodies and handles stored in URDF format. We convert those door assets into USD format with physics APIs from Isaac Sim and manually further verify the correctness of the joint-link relationship. Demonstrations are collected by directly rolling out the hard-coded trajectories in Isaac Gym. We eventually collect about 1k successful legal demonstrations.

\subsection{RLAfford}

RLAfford~\cite{geng2023rlafford} investigates the generalization ability of Deep
Reinforcement Learning models on articulated object manipulation tasks with the
presence of a computer vision model that is co-trained with it in an end-to-end
manner. This work provided a dataset of articulated objects and 8 tasks for benchmarking.

In \textsc{RoboVerse}, we have adapted 4 tasks (open cabinet, open drawer, close cabinet,
close drawer) and in total 40k trajectories from RLAfford.

In the task adaptation, we included 40 articulated objects from the RLAfford
dataset, and uses the same robot description file from RLAfford. Then we record
1000 trajectories for each object in its corresponding task.

The trajectory recording is achieved with several hooks we inserted into the
original RLAfford codebase. The hooks are used to extract and maintain the recordings
at different stages of simulation. We evaluated the released RLAfford model with
hook-inserted scripts. In the initialization stage, objects and robots are
initialized with randomization, their pose, and DoF information are recorded. For
each simulation step, the DoF position information of objects and robots is
recorded in the trajectories. In the end, for each object, a separate trajectory
file of 1000 different trajectories is saved in the \textsc{RoboVerse} supported format.

\subsection{LIBERO}
LIBERO~\cite{liu2023libero} manages data loading and task execution through a combination of INIT(initialization files), BDDL (Behavior Description Definition Language), and HDF5 datasets. Specifically, the initialization files define scene layouts, object properties, and basic task goals; the BDDL format captures semantic details and object affordances; and the HDF5 files store structured data such as object positions and robot actions for dynamic retrieval at runtime.

To migrate a LIBERO task into  \textsc{MetaSim}, we parse the relevant BDDL file to identify which objects are involved and what type of manipulation context is required. Then we get the robot and object initial states from the INIT files, followed by the corresponding robot actions from the HDF5 dataset. These elements are combined into our \texttt{PKL} file format while also recording the participating objects in our \texttt{MetaCfg}. This process ensures that all necessary components of a LIBERO task, initial states, and action data, are fully translated and ready for execution in \textsc{MetaSim}.

We further augment the data by randomly sampling initial positions around each LIBERO demonstration, thus increasing the effective number of demos well beyond the original 50 per task. The spatial sampling range is dynamically chosen based on the task context and object dimensions, ensuring that the augmented configurations remain physically plausible.

\section{Task Generation}

\subsection{Robot \& Object Generation Protocol}
\label{sec:robot-object-generation}

\noindent

\begin{figure*}[t]
    \centering
    \includegraphics[width=\linewidth]{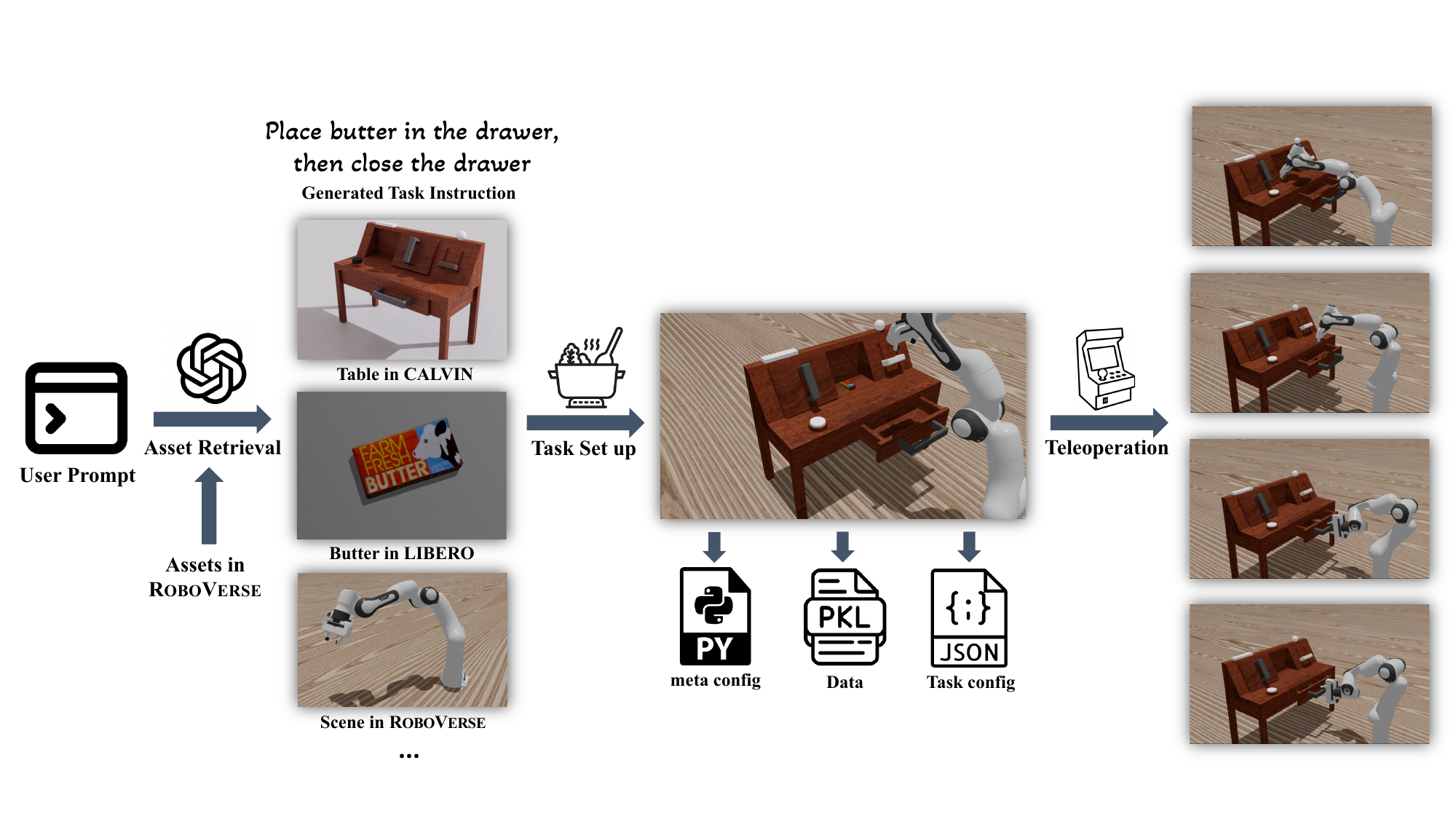}
    \caption{Illustration of the two-phase generation protocol. A user prompt guides the LLM to propose an overall task and item list. The system then refines object positions and merges them into a final initial state.}
    \label{fig:ai-assisted-pipeline}
\end{figure*}

Our task generation pipeline (\fref{fig:ai-assisted-pipeline}) begins with a \emph{user prompt} describing the desired theme or constraints of a robotic task (\eg, "place the butter in the drawer and close it"). From here, the system proceeds in two main phases, mediated by large generative model calls:

\begin{enumerate}
    \item \textbf{\texttt{call\_gpt\_to\_generate\_task()}: Conceptual Task Generation.} 
    This initial function queries the model for a high-level task overview. It requests:
    \begin{itemize}
        \item A \textit{unique task name} (\eg, ``ButterDrawerTask'').
        \item A short, \textit{human-readable instruction} (\eg, ``Place the butter in the drawer, then close the drawer.'').
        \item A candidate list of \textit{robots and objects} to appear in the scenario, referencing an internal asset library (see below).
    \end{itemize}
    The large generative model draws on its generative abilities to propose creative or contextually relevant tasks, while remaining loosely guided by the user prompt~\cite{wang2023robogen, wang2023gensim, pmlr-v229-ha23a, zhou2022learning}. As shown in \fref{fig:ai-assisted-pipeline}, the model might retrieve a ``drawer'' asset from a different benchmark and a ``butter'' asset from a separate dataset, combining them into a single scene idea.

    \item \textbf{\texttt{call\_gpt\_to\_get\_init\_state()}: Physical Layout Refinement.} 
    After receiving the conceptual description, we call the model again to specify \texttt{x,y} coordinates for each listed item. During this second phase, user can provide the prompts that include minimal bounding constraints (\eg, permissible table edges, object height) to help modelgenerate various initial states by few-shot learning.
  
\end{enumerate}

% \noindent

\noindent\textbf{Asset Library.} 
To ground the large generative model’s outputs in realistic data, we maintain an \emph{asset library} (via JSON files) that describes each robot or object’s core attributes (\eg, assets filepath, default rotation, size). The two core functions above selectively pull from this library.

% \noindentx

\noindent\textbf{Input and Output Format.}
\begin{itemize}
    \item \textit{Input}: A user prompt (\eg, “create a tabletop scene with a random container and a snack food”). The pipeline loads relevant asset definitions and passes them to the large generative model calls.
    \item \textit{Output}: A merged \texttt{init\_state} or “initial state” dictionary capturing the initial state config needed for simulation: the chosen robot/object list, each item’s final \texttt{x,y,z} coordinate, and the textual instructions, as shown in the right half of \fref{fig:ai-assisted-pipeline}.
\end{itemize}

\section{Teleoperation}

\begin{figure*}[t]
  \centering
  \makebox[\linewidth]{
    \includegraphics[width=0.99\linewidth, trim={0mm, 0mm, 0mm, 0mm}, clip]{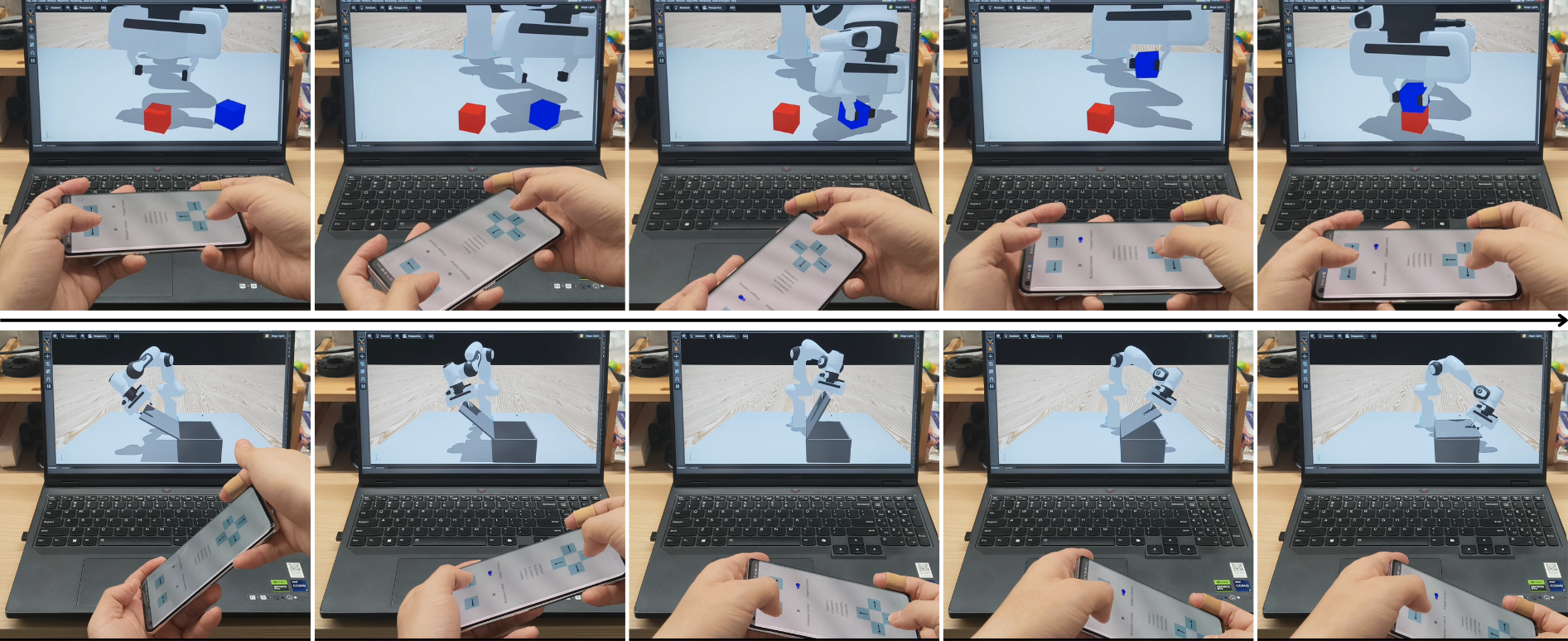}
    }
  \caption{
  Sequential demonstration of smartphone-based control for stack cube and close box tasks.
  }
  \label{fig:2_teleop_demo}
\end{figure*}

Ensuring flexible and intuitive remote operation is critical in robotic teleopration system, particularly when collecting large volumes of high quality data. 
In this work, we designed a suite of input methods to facilitate robot teleopration within the \textsc{MetaSim} infrastructure. 
By supporting keyboard, DualSense Joystick, smartphone, and VR-based controls, our system accommodates varying user preferences and experimental needs. This section details our design rationale, implementation steps, and practical considerations for each control interface. 

\subsection{Keyboard}
Keyboard input is an accessible method for controlling robots in simulation. Our implementation supports multi-key combinations for diagonal movement and enables full six-degree-of-freedom manipulation of the end effector. Translational movement follows the world coordinate frame (UP: +X, DOWN: -X, LEFT: +Y, RIGHT: -Y, ‘e’: +Z, ‘d’: -Z), while rotations in the local EE frame are controlled via ‘q’/‘w’ (roll), ‘a’/‘s’ (pitch), and ‘z’/‘x’ (yaw). The spacebar toggles the gripper. 
To assist users and avoid hotkey conflicts with the simulation viewer, we provide an operation window displaying instructions using \texttt{pygame}.
While efficient and hardware-independent, this method lacks 3D spatial representation, reducing user intuition. Additionally, Euler angle-based rotation control risks gimbal lock, potentially leading to loss of rotational degrees of freedom and failure in certain configurations. 
\subsection{Smartphone}

\begin{figure}[!t]
    \centering
    % trim=left bottom right top, unit=points
    \includegraphics[width=1\linewidth, trim=0 0 0 0, clip]{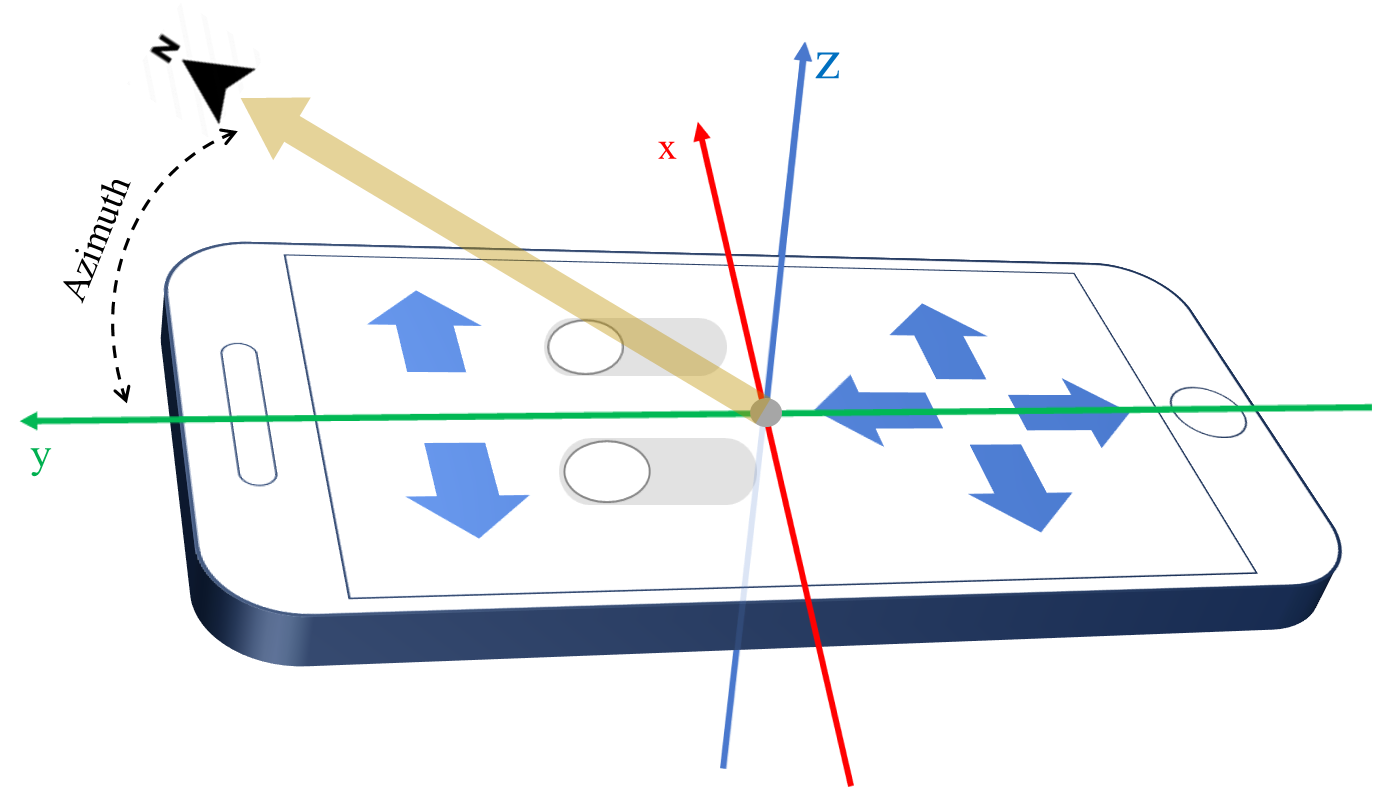}
    \caption{
    Visualization of the smartphone’s local coordinate system, world-frame orientation, and app functionality: six buttons control translation, and two switches toggle orientation control and gripper state.
    }
    \label{fig:smartphone}
\end{figure}

\begin{figure}[!t]
    \centering
    % trim=left bottom right top, unit=points
    \includegraphics[width=1\linewidth, trim=0 0 0 0, clip]{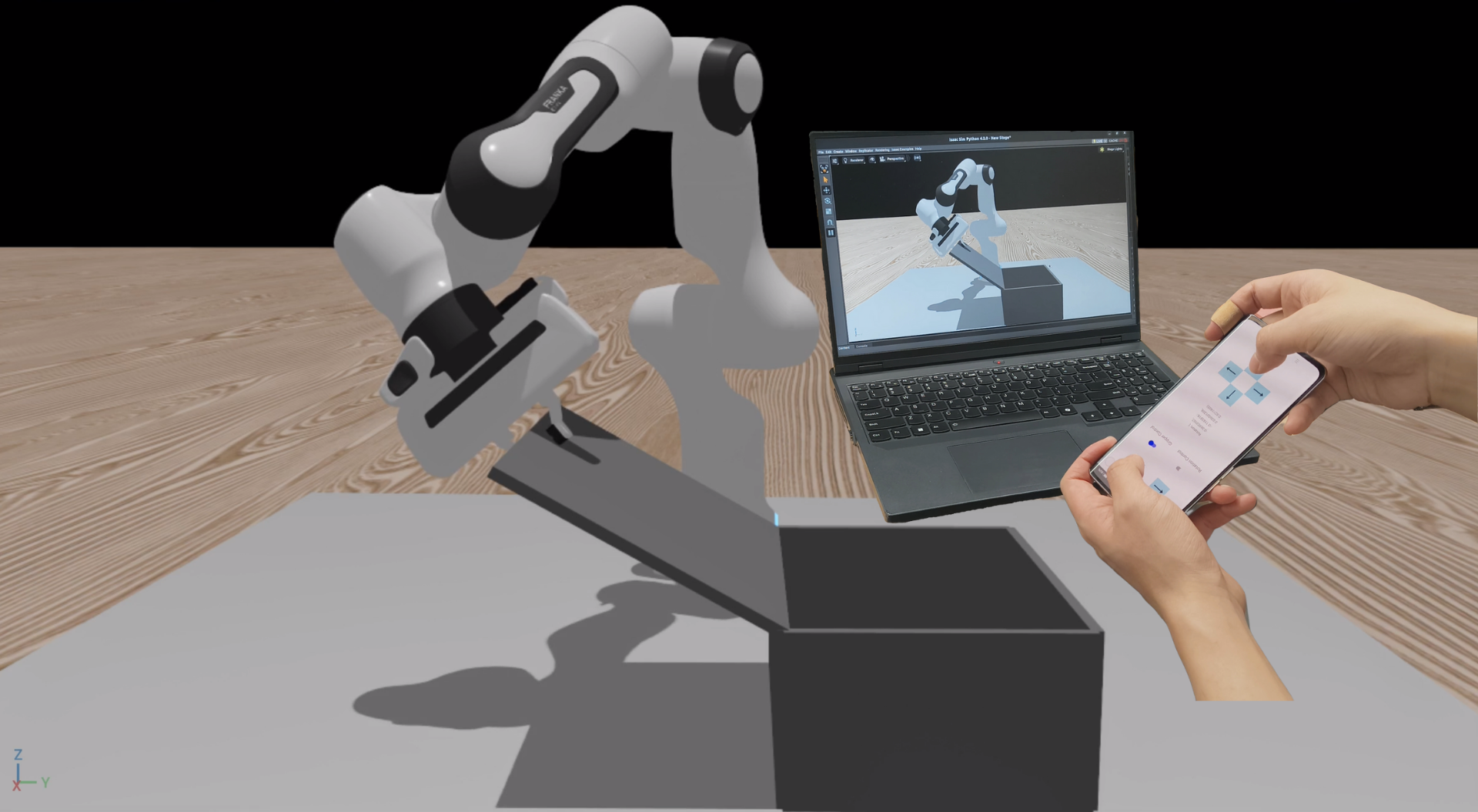}
    \caption{
    The smartphone app enables 6-DoF control using orientation sensing and multi-touch buttons for translation commands, while the simulated robot's movements are visualized in real-time on the workstation.
    }
    \label{fig:human_playing_hands}
\end{figure}

Modern smartphones, equipped with advanced sensors and wireless communication, offer an ideal low-cost solution for intuitive teleoperation from any location. 
However, existing smartphone-based 6-DoF methods, such as those relying on accelerometers or vision-based Visual Inertial Odometry (VIO) systems (\eg, ARKit), suffer from instability due to sensor noise, low update rates, or weak visual features~\cite{ha2024umi, mandlekar2018roboturk, mandlekar2020human, mandlekar2021matters}.  
Additionally, no open-source Android app exists for such implementations.
To overcome these limitations, we adopt a hybrid approach: using smartphone orientation for motion control and on-screen buttons for precise translation. 
Unlike the keyboard interface, where roll, pitch, and yaw are controlled incrementally via discrete keypresses (i.e., delta orientation adjustments), the smartphone directly provides absolute orientation data in the form of quaternions. Quaternions, due to their compactness and immunity to gimbal lock, allow for a more stable and accurate representation of the smartphone’s orientation in the world frame.
As illustrated in \fref{fig:smartphone}, real-time data from the smartphone’s inclination, rotation, and magnetic field sensors is fused to compute spatial orientation with ±5° accuracy at a frequency of 50 Hz. This data is transmitted via WebSocket, ensuring low-latency communication. 
The app interface features six buttons for translation control in the local coordinate system and two switches for toggling orientation updates and gripper control. 
Multi-touch input is supported to enable users to send combined control signals, such as simultaneous movement along multiple axes, improving control flexibility and efficiency. 
As shown in the \fref{fig:human_playing_hands} and \fref{fig:2_teleop_demo}, tilting the smartphone controls the gripper's orientation, while combining multi-touch signals from on-screen buttons enables precise and complex manipulation in 3D space.
However, to mitigate magnetic interference, users should maintain a minimum distance of 10 cm from strong magnetic sources such as laptops and other electronic devices. 
This design optimizes resource utilization, providing a high-precision 6-DoF remote operation experience at minimal cost, rivaling professional-grade teleoperation systems.

\begin{table}[bt]
    \huge
    \centering
    \begin{adjustbox}{width=\textwidth}
        \begin{tabular}{c|cc|ccc}
            \toprule
            \multirow{2}{*}{Method} & \multicolumn{2}{c|}{Simple} & \multicolumn{3}{c}{Language-conditioned Grasping} \\ 
           &  PickCube & MoveSliderLeft & Object Set 1 & Object Set 2 & Object Set 3 \\
            \midrule
            OpenVLA~\cite{kim24openvla} & 40.0 & 45.0 & 46.0 & 33.3 & 14.4\\
            Octo~\cite{octo_2023} & 50.0 & 30.0  & 42.0 & 14.4 & 2.2 \\
             \bottomrule
        \end{tabular}
    \end{adjustbox}
    \caption{\textbf{Vision-Language-Action (VLA) Model Results on \textsc{RoboVerse} Imitation Learning Benchmark.} Constrained with time and resources, we report VLA models' results on two simple tasks from \textsc{RoboVerse} and grasping tasks with diverse and challenging language instructions. We split 58 objects in GraspNet into three sets, each containing progressively more challenging objects based on their geometry.}
    \label{table:vla}
\end{table}

\subsection{Others}
Beyond keyboard and smartphone controls, our system incorporates support for DualSense Joysticks and VR controllers. The DualSense joystick provides ergonomic advantages and high-fidelity analog inputs for nuanced velocity control, mapping triggers and joysticks seamlessly to robot motion. The VR interface enhances spatial awareness and precision by enabling natural gestures and directional cues for control.

Future work could extend VR capabilities by integrating haptic feedback to improve user immersion and task accuracy. Additionally, the modular design of our system facilitates the integration of emerging input devices with minimal development effort.

\section{Real2Sim Toolset for Asset and Task Generation}

\subsection{Overview}
The \emph{Real2Sim} toolset, specifically \emph{Video2URDF}, provides a systematic pipeline to reconstruct environment geometry and robotic assets from monocular video input. By leveraging advanced reconstruction techniques, this pipeline produces meshes and unified robot descriptions that can be used in simulation-based experiments. In doing so, it helps bridge the gap between real-world data and simulated environments, enabling more accurate and comprehensive benchmarking~\cite{lou2024robogsphysicsconsistentspatialtemporal}

\subsection{Components}
\subsubsection{Gaussian Splatting Reconstruction} \label{sec:gaussian_splatting} The first step in the pipeline involves \emph{Gaussian splatting~\cite{kerbl3Dgaussians}}, which converts monocular video frames into a set of Gaussian kernels for rendering~\cite{ye2024gsplatopensourcelibrarygaussian}. This representation captures key scene features such as depth, color, and collision boundaries in a compact and efficient way. As a result, it provides a visually faithful preview of the scene and serves as an intermediate step before detailed mesh reconstruction.

\subsubsection{Mesh Reconstruction} \label{sec:mesh_reconstruction} Once the high-level scene structure is represented by Gaussian splatting, we perform mesh reconstruction to obtain a more precise geometric model utilize tsdf extraction~\cite{zeng20163dmatch,ye2024gaustudio,ye2024stablenormal,Huang2DGS2024}. This step recovers the meshes of: \begin{itemize} \item The \emph{environment}, including rigid, immovable structures (\eg, a table). \item The \emph{manipulatable object}, which is central to the task at hand. \item The \emph{robotic arm and end effector}, assumed to have a deterministic configuration during real-to-sim and sim-to-real transitions. \end{itemize}

We use a visual-language model (VLM) and available CAD design information to generate a unified URDF (or MJCF) description for these components. This division of the workspace follows the notion of \texttt{worldconfig} in cuRobo~\cite{curobo_report23}, ensuring that each element of the scene (robot, object, environment) is cleanly separated and can be easily adapted or replaced as needed.

\subsubsection{Loading the URDF into the Simulation Environment} \label{sec:urdf_loading} After the URDF (or MJCF) files are generated, the final step is to import them into a simulator, such as MuJoCo~\cite{todorov2012mujoco} in \textsc{RoboVerse}. This allows researchers to configure tasks that accurately reflect real-world scenarios, forming a benchmark for training and evaluating robotic manipulation algorithms. The resulting simulated environment benefits from high-fidelity geometry and a consistent representation of the physical workspace.

\subsubsection{Real-to-Sim boost Sim-to-Real Performance} 
\begin{figure*}[htbp]
    \centering
    \includegraphics[width=\linewidth]{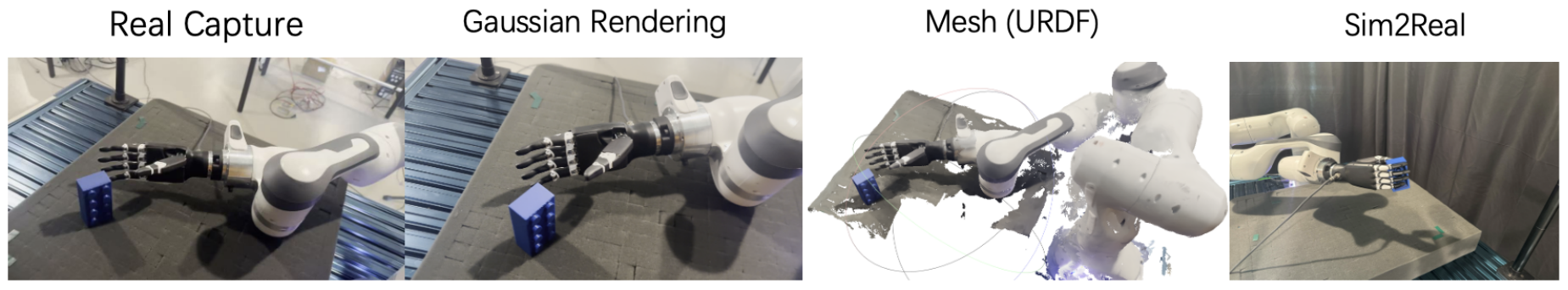}
    \vspace{-0.8cm}
    \caption{Visualization of our real2sim pipeline for robotic grasping.}
    \label{fig:real2sim_policy}
\end{figure*}

We train model on our real2sim module compared with DexGraspNet~\cite{zhang2024dexgraspnet}, demonstrating 80\% success rate compared to the 50\% baseline from DexGraspNet. We use our real2sim assets in physics-based simulations that closely replicate real-world grasping conditions, enabling robust grasp execution. See \fref{fig:real2sim_policy} for visualization.

\subsection{Limitations and Challenges.} While the Real2Sim pipeline effectively reconstructs most of the relevant geometry, it struggles with completely unseen meshes and complex material properties~\cite{zheng2025gstargaussiansurfacetracking}. Furthermore, parameters such as friction and mass are inherently difficult to estimate purely from visual data, introducing uncertainties that may affect simulation fidelity. Despite these challenges, Real2Sim offers a powerful approach to rapidly generating simulation-ready assets for benchmarking in robotic manipulation tasks.

\section{Domain Randomization}

\subsection{Scene Randomization}
For scene randomization, we curate 3D simulatable scene assets from existing 3D scene datasets~\cite{fu20213d,gong2023arnold,deitke2022️proc,jia2025sceneverse}. Specifically, we convert all assets to the USD format for integration. Additionally, we employ the articulated scene generation method PhyScene~\cite{yang2024physcene} to create realistic scenes with articulated objects and mix the generated room-level scenes with house-level 3D scenes like ProcTHOR for greater diversity. We replay demonstrations in these scenes by selecting surfaces (\eg, floors, tables) that provide sufficient workspace, guided by heuristic-based spatial constraints, following~\cite{gong2023arnold}.

\subsection{Visual Material Randomization}
It's optinal to attach random visual material to object surfaces. Visual materials are randomly selected from a curated subset of ARNOLD~\cite{gong2023arnold} and vMaterials~\cite{vMaterials}, providing more the 300 high-quality visual material candidates.
Additionally, user can also randomize the reflection properties of a given visual material, by setting roughness, specular, and metallic to random number between 0 and 1.

\subsection{Light Randomization}
Two lighting configurations are supported: distant light and cylinder light arrays. For distant lighting, the polar angle of the light source is randomized. For cylinder lighting, a randomly generated $n \times m$ matrix of cylinder lights, each with a randomized size, is added at a fixed height above the agents. In both configurations, the intensity and color temperature of the lights are randomized within physically plausible ranges.

\subsection{Camera Randomization}
A total of 59 candidate camera poses are carefully selected, with the majority oriented to face the robot directly and a smaller subset positioned at side-facing angles.

\section{Navigation and Locomotoin Tasks}

\subsection{Navigation Tasks}

To integrate vision-and-language navigation into Isaac Sim, we first correct the error-containing instructions by refining incorrect punctuation and grammar using ChatGPT. Next, we validate the ground truth trajectory by sweeping the robot's 3D model (based on the ground truth trajectory) through the scene. The trajectory is deemed invalid if collisions occur between the robot and the scene. Additionally, we adopt the same evaluation metrics as VLN-CE~\cite{Krantz2020BeyondTN}. For controlling the robot, we provide two different types of mobile embodiments, including a Unitree Go2 robot dog and a JetBot wheeled robot, making our task suitable for a variety of policies (with different navigation capabilities).

\begin{figure*}[hbt] \centering
    \includegraphics[width=\linewidth]{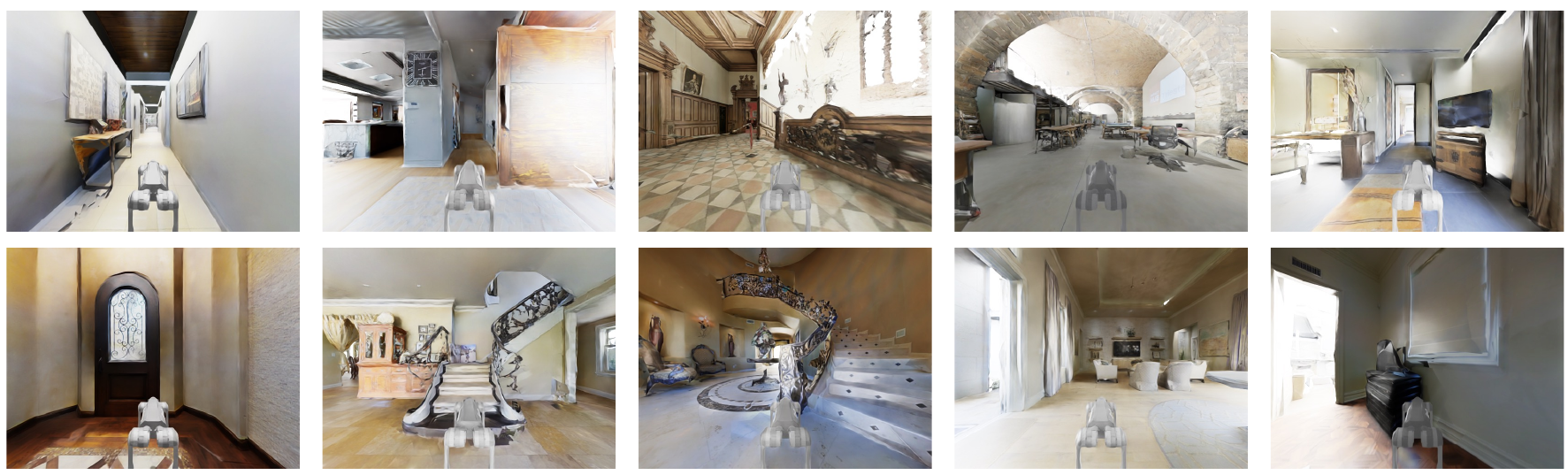}
    \caption{Navigation gallery. We deploy the Unitree Go2 robot within Matterport 3D environments. The robot is tasked with navigating the environment based on provided instructions. }
    \label{fig:navigation_gallery}
\end{figure*}

\subsection{Humanoid Tasks}
We migrated the data samples from the Humanoid-X dataset~\cite{uh1}, and re-implemented the inference pipeline of UH-1~\cite{uh1} in our framework. We use the Unitree-H1-2 humanoid robot as the simulated embodiment and set up the locomotion and humanoid pose control task in our framework. The humanoid pose control task is to control the humanoid robot to follow some human poses while maintaining its stability on the ground. The demonstrated poses in our framework include arms crossing, boxing, dancing, left and right punch, playing violin, playing guitar, praying, waving to a friend, \etc. Our pretrained policy can successfully follow the demonstrated pose to control a humanoid robot while maintaining stable locomotion in IssacGym, and also obtain a decent performance in IssacLab. The humanoid environment and task configurations are highly flexible and scalable, and we are able to support more humanoid pose control tasks from Humanoid-X without modifying the infrastructure.

\subsection{HumanoidBench}

HumanoidBench~\cite{sferrazza2024humanoidbench} is a high-dimensional simulated benchmark designed to accelerate research in humanoid robot learning, focusing on whole-body locomotion and manipulation tasks. The benchmark features a humanoid robot equipped with dexterous hands, enabling a wide range of complex interactions in human-like environments.

\textit{Tasks and Assets}: We migrate three fundamental locomotion tasks: run, walk, and stand. These tasks are designed to test the robot's ability to maintain balance, achieve forward motion, and stabilize in a standing position. The primary robot model used is the Unitree H1, augmented with two dexterous Shadow Hands, though the environment supports other humanoid models such as Unitree G1 and Agility Robotics Digit.

\textit{Demonstrations}: While HumanoidBench does not provide pre-collected demonstrations, it supports the use of reinforcement learning algorithms to generate task-specific policies. The benchmark is designed to facilitate learning from scratch, with dense and sparse reward structures to guide the learning process.

\textit{Success Checkers}: Each task in HumanoidBench is equipped with a success checker that evaluates task completion based on predefined criteria. For example, in the walk task, success is determined by the robot's ability to maintain a forward velocity of 1 m/s without falling, while in the stand task, success is measured by the robot's ability to maintain a stable upright posture for a specified duration.

\begin{figure*}[hbt]
    \centering
    \includegraphics[width=\linewidth]{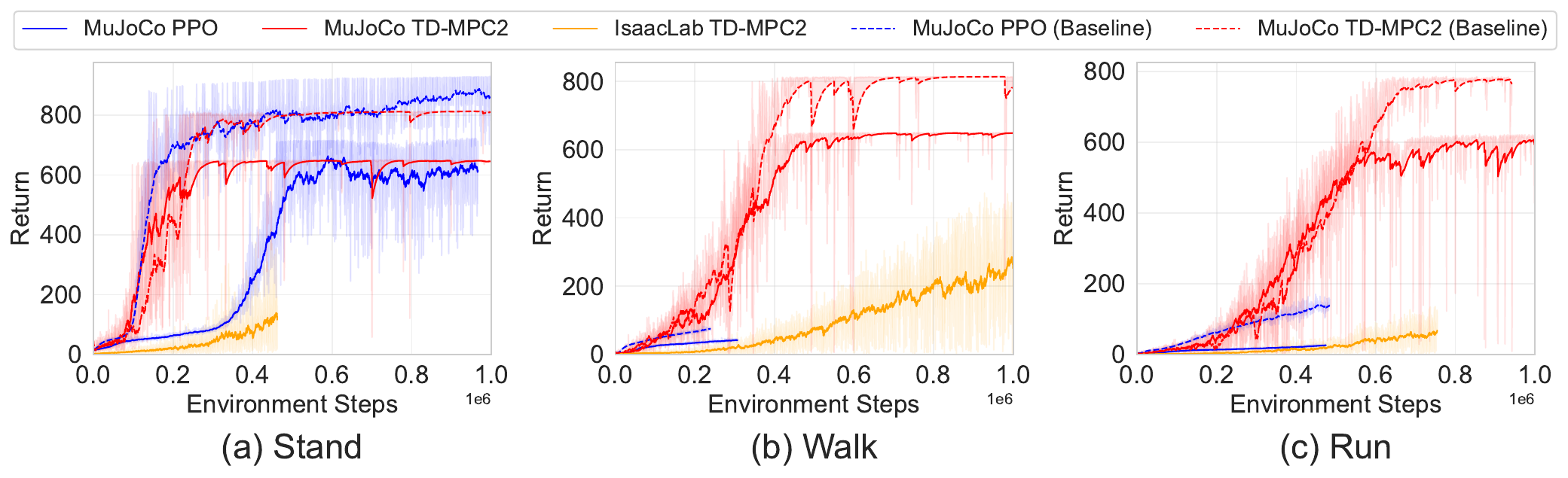}
    \caption{\textbf{Learning curves of RL algorithms on HumanoidBench task migratation}: We also run PPO in the Isaac Sim handler in RoboVerse, but it is not visible in the plot since it only achieves very low returns.}
    \label{fig:HB_task_comparison}
\end{figure*}

\begin{figure*}[hbt]
    \centering
    \includegraphics[width=\linewidth]{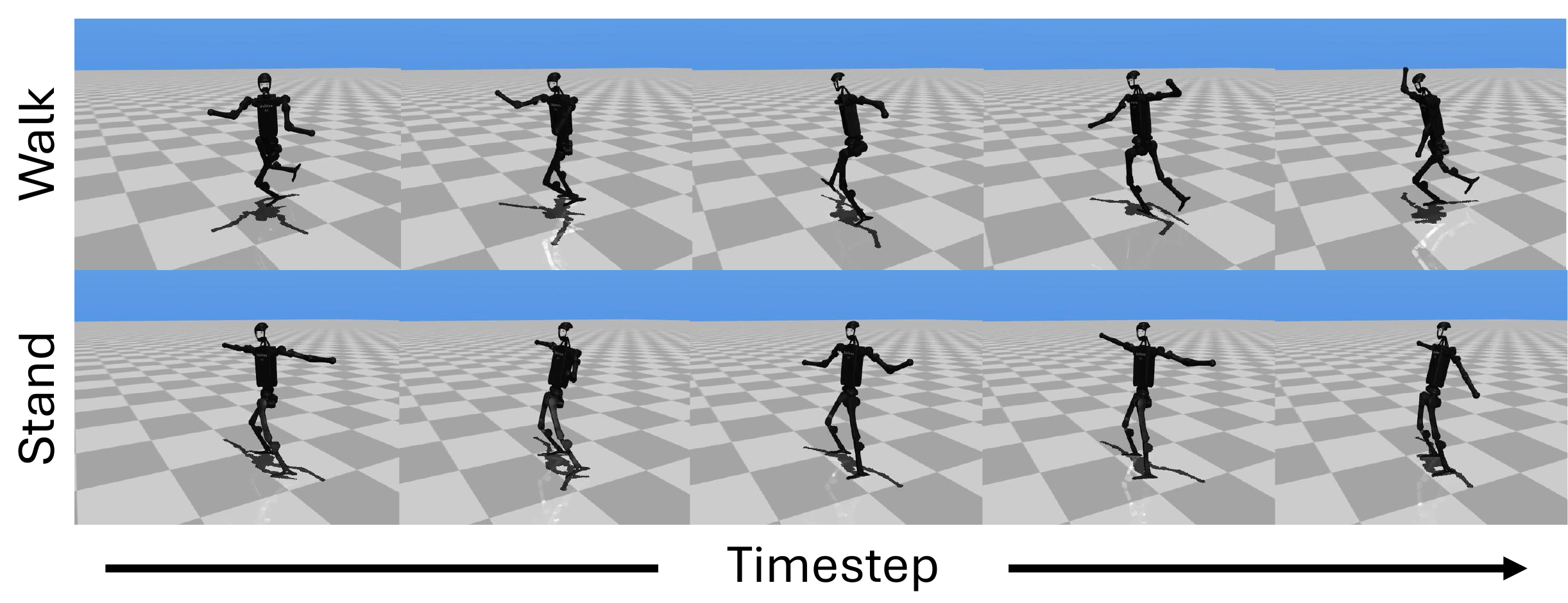}
    \caption{Demonstration of TD-MPC2 policys trained in the RoboVerse MuJoCo simulator on the Walk and Stand tasks migrated from the HumanoidBench benchmark}
    \label{fig:HB_policy_demo}
\end{figure*}

\textit{Experiment and Results}: We trained the walk, stand, and run tasks in both the RoboVerse MuJoCo and Isaac Sim handlers using the PPO and TD-MPC2~\cite{hansen2022tdmpc,hansen2024tdmpc2} algorithms, and compared the results with the HumanoidBench baseline based on the original MuJoCo environment. As shown in \fref{fig:HB_task_comparison} and \fref{fig:HB_policy_demo}, the training curves from the RoboVerse MuJoCo handler eventually converged and approached the performance of HumanoidBench, validating the feasibility of the RoboVerse reinforcement learning infrastructure. Additionally, we trained the same tasks in the RoboVerse Isaac Sim handler with identical configurations. While training efficiency in Isaac Sim was comparatively lower under non-parallelized settings (to maintain configuration consistency), it still demonstrated a clear upward trend in reward accumulation. This confirms the rapid migration capability of the MetaSim framework and highlights its potential to enable sim-to-sim learning while leveraging the strengths of different simulators, such as Isaac Sim's support for GPU-accelerated large-scale parallel training.

\section{\textsc{RoboVerse} Benchmark Set up Details}

\subsection{Generalization Levels}
To systematically evaluate the generalization capability of a robot policy, we establish a benchmark based on a carefully curated asset set designed for domain randomization. This asset set encompasses a diverse range of environmental factors, including materials, textures, lighting conditions, scene configurations, and camera perspectives. By leveraging this set, we assess how well different policies generalize to unseen conditions. Specifically, we split the available assets into a {9:1 ratio for training and testing}, ensuring that the testing environment contains novel variations not encountered during training. Below, we detail the key components of this domain randomization setup:

\begin{itemize}
    \item \textbf{Table, Ground, and Wall.}  
    In tasks where a predefined scene is absent, we incorporate \textbf{walls (and ceilings)} to introduce structural complexity. Additionally, {customizable tables} are included for tasks requiring tabletop interactions. The visual materials applied to these elements are randomly sampled from a carefully curated subset of {ARNOLD}~\cite{gong2023arnold} and {vMaterials}~\cite{vMaterials}, ensuring a diverse range of appearances. The table features approximately {300 distinct material options}, while both the {wall and ground} have around {150 material choices} each. This variation enhances the robustness of the learned policy by exposing the model to a wide spectrum of surface appearances and textures.

    \item \textbf{Lighting Conditions.}  
    We introduce two distinct lighting scenarios: {distant lighting} and {cylinder light arrays}, each designed to test the adaptability of the learned policy to different illumination conditions.  
    \begin{itemize}
        \item \textbf{Distant Light:} The {polar angle of the light source} is randomized within a predefined range, influencing the way shadows and reflections appear in the scene.
        \item \textbf{Cylinder Light Arrays:} A randomized \( n \times m \) matrix of cylinder lights, varying in {size and intensity}, is placed at a fixed height above the agent.
    \end{itemize}
    In both configurations, {light intensity and color temperature} are randomly varied within reasonable limits to ensure that the model encounters a broad range of lighting effects.

    \item \textbf{Camera Poses.}  
    To further evaluate the robustness of visual perception, we carefully select {59 candidate camera poses}, strategically positioned to provide diverse viewpoints. The majority of these cameras are oriented {directly towards the robot}, ensuring consistent frontal perspectives, while a subset is placed at {side-facing angles} to introduce additional viewpoint variability.

    \item \textbf{Reflection Properties.}  
    To simulate the wide range of reflective surfaces encountered in real-world environments, we randomize key material reflection properties, including {roughness, specular intensity, and metallic characteristics}. These properties are adjusted within reasonable physical ranges to ensure that the robot policy learns to handle various levels of surface reflectivity.
\end{itemize}

By integrating these domain randomization techniques into our benchmark, we create a controlled yet diverse testing environment that challenges the {generalization ability} of different robot policies. This setup ensures that trained policies are not merely overfitting to a limited set of conditions but are instead capable of adapting to a broader range of real-world variations.

\subsection{\textsc{RoboVerse} Benchmark Protocol}
We rigorously design a training and evaluation protocol to ensure a structured and reliable assessment of the policy's performance. Given the training data, the policy learns to imitate the demonstrated behavior. For evaluation, we provide a standardized API that enables systematic assessment. As mentioned earlier, the training and evaluation follow a 9:1 ratio, ensuring that the policy is tested on novel scenarios not encountered during training.

\section{Policy Training Details}
\subsection{Implementation Details}
\label{supp:policy_implementation}
For specialist models, we train from scratch with action in $9$-dim robot joint state space. 
Diffusion Policy~\cite{chi2023diffusionpolicy} is implemented based on its original framework. We search several key hyperparameters, including observation and prediction length, to optimize performance for our tasks.
ACT~\cite{zhao2023learning} is implemented with the original architecture and hyper-parameters, except that the batch size has been increased to $512$, with learning rate correspondingly enlarged to $1e-4$ to accelerate convergence. We train ACT on one A100 GPU for $2000$ epochs and evaluate with the best checkpoints on the validation set.

For generalist models, the action is pre-processed into delta end-effector position space from absolute end-effector position space, and the gripper action is binarized to $\{0,+1\}$. Owing to the lack of time and resources, we are only able to fine-tune the generalist models in the single-task setting. For each task, OpenVLA~\citep{kim24openvla} is LoRA~\cite{hu2021loralowrankadaptationlarge} fine-tuned (rank$=32$) with 8 A100 GPU under official settings to convergence and reaches over 95\% action token accuracy as proposed by~\cite{kim24openvla} during the training stage. During evaluations, we employ cuRobo~\cite{curobo_report23} as the inverse-kinematics solver to transform the action to robot joint state space.
\subsection{Diffusion Policy} 
We implemented the training and validation code for \textit{Diffusion Policy} based on the requirements of our tasks and relevant research papers.

Modeling Diffusion Policy as Denoising Diffusion Probabilistic Models (DDPMs), we train a noise predictor network:
\begin{equation}
\widehat{\epsilon^{k}}=\epsilon_{\theta}\left(a^{k}, s, k\right)
\end{equation}
that takes in noisy actions $a^k$, current observations $s$, and denoising iterations $k$ and predicts the noise $\widehat{\epsilon^{k}}$.

As for observation $s$, We use ResNet18 to extract the features of scene images $f_{img}$ and use 3-layer MLP to extract the features of robot joint states $f_{robot}$. $f_{img}$ concatenating with $f_{robot}$ is just the conditioning input for Diffusion Policy.

During training, we randomly choose a denoising step $k$ and sample noise $\epsilon^k$ added to the unmodified sample $a^0$. Our training loss is the difference between $\epsilon^k$ and predicted noise:
\begin{equation}
    L_{DP} = MSELoss(\epsilon^k, \widehat{\epsilon^k})
\end{equation}

During inference time, our policy starts from random actions $a^K$ and denoises for $K$ steps to obtain the final action predictions. At each step, the action is updated following: 
\begin{equation}
    a^{k-1}=\alpha\left(a^{k}-\gamma \epsilon_{\theta}\left(a^{k}, s, k\right)+\mathcal{N}\left(0, \sigma^{2} I\right)\right)
\end{equation}, 
where $\alpha$, $\beta$ and $\gamma$ are hyperparameters.

\begin{figure*}[tb]
    \centering
    \includegraphics[width=\linewidth]{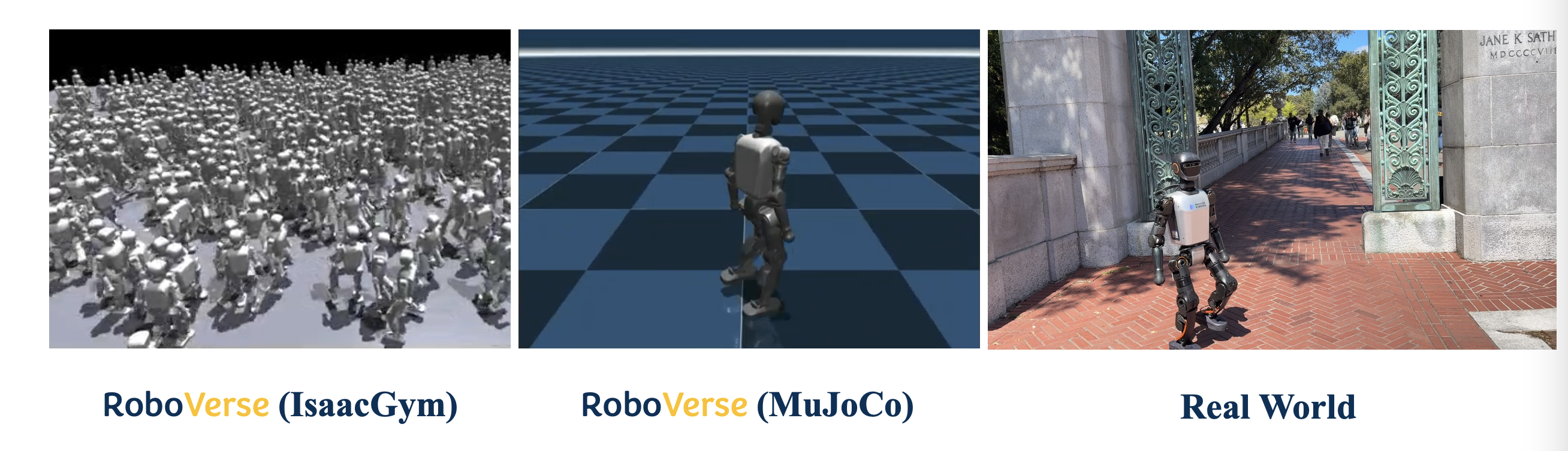}
    \caption{\textbf{Visualization of Sim-to-Sim-to-Real Experiments.}}
    \label{fig:s2s2r}
\end{figure*}

\section{World Model Details}
\subsection{Methodology} We adopt a video generation framework based on Latte~\cite{ma2024lattelatentdiffusiontransformer}—a transformer-driven latent diffusion model equipped with an efficient spatial-temporal attention mechanism. For action conditioning, we use frame-level Adaptive Layer Normalization~\cite{perez2017filmvisualreasoninggeneral} (AdaLN), following insights from IRASim~\cite{zhu2024irasimlearninginteractiverealrobot} that show more precise control of the gripper with frame-level conditioning compared to video-level conditioning.

In the forward pass, raw video frames are encoded using a frozen autoencoder from Stable Diffusion~\cite{podell2023sdxlimprovinglatentdiffusion}. The first frame serves as the initial condition, while noise is introduced into the latent representation of subsequent frames during training. Both the noise schedule and action conditions (gripper states with either Cartesian position plus orientation or joint position) are encoded by separate MLPs into latent space and then added together.

These noisy latent frames are then fed into a transformer composed of alternating spatial and temporal attention blocks, where action conditions are applied at each frame via AdaLN. For inference, we employ DDIM~\cite{song2022denoisingdiffusionimplicitmodels} as a denoising scheduler, using 200 sampling steps.

\subsection{Data Preparation} The DROID~\cite{khazatsky2024droid} dataset's episodes typically last from 120 to 360 frames. To amplify motion, we skip every 6 frames, effectively reducing the frame rate to 4 fps with sequence lengths from 20 to 60. In the \textsc{RoboVerse} simulation, we adjust the control frequency so that most episodes span 20 to 60 frames, mirroring the number of frames of DROID in one episode. We filter out any sequence shorter than 20 or longer than 60 frames, resulting in about 50,000 unique episodes from DROID. 

We only generate 50,000 unique \textsc{RoboVerse} episodes due to time and resource constraints. The full-scale \textsc{RoboVerse} is planned to train more capable world models in future works.

We exclude the gripper camera view because the model struggles with drastic camera pose changes, which leads to poor frame generation quality. Since we consider left and right camera views as separate samples, each dataset effectively doubles to 100,000 samples.

\begin{figure}[tb]
    \centering
    \includegraphics[width=\linewidth]{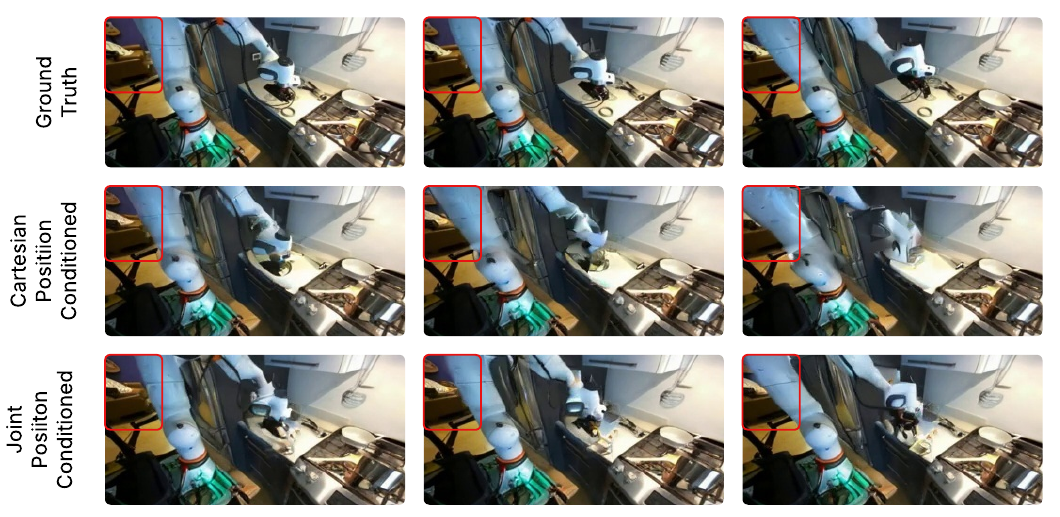}
    \caption{Visualization of ground truth and predicted frames by models conditioned on cartesian position (plus orientation) and joint position.}
    \label{fig:wm_joint_vs_cartesian}
\end{figure}

% \begin{figure}[ht]
%     \centering
%     \includegraphics[width=\linewidth]{fig/wm_roboverse.pdf}
%     \caption{Visualization of ground truth and predicted frames by models trained on DROID-RoboVerse dataset.}
%     \label{fig:world_model_roboverse}
% \end{figure}

\subsection{Experiments} Our experiments involve training three datasets, DROID-50K, RoboVerse-50K, and DROID-RoboVerse-100K, on 8 NVIDIA H100 GPUs. We use a spatial resolution of 240×320 and sequences of 16 frames per episode. Starting with a model of 100M parameters and a batch size of 16, training converges at around 100K steps on \textsc{RoboVerse} and 200K steps on DROID. 

We first compare Cartesian position plus orientation to joint positions as action conditions and find that using joint positions as action conditions yields more precise gripper movement control in frame generation, as shown in \fref{fig:wm_joint_vs_cartesian}. We believe it is due to joint positions being less ambiguous than Cartesian position plus orientation as the robot states representation. 

However, generation quality remains suboptimal when training on the DROID-50K or DROID-RoboVerse-100K datasets and validating on DROID samples due to the complexity of DROID scenes. Scaling the model to 500M parameters and reducing the batch size to 8 leads to better preservation of object geometry, as does the prediction of robot arm movement.

As discussed in the main paper, although the larger model trained on DROID-RoboVerse-100K shows an improved understanding of object shapes in DROID samples compared to the model trained on DROID-50K, it still struggles with intricate real-world physics. In contrast, training with RoboVerse-50K or DROID-RoboVerse-100K and validating on \textsc{RoboVerse} scenes produces more physically and geometrically consistent predictions. % as shown in \fref{fig:world_model_roboverse}. 

We believe it is because \textsc{RoboVerse} offers cleaner backgrounds, more comprehensive views of the robotic arm, and the implementation of domain randomization and augmentation. By comparison, many DROID frames contain cluttered backgrounds or incomplete arm visibility, creating challenges for learning robust temporal dynamics from raw pixels.

\end{document}